\documentclass[10pt,journal,compsoc]{IEEEtran}
\usepackage{amsmath,amsfonts}
\usepackage{algorithmic}
\usepackage{algorithm}
\usepackage{array}
\usepackage[caption=false,font=normalsize,labelfont=sf,textfont=sf]{subfig}
\usepackage{textcomp}
\usepackage{stfloats}
\usepackage{url}
\usepackage{verbatim}
\usepackage{graphicx}
\usepackage{booktabs}
\usepackage{multicol}
\usepackage{multirow}
\usepackage{color}
\usepackage{xcolor}
\usepackage{colortbl}
\usepackage{hyperref}
\usepackage{balance}
\usepackage{hyperref,ulem}
\definecolor{C1}{HTML}{93BFCF}
\definecolor{C2}{HTML}{A0C3D2}
\definecolor{C3}{HTML}{BDCDD6}
\definecolor{C4}{HTML}{EEE9DA}
\definecolor{C5}{HTML}{FFF1DC}
\definecolor{C6}{HTML}{E8D5C4}
\definecolor{C7}{HTML}{EEEEEE}
\definecolor{C8}{HTML}{BCEE68}

\ifCLASSOPTIONcompsoc
  \usepackage[nocompress]{cite}
\else
  \usepackage{cite}
\fi
\ifCLASSINFOpdf
\else
\fi
\begin{document}
%
% paper title
% Titles are generally capitalized except for words such as a, an, and, as,
% at, but, by, for, in, nor, of, on, or, the, to and up, which are usually
% not capitalized unless they are the first or last word of the title.
% Linebreaks \\ can be used within to get better formatting as desired.
% Do not put math or special symbols in the title.
\title{3D Gaussian Splatting as a New Era: A Survey}
%
%
% author names and IEEE memberships
% note positions of commas and nonbreaking spaces ( ~ ) LaTeX will not break
% a structure at a ~ so this keeps an author's name from being broken across
% two lines.
% use \thanks{} to gain access to the first footnote area
% a separate \thanks must be used for each paragraph as LaTeX2e's \thanks
% was not built to handle multiple paragraphs
%
%
%\IEEEcompsocitemizethanks is a special \thanks that produces the bulleted
% lists the Computer Society journals use for "first footnote" author
% affiliations. Use \IEEEcompsocthanksitem which works much like \item
% for each affiliation group. When not in compsoc mode,
% \IEEEcompsocitemizethanks becomes like \thanks and
% \IEEEcompsocthanksitem becomes a line break with idention. This
% facilitates dual compilation, although admittedly the differences in the
% desired content of \author between the different types of papers makes a
% one-size-fits-all approach a daunting prospect. For instance, compsoc 
% journal papers have the author affiliations above the "Manuscript
% received ..."  text while in non-compsoc journals this is reversed. Sigh.

\author{Ben Fei$^*$, Jingyi Xu$^*$, Rui Zhang$^*$, Qingyuan Zhou$^*$, Weidong Yang$^{\dagger}$, and Ying He$^{\dagger}$% <-this % stops a space
\IEEEcompsocitemizethanks{\IEEEcompsocthanksitem 
\textit{Ben Fei, Jingyi Xu, Rui Zhang, and Qingyuan Zhou contribute equally to this work and are joint first authors.
}
\IEEEcompsocthanksitem \textit{Corresponding authors: Weidong Yang and Ying He}

\IEEEcompsocthanksitem Ben Fei, Jingyi Xu, Rui Zhang, Qingyuan Zhou, and Weidong Yang are with the School of Computer Science, Fudan University, Shanghai, China, 200433 (e-mails: $\{\text{bfei21}|\text{jyxu22}|\text{22210240379}|\text{zhouqy23}\}$@m.fudan.edu.cn;  wdyang@fudan.edu.cn).
\IEEEcompsocthanksitem Ying He is with the School of Computer Science and Engineering, Nanyang Technological University, Singapore, 639798 (email: yhe@ntu.edu.sg).
\IEEEcompsocthanksitem This project was partially supported by the China Scholarship Council and the Ministry of Education, Singapore, under its Academic Research Fund Grants (MOE-T2EP20220-0005 \& RT19/22).
}% <-this % stops an unwanted space
% \thanks{Manuscript received April 19, 2005; revised August 26, 2015.}
}

% note the % following the last \IEEEmembership and also \thanks - 
% these prevent an unwanted space from occurring between the last author name
% and the end of the author line. i.e., if you had this:
% 
% \author{....lastname \thanks{...} \thanks{...} }
%                     ^------------^------------^----Do not want these spaces!
%
% a space would be appended to the last name and could cause every name on that
% line to be shifted left slightly. This is one of those "LaTeX things". For
% instance, "\textbf{A} \textbf{B}" will typeset as "A B" not "AB". To get
% "AB" then you have to do: "\textbf{A}\textbf{B}"
% \thanks is no different in this regard, so shield the last } of each \thanks
% that ends a line with a % and do not let a space in before the next \thanks.
% Spaces after \IEEEmembership other than the last one are OK (and needed) as
% you are supposed to have spaces between the names. For what it is worth,
% this is a minor point as most people would not even notice if the said evil
% space somehow managed to creep in.

% The paper headers
\markboth{Journal of \LaTeX\ Class Files,~Vol.~14, No.~8, August~2015}%
{Shell \MakeLowercase{\textit{et al.}}: Bare Demo of IEEEtran.cls for Computer Society Journals}
% The only time the second header will appear is for the odd numbered pages
% after the title page when using the twoside option.
% 
% *** Note that you probably will NOT want to include the author's ***
% *** name in the headers of peer review papers.                   ***
% You can use \ifCLASSOPTIONpeerreview for conditional compilation here if
% you desire.

% The publisher's ID mark at the bottom of the page is less important with
% Computer Society journal papers as those publications place the marks
% outside of the main text columns and, therefore, unlike regular IEEE
% journals, the available text space is not reduced by their presence.
% If you want to put a publisher's ID mark on the page you can do it like
% this:
%\IEEEpubid{0000--0000/00\$00.00~\copyright~2015 IEEE}
% or like this to get the Computer Society new two part style.
%\IEEEpubid{\makebox[\columnwidth]{\hfill 0000--0000/00/\$00.00~\copyright~2015 IEEE}%
%\hspace{\columnsep}\makebox[\columnwidth]{Published by the IEEE Computer Society\hfill}}
% Remember, if you use this you must call \IEEEpubidadjcol in the second
% column for its text to clear the IEEEpubid mark (Computer Society jorunal
% papers don't need this extra clearance.)

% use for special paper notices
%\IEEEspecialpapernotice{(Invited Paper)}

% for Computer Society papers, we must declare the abstract and index terms
% PRIOR to the title within the \IEEEtitleabstractindextext IEEEtran
% command as these need to go into the title area created by \maketitle.
% As a general rule, do not put math, special symbols or citations
% in the abstract or keywords.
\IEEEtitleabstractindextext{%
\begin{abstract}
3D Gaussian Splatting (3D-GS) has emerged as a significant advancement in the field of computer graphics and 3D vision, offering explicit scene representation and novel view synthesis without the reliance on neural networks. This technique has found diverse applications in areas such as robotics, urban mapping, autonomous navigation, and virtual reality/augmented reality, just name a few. Given the growing popularity and expanding research in 3D-GS, this paper presents a comprehensive survey of relevant papers from the past year. We organize the survey into taxonomies based on characteristics and applications, providing an introduction to the theoretical underpinnings of 3D-GS. The survey aims to introduce the theoretical foundations of 3D Gaussian Splatting and provide a reference for new researchers while inspiring future research directions.
\end{abstract}

% Additionally, we present a benchmark comparison of performance and speed among leading 3D Gaussian Splatting models. 

% Note that keywords are not normally used for peerreview papers.
\begin{IEEEkeywords}
3D Gaussian Splatting, differentiable rendering, 3D reconstruction, generation, perception, virtual humans, manipulation.
\end{IEEEkeywords}}

% make the title area
\maketitle

% To allow for easy dual compilation without having to reenter the
% abstract/keywords data, the \IEEEtitleabstractindextext text will
% not be used in maketitle, but will appear (i.e., to be "transported")
% here as \IEEEdisplaynontitleabstractindextext when the compsoc 
% or transmag modes are not selected <OR> if conference mode is selected 
% - because all conference papers position the abstract like regular
% papers do.
\IEEEdisplaynontitleabstractindextext
% \IEEEdisplaynontitleabstractindextext has no effect when using
% compsoc or transmag under a non-conference mode.

% For peer review papers, you can put extra information on the cover
% page as needed:
% \ifCLASSOPTIONpeerreview
% \begin{center} \bfseries EDICS Category: 3-BBND \end{center}
% \fi
%
% For peerreview papers, this IEEEtran command inserts a page break and
% creates the second title. It will be ignored for other modes.
\IEEEpeerreviewmaketitle

% Computer Society journal (but not conference!) papers do something unusual
% with the very first section heading (almost always called "Introduction").
% They place it ABOVE the main text! IEEEtran.cls does not automatically do
% this for you, but you can achieve this effect with the provided
% \IEEEraisesectionheading{} command. Note the need to keep any \label that
% is to refer to the section immediately after \section in the above as
% \IEEEraisesectionheading puts \section within a raised box.

% The very first letter is a 2 line initial drop letter followed
% by the rest of the first word in caps (small caps for compsoc).
% 
% form to use if the first word consists of a single letter:
% \IEEEPARstart{A}{demo} file is ....
% 
% form to use if you need the single drop letter followed by
% normal text (unknown if ever used by the IEEE):
% \IEEEPARstart{A}{}demo file is ....
% 
% Some journals put the first two words in caps:
% \IEEEPARstart{T}{his demo} file is ....
% 
% Here we have the typical use of a "T" for an initial drop letter
% and "HIS" in caps to complete the first word.

%\vspace{-0.2cm}
\IEEEraisesectionheading{\section{Introduction}\label{sec:introduction}}
%\vspace{-0.2cm}

\IEEEPARstart{3}{D} Gaussian Splatting (3D-GS) has emerged as a prominent technique in the field of computer graphics, particularly in the context of 3D rendering~\cite{kerbl20233d,lu2023scaffold,yu2023mip}. 
3D-GS offers a versatile and powerful approach for efficiently rendering complex scenes with high levels of detail~\cite{wu20234d,cotton2024dynamic}.
By representing objects and surfaces as a collection of Gaussians, Gaussian splatting allows for the efficient and accurate representation of geometry and appearance properties~\cite{guedon2023sugar, yu2023mip}. 
3D-GS overcomes the limitations of volume rendering methods by providing a more flexible and adaptive representation of 3D objects~\cite{kerbl20233d}. 
Additionally, Gaussian splatting enables realistic rendering of various visual effects such as depth-of-field and soft shadows, making it a valuable tool in computer graphics research and applications~\cite{chung2023depth}.

\begin{figure}[t]
    \centering
    {%
        \includegraphics[width=0.85\linewidth]{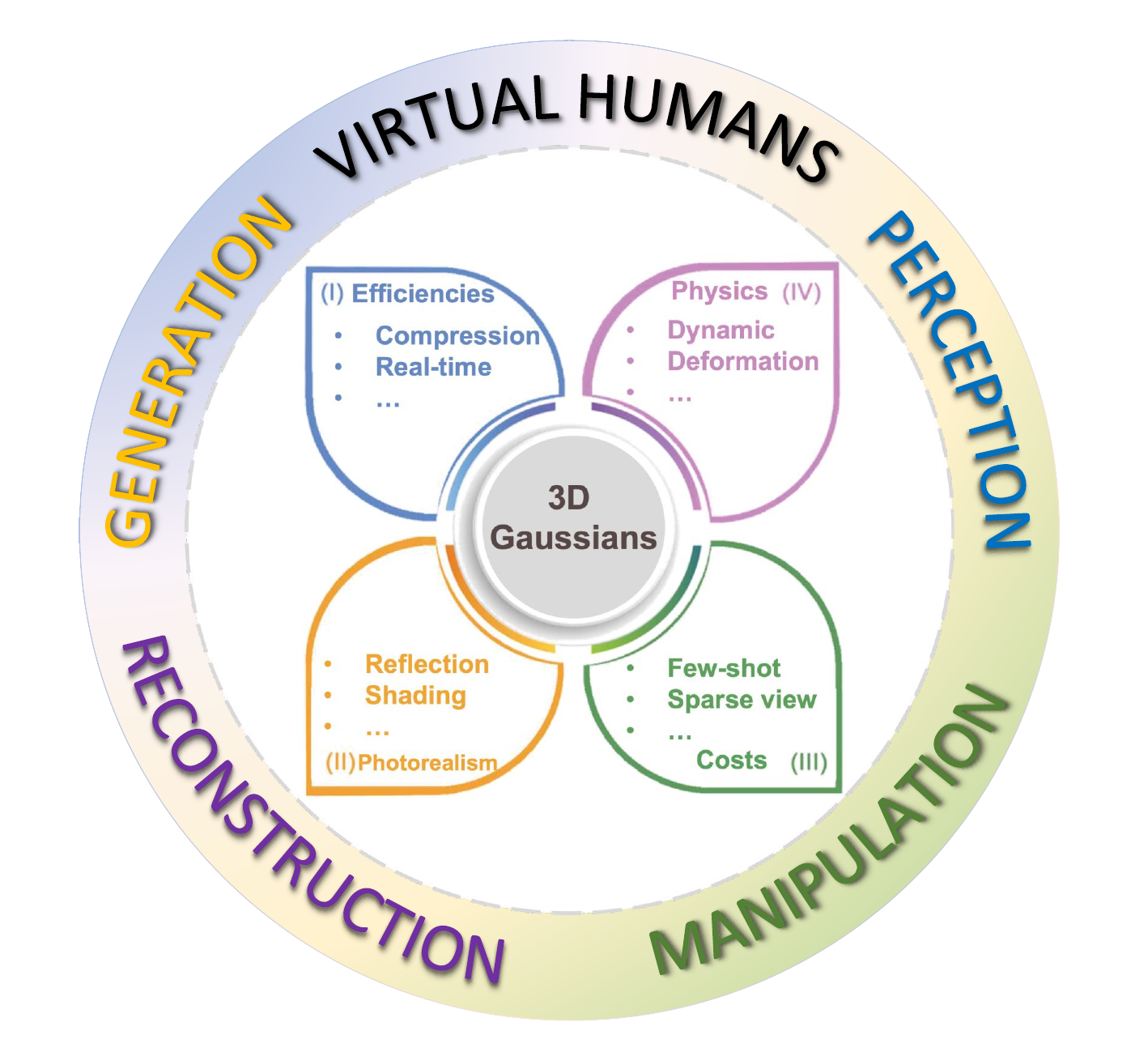}%
        \label{subfig:teaser}%
    }
    % \vspace{-1.2cm}
    {%
        \includegraphics[width=\linewidth]{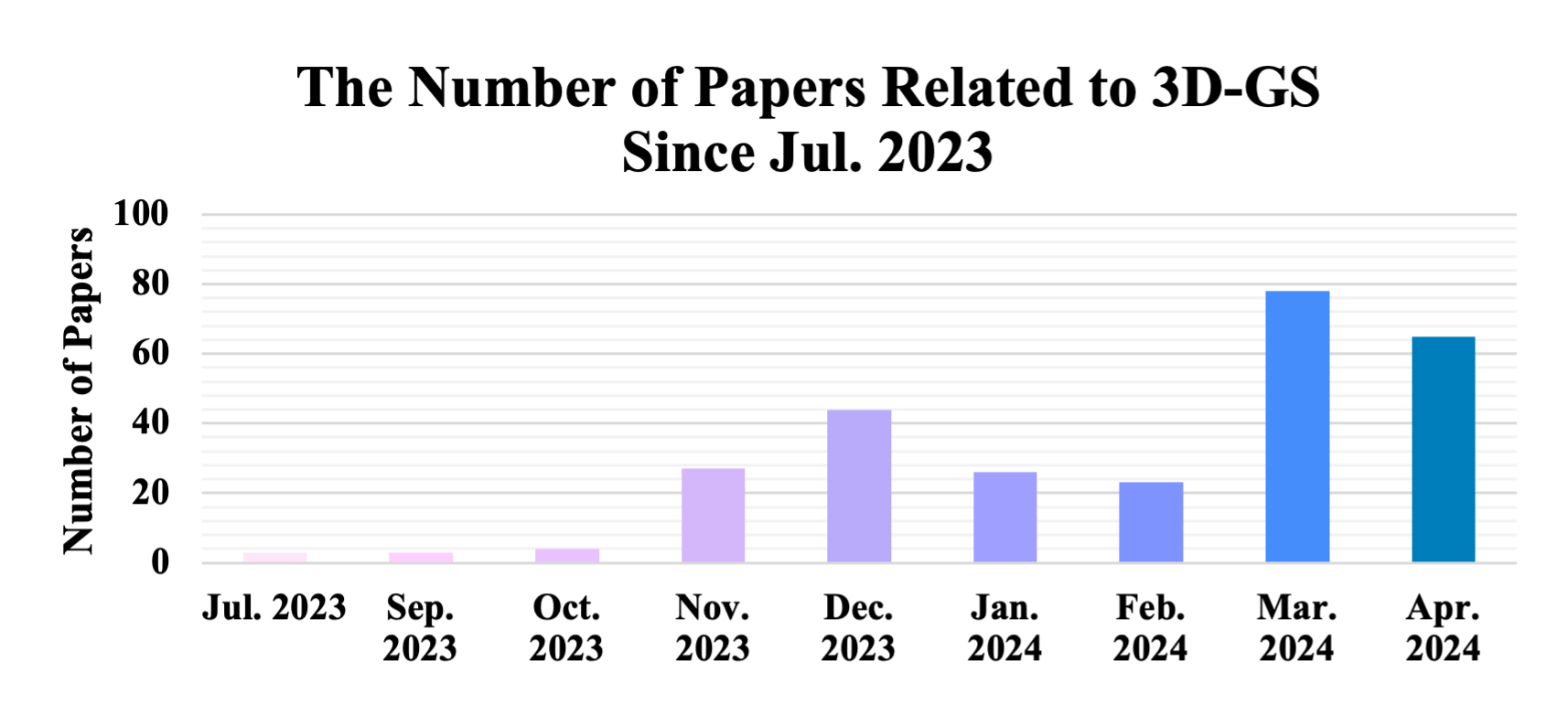}%
        \label{fig:number_paper}%
    }
    % \vspace{-1.0cm}
    \caption{(Top) The structure of this survey. We begin by introducing the optimization of 3D-GS in terms of efficiency, photorealism, costs, and physics. Then, we review 3D-GS on reconstruction, manipulation, perception, generation, and virtual human applications. (Bottom) The monthly number of publications since July 2023.}
    \label{fig:teaser}
    % \vspace{-0.7cm}
\end{figure}

% \begin{figure}
%     \centering
%     \includegraphics[width=\linewidth]{Figures/teaser_update.jpg}
%     \vspace{-0.7cm}
%     \caption{The framework of this survey. The optimization of 3DGS will be first introduced in terms of efficiency, realness, costs, and physics. Then, 3DGS on reconstruction, manipulation, perception, generation, and human applications are comprehensively reviewed.}
%     \label{fig:teaser}
%     \vspace{-0.5cm}
% \end{figure}

\begin{figure*}
    \centering
    \includegraphics[width=\linewidth]{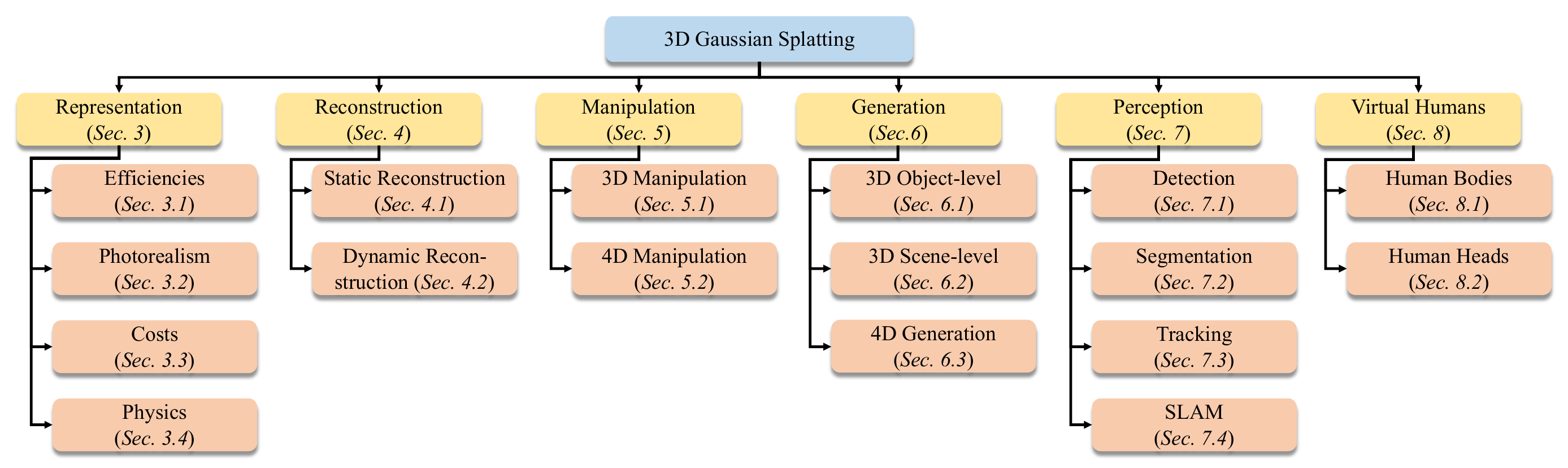}
    % \vspace{-0.9cm}
    
    \caption{Taxonomy of existing 3D Gaussian Splatting derived methods.}
    \label{fig:tax}
    % \vspace{-0.6cm}
\end{figure*}

% Before the era of 3D-GS, Neural Radiance Fields (NeRF) is widely used in the field of computer graphics and 3D scene reconstruction. 
% NeRF enables the generation of visually realistic scenes with a limited number of input views, presenting a novel approach to image synthesis.
% Nevertheless, NeRF faces challenges and limitations, particularly in terms of computational efficiency and controllability.
% Besides, for novel-view synthesis, the introduction of structure-from-motion (SfM) and subsequent advancements in multiview stereo (MVS) algorithms provided a more robust framework for 3D scene reconstruction, paving the way for more sophisticated view synthesis algorithms. 
% NeRF leverages neural networks to map spatial coordinates to color and density to take a step further in novel-view synthesis. 
% The success of NeRF lies in its ability to create continuous, volumetric scene representations, resulting in outputs with unparalleled detail and realism. 
% However, NeRFs are computationally intensive, often requiring extensive training times and substantial resources for high-resolution rendering.
% Therefore, due to its rapid training time and real-time rendering capabilities, 3D-GS has emerged as a popular alternative to Nerf in specific scenes.

The objective of this paper is to provide a comprehensive overview of 3D Gaussian Splatting, exploring its underlying advancements and applications. 
Through an extensive review of existing literature, we aim to present a detailed analysis of the techniques and algorithms used in 3D-GS, including the mechanisms of Gaussian generation, reconstruction, manipulation, perception, and human applications.
In addition to examining the current state-of-the-art in 3D-GS, this paper also aims to highlight the challenges and open research questions in this field. 
We delve into topics such as efficient data structures for handling large-scale scenes, optimization techniques for real-time rendering, and the integration of 3D-GS with other rendering algorithms. 
Furthermore, we discuss various applications of 3D-GS, ranging from robotics to autonomous navigation.
By identifying these research gaps, we hope to inspire future work and foster advancements in the field of 3D Gaussian Splatting.

% %Overall, this comprehensive review  contributes to a deeper understanding of Gaussian splatting and its potential applications in various domains. 
Our main contributions are listed below.

\begin{itemize}
    \item \textbf{Unified framework with systematic taxonomy.} We introduce a unified $\&$ practical framework for categorizing existing works in 3D Gaussians. This framework divides the field into six main aspects. Additionally, we provide detailed taxonomies of applications of 3D Gaussians offering a comprehensive perspective of the field. 
    \item \textbf{Comprehensive and up-to-date review.} Our survey presents an extensive and up-to-date review of 3D-GS, covering both classical and cutting-edge approaches. For each category, we provide fine-grained classification and concise summaries. 
    To the best of our knowledge, our survey presents the first review of 3D Gaussians.
    \item \textbf{Insights into future directions in 3D-GS.} We highlight the technical limitations of current research and propose several promising avenues for future work, aiming to inspire further advancements in this rapidly evolving field. %Special emphasis is given to exploring the potential roles of 3D-GS, offering insights into their future applications.
\end{itemize}
% % \vspace{-0.2cm}

The remaining part of this survey is organized as follows: 
Section~\ref{sec:Background} provides an introduction to the background knowledge of 3D-GS. 
Section~\ref{sec:Representation} presents a systematic review of the methods used to optimize 3D-GS, including considerations such as rendering efficiency, realism of the rendered images, costs, and the physics involved in 3D-GS.
Additionally, Section~\ref{sec:Reconstruction} reviews recently proposed methods for reconstructing meshes, while Section~\ref{sec:Manipulation} compares and summarizes the techniques employed for manipulating 3D-GS.  
Furthermore, Section~\ref{sec:Generation} examines the methods for 3D generation from the perspectives of both object- and scene-level. 
Sections~\ref{sec:Perception} and ~\ref{sec:Human} summarize the applications of 3D-GS in perception and human body, respectively. 
Finally, Section~\ref{sec:Discussion} discusses several promising future directions for 3D-GS.

%\vspace{-0.5cm}
\section{Background}\label{sec:Background}
%\vspace{-0.1cm}

\textbf{Traditional Point-based Rendering.}
The objective of point-based rendering techniques is to generate realistic images by rendering a set of distinct geometric primitives.
Zwicker et al.~\cite{zwicker2001surface} proposed a technique to render ellipsoid-shaped splats, enabling them to occupy multiple pixels. This method enhances the production of hole-free images by leveraging the mutual overlap between the splats, which proves more efficient than employing a solely point-based representation.
Kopanas et al.~\cite{kopanas2021point} presented a novel differentiable point-based pipeline that integrates bi-directional Elliptical Weighted Average splatting, a probabilistic depth test, and efficient camera selection as its fundamental elements.
In recent years, a series of techniques have been developed to enhance the rendering process. These advancements include the introduction of a texture filter specifically designed for anti-aliasing rendering~\cite{zwicker2001ewa}, improvements in rendering efficiency~\cite{botsch2003high}, and the resolution of discontinuous shading issues~\cite{rusinkiewicz2000qsplat}.

Traditional point-based rendering methods primarily focus on generating high-quality rendered outputs using a specified geometry. However, the advent of new implicit representation techniques has spurred researchers to explore point-based rendering with neural implicit representation. This approach does not depend on a predefined geometry for the purpose of 3D reconstruction.
An exemplary work in this field is Neural Radiance Fields (NeRF)~\cite{mildenhall2021nerf}, which utilizes an implicit density field to model geometry and employs an appearance field to predict view-dependent color $c_i$. Point-based rendering is employed to combine the colors of all sample points along a camera ray, resulting in the generation of a pixel color $C$:
\begin{equation}
    C=\sum_{i=1}^N c_i \alpha_i T_i
\end{equation}
where $N$ represents the number of sample points along a ray. The view-dependent color and opacity value for the $i$-th point on the ray are denoted as $\alpha_i=\exp \left(-\sum_{j=1}^{i-1} \sigma_j \delta_j\right)$. Here, $\sigma_j$ refers to the density value of the $j$-th point. The accumulated transmittance is given by $T_i=\prod_{j=1}^{i-1}\left(1-\alpha_j\right)$.

The rendering process of 3D-GS exhibits similarities to NeRF, yet there exist two fundamental distinctions between them. Firstly, 3D-GS directly models opacity values, whereas NeRF converts density values into opacity values. Secondly, 3D-GS employs rasterization-based rendering, eliminating the need for sampling points, whereas NeRF necessitates dense sampling throughout the 3D space.

\textbf{Gaussian Splatting Radiance Fields.}
\label{sec:bg-3dgs}
Gaussian Splatting Radiance Field~\cite{kerbl20233d} also referred to as 3D Gaussian Splatting, is an explicit radiance field-based scene representation that represents a radiance field using a large number of 3D anisotropic balls, each modeled using a 3D Gaussian distribution (Eq.~\ref{eq:bg1}). More concretely, each anisotropic ball has mean $\mathcal{M} \in \mathbb{R}^3$, covariance $\Sigma$, opacity $\alpha \in \mathbb{R}$ and spherical harmonics parameters $\mathcal{C} \in \mathbb{R}^k$ ( $k$ is the degrees of freedom) for modeling view-dependent color. For regularizing optimization, the covariance matrix is further decomposed into rotation matrix $\mathbf{R}$ and scaling matrix $\mathbf{S}$ by Eq~\ref{eq:bg2}. These matrices are further represented as quaternions $r \in \mathbb{R}^4$ and scaling factor $s \in \mathbb{R}^3$.
% \vspace{-0.1cm}
\begin{equation}
    G(X)=e^{-\frac{1}{2} \mathcal{M}^T \Sigma^{-1} \mathcal{M}}
    \label{eq:bg1}
\end{equation}
\begin{equation}
    \Sigma=\mathbf{R S S}^T \mathbf{R}^T
    \label{eq:bg2}
\end{equation}
% \vspace{-0.1cm}
For this scene representation, view rendering is performed via point splatting~\cite{yifan2019differentiable}.
Specifically, all Gaussian balls in the scene are first projected onto the $2 \mathrm{D}$ image plane, and their color is computed from spherical harmonic parameters. Then, for every $16 \times 16$ pixel patch of the final image, the projected Gaussians that intersect with the patch are sorted by depth. For every pixel in the patch, its color is computed by alpha compositing the opacity and color of all the Gaussians covering this pixel by depth order, as in Eq.~\ref{eq:bg3}.
% \vspace{-0.1cm}
\begin{equation}
    C=\sum_{i \in N_{\text {cov }}} c_i \alpha_i \prod_{j=1}^{i-1}\left(1-\alpha_j\right)
    \label{eq:bg3}
\end{equation}
% \vspace{-0.1cm}
where, $N_{\text{cov}}$ represents the splats that cover this pixel, $\alpha_i$ represents the opacity of this Gaussian splat multiplied by the density of the projected 2D Gaussian distribution at the location of the pixel, and $c_i$ represents the computed color.

\textbf{Datasets.}\label{sec:Datasets}
Various publicly available datasets are utilized to evaluate the performance of 3D-GS on various tasks. 
Table~\ref{dataset} provides an overview of some of these datasets for 3D-GS in optimization, reconstruction, manipulation, generation, perception, and virtual humans. 
% These datasets have different properties, which are summarized in the table. 

\begin{table*}[htbp]
\centering
\caption{List of commonly used datasets for 3D Gaussians.}
% \vspace{-0.3cm}
\resizebox{\textwidth}{!}
{%
\begin{tabular}{lccccccc}
\toprule[1.5pt]
\rowcolor{C7!50}Dataset        & Year & Samples        & Camera Views & Type              & Description   & Task    & \#Number                              \\ \midrule[1pt]
 
\rowcolor{C6!50}  Mip-NeRF 360~\cite{barron2022mip}  &CVPR'22  & 9 & 360 viewpoint     & Videos & \begin{tabular}[c]{@{}c@{}}Outdoor and indoor scene containing a complex \\central object or area with a detailed background  \end{tabular} & Representation \& Reconstruction  & 49 \\

\rowcolor{C6!50}  Tanks and Temples~\cite{knapitsch2017tanks} & ToG'17 & 14 &   263-1107   & Videos & \begin{tabular}[c]{@{}c@{}}Ground-truth captured by high quality \\industrial laser scanner\end{tabular} & Representation \& Reconstruction & 34 \\

\rowcolor{C6!50}  DeepBlending~\cite{hedman2018deep}  &TOG'18  & 19 &  12-418    & Images & \begin{tabular}[c]{@{}c@{}} Images with suicient variety of scene content \end{tabular} & Representation \& Reconstruction & 18 \\

\rowcolor{C6!50}  Plenoptic Video~\cite{li2022neural} & CVPR'22 & 6 &   21   & Videos & \begin{tabular}[c]{@{}c@{}} Captured daily events with challenging scene motions\end{tabular} & Representation \& Reconstruction & 3 \\

\rowcolor{C6!50}  D-Nerf~\cite{pumarola2021d}    & CVPR'21 & 8 & 360 viewpoint     &  Synthetic videos &   \begin{tabular}[c]{@{}c@{}} Animated objects with complex geometries \\  and non-Lambertian materials \end{tabular}  & Representation \& Manipulation  &  9 \\

\rowcolor{C6!50}    DyNeRF~\cite{li2022neural}  & CVPR'22  & 6  & 20   & Videos &  \begin{tabular}[c]{@{}c@{}} Time-synchronized and calibrated multi-view \\  videos that covers challenging 4D scenes \end{tabular}  & Representation & 3 \\

\rowcolor{C6!50}    NeRF-LLFF~\cite{mildenhall2019llff}  &  SIGGRAPH'19 & 8 & forward-facing views & images &  A subset of Real Iconic  & Representation & 11 \\

\rowcolor{C5!50}  RealEstate10k~\cite{zhou2018stereo}  &ToG'18 & about 80k &  -   & Videos & \begin{tabular}[c]{@{}c@{}}Video clips on YouTube shot from a moving camera\end{tabular} & Reconstruction  & 3 \\

\rowcolor{C5!50}  ACID~\cite{liu2021infinite}  &ICCV'21 & 765  &  -   & Videos & \begin{tabular}[c]{@{}c@{}} A dataset of aerial landscape videos\end{tabular} & Reconstruction  & 2 \\

\rowcolor{C5!50}  ShapeNet~\cite{chang2015shapenet}  &2015 & about 60k  &  50   & Synthetic images & \begin{tabular}[c]{@{}c@{}} A richly-annotated, large-scale dataset of 3D shapes\end{tabular} & Reconstruction & 1 \\

\rowcolor{C4!50}  Tensor4D~\cite{shao2023tensor4d}    & CVPR'23 & 4 & 1, 4, 12     & Videos &   \begin{tabular}[c]{@{}c@{}}Dynamic half-body human videos by four\\ sparsely positioned, fixed RGB cameras\end{tabular}   & Manipulation & 5 \\

\rowcolor{C4!50}  NeRF-DS~\cite{pumarola2021d}    & CVPR'23 & 7 & 2 forward-facing views  & Videos &   Dynamic specular dataset & Manipulation  & 2 \\

\rowcolor{C4!50}  CoNeRF~\cite{kania2022conerf}    & CVPR'22 & 7 &  -  & Videos &   Real face scenes   & Manipulation & 1 \\

\rowcolor{C4!50}  InterHand2.6M~\cite{moon2020interhand2}    & ECCV'20 & 36 &  80-140 & Videos &   \begin{tabular}[c]{@{}c@{}} 2.6M labeled single and \\ interacting hand frames \end{tabular}  & Manipulation  & 1 \\

\rowcolor{C3!50}   NVOS~\cite{ren2022neural} & ICCV'22 & 7 &   20-62   &  Images &  \begin{tabular}[c]{@{}c@{}}  Real-world front-facing scenes with annotated object masks based on LLFF\end{tabular}  & Perception  & 1 \\

\rowcolor{C3!50}    SPIn-NeRF~\cite{mirzaei2023spin} & CVPR'23 & 10 & 100   & Images &  \begin{tabular}[c]{@{}c@{}} Real-world forward-facing
scenes with human annotated object masks \end{tabular} & Perception & 2 \\

\rowcolor{C3!50}    LERF~\cite{kerr2023lerf} & ICCV'23 & 13 & -   & Videos &  \begin{tabular}[c]{@{}c@{}} A mixture of in-the-wild and posed long-tail scenes\end{tabular}  & Perception & 5 \\

\rowcolor{C3!50}    Replica~\cite{straub2019replica}  &arXiv'19  & 18 &  -   &  Videos & \begin{tabular}[c]{@{}c@{}}  Highly photo-realistic 3D indoor scene dataset at room and building scale\end{tabular} & Perception & 14 \\

\rowcolor{C3!50}    KITTI~\cite{geiger2012we}  &CVPR'12  & 22 &  -   &  Videos & \begin{tabular}[c]{@{}c@{}}   A city-scale dataset created for autonomous driving research\end{tabular} & Perception & 5 \\

\rowcolor{C2!50}    Objaverse 1.0~\cite{deitke2023objaverse} & CVPR'23 & 818k &   -   & 3D objects &  \begin{tabular}[c]{@{}c@{}} A large dataset of over 800K 3D models with descriptive captions, tags, and animations.\end{tabular} & Generation & 2 \\

\rowcolor{C2!50}    OmniObject3D~\cite{wu2023omniobject3d} & CVPR'23 & 6k &   -   & 3D objects &  \begin{tabular}[c]{@{}c@{}} Real-scanned 3D object captured with both 2D and 3D sensors, providing textured meshes, \\point clouds, multi-view rendered images, and multiple real-captured videos.\end{tabular}  & \begin{tabular}[c]{@{}c@{}} Generation \&  Perception \\ \& Reconstruction \end{tabular}  & 4 \\

\rowcolor{C1!50}     People-Snapshot~\cite{alldieck2018video} & CVPR'18 & 24 &    1  & Videos &  \begin{tabular}[c]{@{}c@{}} Containing subjects captured with varying sets of garments \\ and with three different background scenes \end{tabular} & Virtual Humans & 7 \\

\rowcolor{C1!50}     DynaCap~\cite{habermann2021real} & ToG'21 & 5 &    50-101  & Videos &  \begin{tabular}[c]{@{}c@{}} Recording 5 subjects wearing different types of apparel \\ and performing a wide range of motions like "dancing"\end{tabular} & Virtual Humans & 1 \\

\rowcolor{C1!50}     ZJU-Mocap~\cite{peng2021neural} & CVPR'21 & 9 &    21  & Videos &  9 dynamic
human videos with 60 to 300 frames & Virtual Humans  & 9 \\

\rowcolor{C1!50}     THuman4~\cite{zheng2022structured} & CVPR'22 & 3 &    24  & Videos &  multi-view
human videos ranging from 2500 to 5000 frames & Virtual Humans & 3 \\

\rowcolor{C1!50}     NeuMan~\cite{jiang2022neuman} & ECCV'22 & 6 &    1  & Videos &  \begin{tabular}[c]{@{}c@{}} Monocular videos featuring a single individual \\ captured using a mobile phone \end{tabular}  & Virtual Humans & 1 \\

\rowcolor{C1!50}     ActorsHQ~\cite{icsik2023humanrf} & arXiv'23 & 16 &    160  & Videos &  \begin{tabular}[c]{@{}c@{}} Containing 8 actors with casual clothing such as skirts or shorts\end{tabular}  & Virtual Humans & 2
\\ \bottomrule[1.5pt]
\end{tabular}%
}\label{dataset}
% \vspace{-0.5cm}
\end{table*}

% \input{SectionsV1/Optimization}
%\vspace{-0.6cm}
\section{Representation}\label{sec:Representation}
%\vspace{-0.1cm}

\begin{figure}[htbp]
    \centering
    \includegraphics[width=\linewidth]{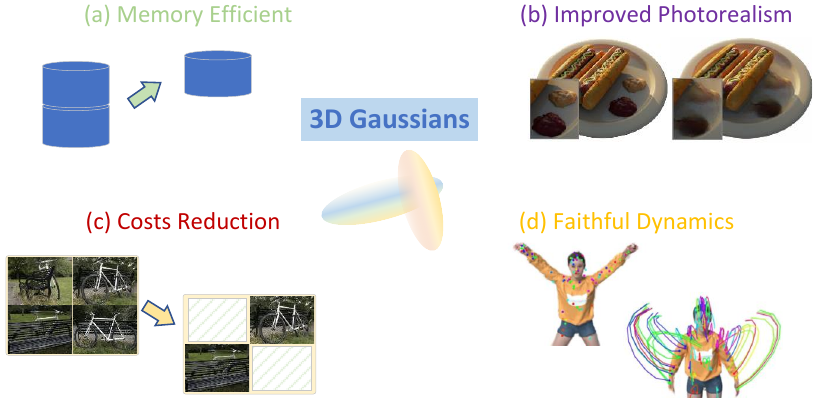}
    %\vspace{-0.8cm}

    \caption{An illustration of refining the representation of 3D-GS: (a) efficiencies, (b) photorealism, (c) costs, and (d) physics. Images courtesy of ~\cite{jiang2023gaussianshader,kratimenos2023dynmf,zhu2023fsgs}.}
    \label{fig:opt}
    %\vspace{-0.2cm}
\end{figure}

Despite the already showcased capability and efficiency of 3D-GS, there is still room for further improvements, including (as depicted in Fig~\ref{fig:opt}): (a) Making 3D-GS more memory efficient for real-time rendering; (b) Enhancing the quality of rendered images; (c) Reducing the costs for synthesizing novel views; (d) Enabling 3D Gaussians to represent dynamic scenes with faithful dynamics.

%\vspace{-0.5cm}
\subsection{Efficiencies}
%\vspace{-0.1cm}
The parameters within millions of Gaussians that represent the scene require an enormous memory space, hence the reduction of memory usage while maintaining quality is critical and beneficial for real-time rendering.

% Add ~\cite(), total 5 + 2

% First grid gaussian method
% Scaffold-GS
Encouraged by the grid-guided NeRF, Lu et al.~\cite{lu2023scaffold} proposed a memory-efficient Scaffold-GS, which maintains rendering quality and speed. 
Scaffold-GS exploits underlying scene structure to help prune excessively expanded Gaussian balls. 
It utilizes initialized points from Structure-from-motion (SfM)~\cite{schonberger2016structure} to construct a sparse grid of anchor points, to each of which a set of learnable Gaussians is attached.
Attributes of these Gaussians are predicted on the fly according to specific anchor features. 
Furthermore, a policy guided by the aggregated gradients of the neural Gaussians is adopted to grow anchor points wherever a pruning operation eliminates significant and trivial anchors.
% An additional volume regularization loss term~\cite{lombardi2021mixture} is added to encourage the Gaussians to be small with minimal overlapping.

% Compact3D and EAGLES all use quantization, so they can be described together.
% Compact3D
Vector quantization (VQ) ~\cite{equitz1989new} is widely applied for data compression and encoding. It first categorizes data into clusters of similar vectors and then represents them via a compact codebook. VQ facilitates efficient storage and transmission of data.
Therefore, Navaneet et al.~\cite{navaneet2023compact3d} proposed CompGS that adopts VQ to compress the parameters and reduce the memory usage of 3D-GS. 
Specifically, K-means is performed on the color, spherical harmonics, scale, and rotation vectors separately to obtain relative clusters for efficient quantization. 
The opacity and position parameters are left untouched since the former is a plain scalar and the quantization of the latter will cause overlapping of Gaussians. 
%Furthermore, the indices of codebooks are represented in fewer bits, and the Gaussians are sorted based on their color so that only one index is stored for the first Gaussian of one color and the rest with identical color simply subsequently adds 1.
%to improve memory usage efficiency for 3D-GS.   the attribute vectors of EAGLES
The indices of codebooks are represented in fewer bits to improve efficiency.
% EAGLES
Similarly, Girish et al.\cite{girish2023eagles} introduced EAGLES to utilize VQ for 3D-GS.
Specifically, except for the coefficient of base band color spherical harmonics and scaling, and position vectors, all other attributes are quantized and encoded. 
The continuous approximations are maintained for the back-propagation of gradients. 
%Attributes are then retrieved from the latent quantized vectors via an MLP decoder.
%that gradually increases the rendering size. This leads
Moreover, EAGLES utilizes a coarse-to-fine training strategy to reduce training time and improve rendering quality.
Lee et al.~\cite{lee2023compact} proposed a compact 3D Gaussian representation drastically reducing required storage space. It applies a learnable volume mask to filter out insignificant Gaussians. Then the geometric attributes of selected Gaussians are compressed via residual VQ~\cite{zeghidour2021soundstream} for memory efficiency.

%LightGauss
On the other hand, the lossless octree-based algorithm is also utilized for efficient representation. For instance, Fan et al.~\cite{fan2023lightgaussian} proposed LightGaussian to compress Gaussians for memory efficiency.
Gaussians are first pruned according to their global significance.
Then, the degree of spherical harmonics coefficients are reduced via data distillation and coefficients of trivial Gaussians are quantized. 
%Pseudo-views are randomly sampled to facilitate the knowledge transfer. 
%Furthermore, the position parameter is compressed via a lossless octree-based algorithm and the remaining attributes are saved in half-precision format for further memory saving.
Furthermore, the position parameter is compressed via an octree-based algorithm and the remaining attributes are saved in half-precision format.
Ren et al.~\cite{ren2024octree} introduced Octree-GS which built upon Scaffold-GS and leverages octree-based Level-of-Detail (LOD) representation. The octree is constructed for anchoring initialized Gaussians and each anchor is adaptively refined between adjacent levels of LOD.

% Efficient 3d rep
Moreover, Katsumata et al.~\cite{katsumata2023efficient} proposed to relieve the memory required of 3D-GS for dynamic scene representation. 
It divides the parameters of Gaussians into time-invariant and time-varying ones. 
% The former contains the position and rotation which are estimated via Fourier and linear approximation respectively. 
%This approach effectively reduces memory consumption when compared to storing each of those parameters in every time step. 
%Furthermore, the flow information is employed via a loss term to overcome the ambiguity between consecutive frames across different time steps.
This approach effectively avoids storing parameters in every time step.
Furthermore, the flow information is employed to reduce the ambiguity between consecutive frames.

\textbf{Challenges:}
Representing scenes with intricate details requires an enormous amount of 3D Gaussians. The tremendous storage space needed for Gaussians not only impedes its application on edge devices but also restrains the rendering speed. On the other hand, excessive insignificant Gaussians could be generated during the reconstruction process. This will not only increase unnecessary memory usage but also degrades rendering quality.

% of insignificant Gaussian 
\textbf{Opportunities:}
Existing vector quantization and pruning methods have demonstrated their effectiveness in compressing 3D-GS for static scenes. However, (i) additional procedures prolong the training time. (ii) There is still room to find the minimum set of Gaussians that could maximize the capability for scene representation. (iii) Extending existing static methods to dynamic scenarios and improving the compactness of dynamic scene representation are still under-explored.

%\vspace{-0.3cm}
\subsection{Photorealism}
% \vspace{-0.1cm}
The aliasing issue and artifacts emerge during the splatting process and resolving them is clearly favorable for the quality and realism of rendered image. 
Besides, the realism of reflections within the scene could be further improved.

% ~\cite(), Total 6 + 1

% Multi-scale AA
Yan et al.~\cite{yan2023multi} introduced a multi-scale approach for mitigating the aliasing effect in 3D-GS. 
They postulate that such an issue is primarily caused by the splatting of large amounts of Gaussions filled in regions with intricate 3D details. 
Hence, they propose to represent the scene at different levels of detail. 
For each level, small and fine-grained Gaussians below the size threshold in each voxel are aggregated into larger Gaussians and then inserted into subsequent coarser levels. 
These multi-scale Gaussians effectively encode both high- and low-frequency signals and are trained with original images and their downsampled counterparts. 
During rendering, Gaussians with appropriate scale are chosen accordingly, thereby improving quality and increasing rendering speed.
% FreGS, Zhang et al.~\cite(zhang2024fregs)
Similarly, Zhang et al.~\cite{zhang2024fregs} proposed FreGS to mitigate the over-reconstruction issue. It utilizes amplitude and phase discrepancies described within Fourier space for regularization. Then, a frequency annealing strategy is adopted for facilitating densification in a coarse-to-fine manner and progressively learns low-to-high frequency components.

% MipSplatting

% Mip-NeRF~\cite{barron2022mip} is proven as an efficient and robust method in the field of NeRF.
Mip-NeRF~\cite{barron2022mip} has been established as an efficient and robust approach within the NeRF domain.
Encouraged by Mip-NeRF, Yu et al.~\cite{yu2023mip} introduced 3D smoothing and 2D Mip filters to resolve ambiguities within the optimization process of 3D Gaussians.
Derived from the Nyquist-Shannon Sampling Theorem~\cite{nyquist1928certain,shannon1949communication}, the 3D filter performs as a Gaussian low-pass filter that restrains the frequency of 3D representation within half of the maximal sampling rate, which derived from multi-view images, to remove high-frequency artifacts. 
On the other hand, 2D filter aims to mitigate the aliasing issue when rendering reconstructed scenes at a lower sampling rate. 
It replaces the screen space dilation filter of 3D-GS and replicates the behavior of a box filter from the physical imaging process~\cite{mildenhall2022nerf,szeliski2022computer}. 
% This principled approach achieves better generalization to out-of-distribution scenarios with unseen camera poses and zoom factors.

Except for the improvement of rendering details, lighting decomposition is also an urgent challenge to be solved since the original 3D-GS works not very well on specific materials.
% Relightable 3DGS
Gao et al.~\cite{gao2023relightable} introduced a framework for photo-realistic rendering. 
It utilizes a set of relightable 3D Gaussian points to represent the scenes.
%The surface normal is regularized by the consistency between the rendered normal and the pseudo normal which is computed from the rendered depth map.
The surface normal is regularized via the consistency between the rendered normal and the pseudo normal computed from the rendered depth map.
%Geometry clues are introduced via the integration of Multi-View Stereo cues~\cite{yao2018mvsnet}. 
%This approach adopts a simplified Disney BRDF model~\cite{yao2022neilf} and assigns extra properties to each Gaussian for rendering. 
The incident light is divided into local and global components that are represented by spherical harmonics of each Gaussian and shared global spherical harmonics multiplied by a visibility term respectively. 
%In order to improve the rendering efficiency and quality, the physical-based rendering color is computed at Gaussian level and additional regularization terms are attached to the optimization process.
Moreover, a technique based on the Bounding Volume Hierarchy~\cite{karras2012maximizing} is introduced to further use ray tracing to compute visibility for shadows.
% GaussianShader
Jiang et al.~\cite{jiang2023gaussianshader} proposed GaussianShader to further enhance the realism of scenes with specular features and reflective surfaces. 
GaussianShader explicitly considers the light-surface interaction and incorporates a simplified approximation of the rendering equation~\cite{kajiya1986rendering} for high-quality rendering with considerably lower time cost. 
To accurately predict normals on discrete 3D Gaussians, the shortest axis is selected as the approximated normal and two additional trainable normal residuals are introduced for regularization, one for outward and the other for inward axis scenarios.
%Furthermore, consistency of normal-geometry is enforced by minimizing the difference between the grad normals derived from rendered depth maps and rendered normals maps using previously predicted normals.
% GS-IR
Liang et al.~\cite{liang2023gs} presented GS-IR that introduces 3D-GS to inverse rendering. 
GS-IR first generates depth and improves it by considering the depth as a linear interpolation of the distances from 3D Gaussians to the image plane. 
Then the pseudo normals are derived by leveraging the calculated depth and are tied to the depth gradient normal via $L_1$ regularization. 
The indirect illumination is modeled by a spherical harmonics-based architecture and the occlusion information is pre-computed and cached. 
Finally, GS-IR adopts differentiable splatting along with a physic-based rendering pipeline, and split-sum approximation~\cite{karis2013real} to mitigate the intractable integrals.
% SpecNeRF  NeRF + 3DGauss Encodeing
Ma et al.~\cite{ma2023specnerf} proposed SpecNeRF to improve the modeling of specular reflections especially for near-field lighting conditions. Different from aforementioned methods that ameliorate reflection, SpecNeRF aims to enhance the ability of NeRF by incorporating 3D-GS as a directional encoding. It utilizes a set of learnable Gaussians as the basis to embed a 5D ray space that includes the ray origin and ray direction. Therefore, the encoding function could vary spatially and the change of spatial features aligns with the behavior of specular components.
% This results in a better modeling of reflections and improves the realness. 
% SpecNeRF also introduced an initialization stage which involves refinement of Gaussian Parameters to facilitate the joint optimization of the Gaussians and NeRF. 
%Additionally, the monocular normals estimated via a prior model~\cite{eftekhar2021omnidata} are utilized in the early training stage to provide a supervised signal for the predicted normals and mitigate the shape-radiance ambiguity.

\textbf{Challenges:}
%Although the projection of 3D Gaussians onto 2D image drastically accelerates the rendering process, it complicates the calculation of occlusion which leads to poor estimation of illumination. In the meantime, the under-regularized 3D-GS fails to capture precise geometry and cannot natively generate accurate normals. Furthermore, the aliasing issue and artifacts deteriorate the quality of rendered images especially when synthesizing for unseen camera views.
(i) The projection of 3D Gaussians onto 2D image drastically accelerates the rendering process, but it complicates the calculation of occlusion which leads to poor estimation of illumination. 
(ii) 3D-GS is under-regularized for capturing precise geometry information and generating accurate normals. 
(iii) Over-constructed Gaussians could cause aliasing and artifacts that deteriorate the quality of rendered images. 
(iv) 3D-GS falls short in scenes with mirror-like objects and intricate reflections. 
(v) Reconstruct scenes from images with imprecise camera pose or motion blur poses new challenges to 3D-GS.

\textbf{Opportunities:}
%View-dependent variations are critical for scenes featuring specular objects and complex reflections. Hence, empowering 3D-GS to capture significant appearance properties is beneficial for enhancing the realness of rendering. To better reduce the aliasing effects, it is worth investigating methods for more effective elimination of redundant Gaussians without compromising its expressive capability. Besides, the lack of rigorous normal estimation and geometry regularization which impedes the amelioration of image quality could be further compensated.
(i) View-dependent variations are critical for scenes featuring specular objects and complex reflections. Hence, empowering 3D-GS to capture significant appearance properties is beneficial for enhancing the realism of rendering. 
(ii) It is essential to filter out and prune excessively constructed Gaussians, which cause artifacts and floater issues, without compromising the expressive capability and rendering quality. 
(iii) Although the lack of rigorous normal estimation regularization which impedes the amelioration of image quality could be compensated by off-the-shell monocular normal estimator, the introduced intrinsic error should be further addressed.

%\vspace{-0.4cm}
\subsection{Costs}
To synthesize novel views with high quality, a substantial number of images is necessary. Loosening this limitation is desirable for further exploring the potential of 3D-GS.

% Total 3 + 1
% DNGaussian, Li et al.~\cite{li2024dngaussian} cvpr24
% ..Touch-GS, Swann et al.~\cite(swann2024touch)
% ..FDGaussian, Feng et al.~\cite(feng2024fdgaussian)
% ..CoherentGS, Paliwal et al.~\cite(paliwal2024coherentgs)

Several works have been proposed to solve the few-shot problem in 3D-GS.
% DepthReg
Chung et al.~\cite{chung2023depth} introduced a depth-regularized approach to avoid overfitting in few-shot image synthesis. 
Geometry constraints are introduced by exploiting both sparse and dense depth maps obtained from COLMAP~\cite{schonberger2016structure} and a monocular depth estimation model respectively. 
To prevent overfitting, this approach adopts unsupervised constraint for geometry smoothness~\cite{godard2017unsupervised} and utilizes Canny edge detector~\cite{canny1986computational} to avoid regularization in edge areas where depth varies significantly.
%Furthermore, the maximum degree of spherical harmonics is restricted to 1 and an early stop optimization strategy is used to further reduce the overfitting.
% FSGS
Zhu et al.~\cite{zhu2023fsgs} proposed FSGS for novel view synthesis under limited observations. 
Inspired by vertex-adding strategy~\cite{zorin1996interpolating}, it gradually grows the initialized 3D Gaussians that are derived from SfM~\cite{schonberger2016structure} to fill the scene.
The sparse Gaussians are utilized to construct a directed proximity graph that connects the nearest K neighbors and the proximity score of each Gaussian is subsequently calculated from the average distance to its neighbors. 
FSGS then proposes proximity-guided Gaussian unpooling to densify Gaussians and exploit Pearson correlation loss to provide a soft constraint.
%between the rendered depth map and the depth map estimated via a well-trained monocular depth estimator. 
Additionally, FSGS employs pseudo-view augmentation through a 2D prior model to mitigate the issue of overfitting to sparse training views.
% SparseGS
Xiong et al.~\cite{xiong2023sparsegs} introduced SparseGS for few-shot view synthesis. 
Similar to SparseNeuS~\cite{long2022sparseneus}, it adopts a depth map for regularization and proposes to apply an additional pre-trained depth estimation model~\cite{lasinger2019towards,miangoleh2021boosting} for computing a patch-based Pearson correlation between the two depth loss terms.
The absence of images for uncovered regions is compensated by utilizing a generative diffusion model and 3D-GS is guided via Score Distillation Sampling~\cite{poole2022dreamfusion}. 
Furthermore, SparseGS devises a pruning strategy to identify and remove the parts where floating artifacts exist or background collapse.
%identifies the parts of a rendered image that suffer from floating artifacts or background collapse and removes them to improve the quality of the synthesized image.
Li et al.~\cite{li2024dngaussian} introduced DNGaussian to exploit depth information to enhance sparse-view reconstruction. Specifically, it proposes hard and soft depth regularization strategies to enforce the nearest Gaussians to compose complete surfaces. Besides, the scaling and rotation parameters are frozen to reduce their negative impact on color reconstruction. DNGaussian also normalizes both local and global depths to correct small local depth errors and be aware of global scale losses respectively.

\textbf{Challenges:}
%The performance of 3D-GS is heavily relies on the quantity and accuracy of initialized sparse points. This default initialization method naturally contradicts the aim of image cost reduction and makes it grueling to achieve. Besides, inadequate initialization could cause overfitting and yield over-smoothed results.
The performance of 3D-GS heavily relies on the quantity and accuracy of initial sparse points. This default initialization method  contradicts the aim of image cost reduction and makes it grueling to achieve.
The sparse input views could not only cause insufficient initialization of Gaussians, which may not only lead to collapsed reconstruction but also increase the chance of over-fitting and yielding over-smoothed results.

\textbf{Opportunities:}
%Employing an additional monocular depth estimation model could provide useful geometry priors to adjust 3D Gaussians for effective coverage of the scene. However, this strong dependency on the estimation accuracy could lead to poor reconstruction of scenes with complex surfaces where the model fails to output accurate prediction. It is promising to further explore methods for effective densification and adjustment of 3D Gaussians and sufficient utilization of geometry information to improve the rendering quality.
(i) Employing an additional monocular depth estimation model provides useful geometry priors to adjust 3D Gaussians for more effective coverage of unseen views. However, this strong dependency on the estimation accuracy could lead to poor reconstruction of scenes with complex surfaces where the model fails to output accurate prediction. 
(ii) Compensating absent views via pre-trained generative models could mitigate the inadequate initialization issue. But the misaligned consistency between generated views and sparse inputs can lead to distortion in reconstructed scenes especially when dealing with scenes comprising complex details and multiple objects. 
(iii) Exploring more diverse visual cues is promising to facilitate the reduction of image costs.

%\vspace{-0.3cm}
\subsection{Physics}
It is beneficial to enhance the ability of 3D Gaussians Splatting by extending it from representing static scenes to 4D scenarios that could incorporate faithful dynamics that align with real-world physics.

% Total 5 + 2
% 3DGStream, Sun et al.~\cite(sun20243dgstream) cvpr24
% GaGS, Lu et al.~\cite(lu20243d)  cvpr24

% 4D Gaussian deformation
In dynamic scenes, learning deformation is more convenient than modeling scenes at every time step.
Wu et al.~\cite{wu20234d} proposed 4D-GS for real-time 3D dynamic scene rendering. 
Instead of directly constructing 3D Gaussians for each time-stamp with excessive memory costs, 4D-GS first employs a spatial-temporal encoder that utilizes multi-resolution K-Planes~\cite{fridovich2023k} and MLP for effective feature extraction. 
%The compact multi-head MLPs then perform as the decoder and separately predict the deformation of the position, rotation, and scaling of 3D Gaussians based on the input feature. 
The compact multi-head MLPs then perform as the decoder and separately predict the deformation of 3D Gaussians based on the input feature.
%This approach learns the Gaussian deformation field that leads to efficient memory usage and fast convergence.
% MDSplat
Meanwhile, Lu et al.~\cite{lu20243d} proposed a framework that incorporates both canonical and deformation fields. The geometric information is first extracted in the canonical space through a combination of MLPs and 3D U-Net. This information is subsequently utilized to estimate time-dependent 6D Gaussian deformation. The density control strategy is also adapted to dynamic scenarios in which it only considers attributes estimated for the current time stamp.
%Duisterhof et al.~\cite{duisterhof2023md} introduced MD-Splatting to perform 3D point tracking while synthesizing dynamic novel views. 
%MD-Splatting adopts a feature encoding technique~\cite{wu20234d,cao2023hexplane} and learns the Gaussian deformation in a metric space rather than a non-metric canonical one. 
Duisterhof et al.~\cite{duisterhof2023md} introduced MD-Splatting that adopts a feature encoding technique~\cite{wu20234d,cao2023hexplane} and learns the Gaussian deformation in a metric space rather than a non-metric canonical one.
The opacity and scale parameters are not inferred since learning them over time would allow Gaussians to disobey the accurate motion of points. 
Furthermore, the rigidity and isometry losses~\cite{luiten2023dynamic} along with a conservation of momentum loss are incorporated to regularize the trajectory.

% DynMF
%Kratimenos et al.~\cite{kratimenos2023dynmf} presented DynMF to synthesize dynamic views in a real-time fashion. 
Moreover, Kratimenos et al.~\cite{kratimenos2023dynmf} presented DynMF that decomposes complex motions within a given scene into a few basis trajectories from which each point or motion can be adequately derived.
The trajectories are then predicted through an MLP which inherently forms a motion neural field. 
The sharing of neural basis across all points yields physically plausible and frame-consistent sequences. 
To prevent the selection of unnecessary basis and enforce strict sparsity, DynMF adopts a stronger loss term that forces each Gaussian to choose only a few trajectories.
% 3DGStream
Sun et al.~\cite{sun20243dgstream} introduced 3DGStream, a two-stage framework that captures dynamics on the fly for free-viewpoint videos. Inspired by~\cite{muller2021real,muller2022instant}, 3DGStream utilizes a multi-resolution voxel grid as scene representation and adopts hash encoding along with shallow MLP as the Neural Transformation Cache (NTC). During the first stage, NTC is trained to learn the translation and rotation of 3D Gaussians for the next time stamp. Then, Gaussians are adaptively spawned under quality control to reconstruct subsequent frames. 
% PhotoRealistic
Yang et al.~\cite{yang2023real} proposed 4DGS to model space and time as an entirety to address general dynamic scene representation and rendering. 
4DGS extends the scaling and rotation matrix that is decomposed from the original covariance matrix~\cite{kerbl20233d} to 4D Euclidean space. 
The derived general 4D Gaussian representation achieves reasonable fitting to 4D manifold while capturing the underlying dynamics of the scene. 
Additionally, 4DGS allows the appearance to vary with viewpoint and colors to evolve through time by exploiting the spherindirical harmonics, realizing a more faithful rendering of the dynamic scenes from the real world.

% PhysGauss
%Xie et al.~\cite{xie2023physgaussian} introduced PhysGaussian to seamlessly integrate physical simulation within 3D-GS for generating novel dynamics and views. 
%PhysGaussian first reconstructs a static scene via 3D-GS and over-skinny Gaussians could be regularized via an optional anisotropic loss. 
To seamlessly integrate physical simulation within 3D-GS, Xie et al.~\cite{xie2023physgaussian} introduced PhysGaussian for generating faithful novel dynamics and views. It first reconstructs a static scene via 3D-GS and over-skinny Gaussians could be regularized via an optional anisotropic loss.
The simulation of continuum mechanics is incorporated through a time-dependent continuous deformation map. 
Gaussian kernels are then treated as discrete particle clouds and simultaneously deform with the continuum. 
To enforce the deformed kernels under the deformation map to be a Gaussian, PhysGaussian utilizes the first-order approximation to characterize the particles undergoing local affine transformations. 
Additionally, the internal area of objects could be optionally filled via the 3D opacity field to assist in the rendering of exposed internal particles. 

\textbf{Challenges:}
The intrinsic sparsity of input point clouds poses an essential challenge for the reconstruction of scenes with faithful dynamics. 
It is even more challenging to capture physically plausible dynamics while maintaining the quality. 
Moreover, large dynamic motions also remain an open question if no physical constraints are applied.

\textbf{Opportunities:}
Objects with large motions could cause unnatural distortion among consecutive frames and incorporating neural networks with learned scene-specific dynamics could improve the fidelity of deformations. Current methods for reconstruction of dynamic scenes mainly focus on indoor object-level deformations and they still require images taken from multiple camera views along with the precise camera pose. Extending 3D-GS to larger dynamic scenes and introducing physical constraints into 3D-GS optimization might be able to relax these limitations in large dynamic scenes, which is beneficial for real-world applications.

%\vspace{-0.3cm}
\section{Reconstruction}\label{sec:Reconstruction}
The widespread adoption of 3D-GS owes much to its exceptional rendering speed and ability to synthesize realistic scenes from novel viewpoints.
Similar to NeRFs, mesh extraction and surface reconstruction in static scenes is a fundamental yet essential aspect.
Further investigation is necessary to tackle challenging scenarios, such as monocular or few-shot situations, which are commonly encountered in practical applications like autonomous driving.
Additionally, the training times of 3D-GS are on the order of minutes, enabling real-time rendering and facilitating the reconstruction of dynamic scenes.

\begin{figure}[t]
    \centering
    \includegraphics[width=\linewidth]{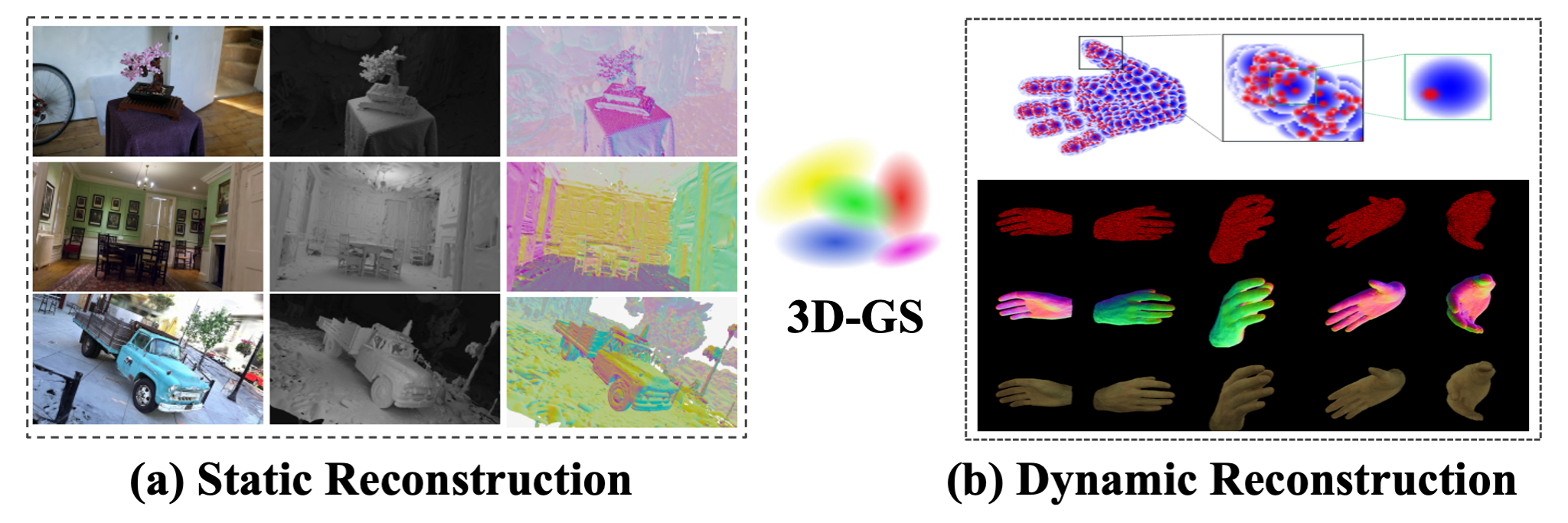}
    %\vspace{-0.9cm}
    \caption{An illustration of 3D-GS on reconstruction: (a) static reconstruction, (b) dynamic reconstruction. Images courtesy of ~\cite{guedon2023sugar,jiang20233d}.}
    \label{fig:recon}
    %\vspace{-0.6cm}
\end{figure}

In this section, as shown in Fig.~\ref{fig:recon}, we delve into exploring related research from two perspectives: static reconstruction and dynamic reconstruction.
% \vspace{-0.4cm}

%\vspace{-0.4cm}
\subsection{Static Reconstruction}
Surface reconstruction and mesh extraction in static scenes is a classic challenge.
However, the explicit representation of 3D-GS distribution introduces a significant level of complexity to this task.
Consequently, several novel methods have been proposed to effectively address this complexity and facilitate static-scene reconstruction.

Gu{\'e}don et al.~\cite{guedon2023sugar} introduced SuGaR for 3D mesh reconstruction and high-quality mesh rendering. SuGaR incorporates a regularization term to promote alignment between the Gaussians and the scene's surface. The Poisson reconstruction method is then used to leverage this alignment and derive a mesh from the Gaussians. 
To bind the Gaussians to the mesh surface, SuGaR presents an optional refinement strategy that optimizes both Gaussians and mesh using 3D-GS rendering. However, the mandatory restrictions on the Gaussians result in a decrease in rendering quality. 
% Nevertheless, these restrictions lead to improved performance in mesh extraction.
Meanwhile, Chen et al.~\cite{chen2023neusg} introduced NeuSG that jointly optimizes NeuS~\cite{wang2021neus} and 3D-GS to achieve highly detailed surface recovery. 
Similar to the regularization term in SuGaR~\cite{guedon2023sugar}, NeuSG incorporates regularizers to ensure that point clouds generated from extremely thin 3D Gaussians closely adhere to the underlying surfaces. NeuSG leverages the benefits of joint optimization, resulting in the generation of comprehensive surfaces with intricate details.

On the other hand, 3D-GS represents a promising solution in monocular and few-shot reconstruction tasks. 
However, a significant challenge in these tasks is the absence of geometric information. 
To this end, numerous methods have been proposed to overcome the lack of geometric information in the perspective.

Few-shot 3D reconstruction allows for reconstructing 3D scenes with limited input data. 
Charatan et al.~\cite{charatan2023pixelsplat} presented pixelSplat for 3D reconstruction from image pairs to address the challenge of scale factor inference by proposing a multi-view epipolar transformer. 
With scale-aware feature maps, pixelSplat predicts the parameters of a set of Gaussian primitives. 
The scene is parameterized through pixel-aligned Gaussians, enabling the implicit densification or deletion of Gaussian primitives during training, avoiding local minima while ensuring a smooth gradient flow. 
Notably, pixelSplat demonstrates exceptional performance in capturing intricate details and accurately inferring 3D structure, particularly in regions of the scene that are only observable from a single reference view.
Zou et al.~\cite{zou2023triplane} utilized two transformer-based networks to perform single-view 3D object reconstruction using a hybrid representation named Triplane-Gaussian, which combines an explicit point cloud with an implicit triplane field. 
By incorporating local image features for projection-aware conditioning within the transformer networks, Triplane-Gaussian enhances consistency with the input observation. 
Triplane-Gaussian leverages both explicit and implicit representations, resulting in efficient and high-quality single-view reconstruction.

Monocular 3D reconstruction is capable of inferring the shape and structure of a 3D scene from 2D images using a single camera. 
The key to monocular 3D reconstruction involves the meticulous analysis of perspective relationships, textures, and motion patterns within the images. By employing monocular techniques, it becomes possible to accurately estimate the distances between objects and discern the overall shape of the scene.
Szymanowicz et al.~\cite{szymanowicz2023splatter} introduced Splatter Image, an ultra-fast method for monocular 3D object reconstruction, demonstrating rapid training and evaluation on both synthetic and real benchmarks without the need for canonical camera poses. 
This approach utilizes a 2D CNN to efficiently process images, predicting a pseudo-image where each pixel is represented by a colored 3D Gaussian. 
% Splatter Image demonstrates rapid training and evaluation on both synthetic and real benchmarks without the need for canonical camera poses. 
% Furthermore, it is also capable of few-shot 3D reconstruction by incorporating cross-view attention.
Das et al.~\cite{das2023neural} introduced Neural Parametric Gaussians (NPGs) for reconstructing dynamic objects from monocular videos. NPGs employs a two-stage reconstruction approach: learning coarse deformation and using it as a constraint for the reconstruction. 
In the initial stage, a coarse parametric point model is implemented based on a low-rank deformation, providing regularization and temporal correspondences. 
The second stage involves optimizing 3D Gaussians, anchored in and deformed by locally oriented volumes. 
NPGs improve non-rigid novel view synthesis from a monocular camera, particularly in challenging scenarios with limited multi-view cues.

%\vspace{-0.4cm}
\subsection{Dynamic Reconstruction}

The high rendering speed and resolution of 3D-GS support dynamic scene reconstruction, including human body tracking and large urban scene reconstruction.

Several methods for dynamic scene reconstruction focus on reconstructing specific dynamic objects or reconstructing small but real-world dynamic scenes from daily life videos.
Jiang et al.~\cite{jiang20233d} introduced 3D-PSHR for real-time dynamic hand reconstruction. 
3D-PSHR incorporates a self-adaptive canonical points upsampling strategy and self-adaptive deformation, enabling pose-free hand reconstruction. 
Furthermore, for texture modeling, 3D-PSHR separates the appearance color into intrinsic albedo and pose-aware shading based on normal deformation.
Lin et al.~\cite{lin2023gaussian} introduced Gaussian-Flow for swift dynamic indoor scene reconstruction and real-time rendering.
% , which facilitates the segmentation, editing, and composition of both static and dynamic 3D scenes. 
Gaussian-Flow introduces the Dual-Domain Deformation Model to capture the time-dependent residual of each attribute through polynomial fitting in the time domain and Fourier series fitting in the frequency domain. Gaussian-Flow can effectively eliminate the necessity of training separate Gaussians for each frame or introducing an additional implicit neural field to model 3D dynamics.
Yang et al.~\cite{yang2023deformable} proposed a deformable 3D-GS for reconstructing scenes by utilizing 3D-GS learned in canonical space with a deformation field to model monocular dynamic scenes. 
This method incorporates an annealing smoothing training mechanism to mitigate the influence of inaccurate poses on the smoothness of time interpolation tasks in real-world datasets. 
This approach attains real-time rendering and high-fidelity scene reconstruction in dynamic scenes.

Reconstructing large dynamic scenes, including real-time dynamic urban scenes and driving scenes, is another challenge.
Lin et al.~\cite{lin2024vastgaussian} introduced VastGaussian for high-fidelity reconstruction and real-time rendering in large scenes. 
VastGaussian incorporates a progressive data partitioning strategy that assigns training views and point clouds to different cells, facilitating parallel optimization and seamless merging. 
Additionally, decoupled appearance modeling is integrated into the optimization process, effectively mitigating floaters caused by appearance variations, which can be discarded after optimization to attain real-time rendering speeds.
Zhou et al.~\cite{zhou2023drivinggaussian} introduced DrivingGaussian for representing and modeling large-scale, dynamic driving scenes. DrivingGaussian also facilitates the simulation of corner cases for various downstream tasks, offering a superior synthesis of surrounding views and reconstruction of dynamic scenes.
Chen et al.~\cite{chen2023periodic} introduced a Periodic Vibration Gaussian (PVG) to modeling large-scale scenes with intricate geometric structures and unconstrained dynamics. 
PVG demonstrates superior performance in both the reconstruction and novel view synthesis of dynamic and static scenes, all without relying on manually labeled object bounding boxes or expensive optical flow estimation.

\textbf{Challenges:}
%As 3D-GS is an explicit representation model for reconstruction, each Gaussian kernel may not necessarily lie on the surface of a certain object, raising a challenge for surface mesh extraction. There is a need to constrain the Gaussian kernel to adhere to the object's surface, but this might lead to a reduction in rendering accuracy.
(i) 3D-GS falls short in capturing intricate geometry, raising a challenge for surface mesh extraction.
There is a need to constrain the Gaussian kernel to adhere to the object's surface, but this might lead to a reduction in rendering quality.
Alternatively, while 3D-GS can integrate with other representations like Signed Distance Functions (SDF) for precise surface modeling, this fusion might incur a decrease in speed.
(ii) Several real-time rendering techniques for dynamic scenes often assume the scene to be static.
(iii) Dynamic scenes with motion blur pose a significant challenge for rendering and reconstruction.

% 高斯的自由度太多，需要优化，高斯之间没有连接关系。解决方案：sugar 2d-gs。效率高 （单一表达）
% sdf 连续，效率低 （组合表达）

\textbf{Opportunities:}
(i) For few-shot reconstruction, integrating with a diffusion model or eliminating the requirement for camera poses could facilitate large-scale training. 
(ii) Additionally, introducing lighting decomposition may extract more realistic surface textures. 
(iii) In dynamic scene reconstruction, prioritizing the optimization of the balance between speed and the preservation of image details may prove to be considerable.

%\vspace{-0.3cm}
\section{Manipulation}\label{sec:Manipulation}
\begin{figure}[t]
    \centering
    \includegraphics[width=\linewidth]{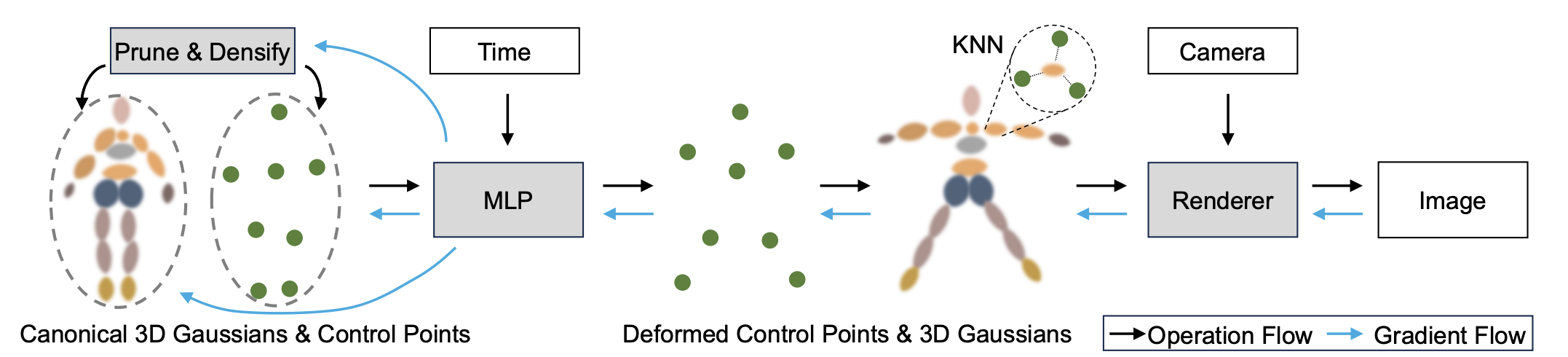}
    %\vspace{-0.8cm}
    \caption{An illustration of employing sparse control points and a deformation MLP to direct 3D Gaussian dynamics. Images courtesy of ~\cite{huang2023sc}.}
    \label{fig:manipulation}
    %\vspace{-0.6cm}
\end{figure}

% \vspace{-0.1cm}
Due to the explicit property of 3D-GS, it has a great advantage for editing tasks as each 3D Gaussian exists individually (Fig.~\ref{fig:manipulation}). It is easy to edit 3D scenes by directly manipulating 3D Gaussians with desired constraints applied.

%\vspace{-0.3cm}
\subsection{3D Manipulation}

\subsubsection{Text-guided or Image-guided Manipulation}
In recent years, there has been a surge in the number of text-guided or image-guided manipulation~\cite{canfes2023text, yu2023towards,hwang2023faceclipnerf}.
% Consequently, text-guided or image-guided manipulation has garnered increased attention.
Furthermore, as access to large language models or visual foundation models becomes more widespread, the utilization of foundation model guided manipulation holds great promise as a future direction.
Fang et al.~\cite{fang2023gaussianeditor} presented GaussianEditor to precisely edit 3D scenes using 3D Gaussians and text instructions. The first step involves extracting the region of interest (RoI) that corresponds to the provided text instruction and aligning it with the 3D Gaussians. This Gaussian RoI is then utilized to control the editing process, enabling fine-grained adjustments.

Most recently, Zhuang et al.~\cite{zhuang2024tip} proposed TIP-Editor, which accepts text and image prompts as well as 3D bounding boxes to specify the editing area. 
Image prompts facilitate the specification of detailed appearance and style for the target content, enhancing the textual description and allowing for precise control.
TIP-Editor utilizes a sequential 2D personalization strategy to enhance the learning of scene representations and reference images.
Additionally, TIP-Editor uses an unambiguous and flexible 3D-GS to facilitate local editing while keeping the background unchanged.

Gao et al.~\cite{gao2024mesh} introduced a mesh-based Gaussian Splatting for representing 3D Gaussians. 
This technique effectively combines the mesh representation with 3D Gaussians, leveraging the mesh to guide the splitting process and enhancing the overall quality of the learned GS. 
They propose a Gaussian deformation strategy on a large scale, which not only utilizes vertex positions but also incorporates deformation gradients to guide the GS.
% By taking advantage of mesh deformation methods, this approach ensures real-time rendering and robustly maintains a high-quality appearance even under substantial deformations.

%\vspace{-0.3cm}
\subsubsection{Non-rigid Manipulation}
Non-rigid objects can change and deform in shape, enabling more realistic simulations of soft objects, biological tissues, and fluids. These objects offer several advantages, including increased authenticity and improved depiction of object deformation and behavior~\cite{tretschk2021non}. Moreover, these models allow for diverse effects as they can respond to external forces and constraints by deforming~\cite{lazova2023control}. However, non-rigid objects also present challenges. They are characterized by their complexity, necessitating careful consideration of factors such as object deformation, continuity, and collision during both editing and simulation. Additionally, the real-time interaction performance of non-rigid objects may be limited in applications, particularly when dealing with large-scale and complex non-rigid objects~\cite{xu2022deforming}.

Pokhariya et al.~\cite{pokhariya2023manus} introduced an articulated 3D Gaussian representation. Specifically, they proposed MANUS-Hand for articulated hands, which leverages 3D-GS to achieve precise shape and appearance representation. 
In addition, they introduced MANUS to combine MANUS-Hand with a 3D Gaussian representation of the object to model contacts accurately. 
They also introduced MANUS-Grasps, an extensive real-world multi-view RGB grasp dataset, comprising over 7 million frames captured by 53 cameras, providing comprehensive 360-degree coverage of more than 400 grasps across various everyday life scenarios. 
% The accuracy of contacts is validated by utilizing paint transfer between the object and the hand.
Lei et al.~\cite{lei2023gart} introduced GART for non-rigid articulated subjects. 
GART approximates the radiance field of the canonical shape and appearance using a Gaussian Mixture Model. 
Additionally, GART can be effectively animated through a learnable forward skinning technique. 
This representation is capable of capturing complex deformations, including loose clothing on human subjects.

%\vspace{-0.3cm}
\subsubsection{Time-efficient Editing}
While 3D-GS is indeed a rapid rendering technique, it is crucial to operate in real-time while editing 3D Gaussians. Consequently, there is an urgent need to develop editing methods for 3D-GS that are time-efficient.

Huang et al.~\cite{huang2023point} proposed Point'n Move to enable interactive manipulation of scene objects through exposed region inpainting. 
The interactivity is enhanced by intuitive object selection and real-time editing. 
To achieve this, they took advantage of the explicit nature and speed of 3D-GS. The explicit representation formulation allows for a dual-stage self-prompting segmentation algorithm, where 2D prompt points are used to create 3D masks. 
This algorithm facilitates mask refinement and merging, minimizes changes, provides a good initialization for scene inpainting, and enables real-time editing without the need for per-editing training. 
At the same time, Chen et al.~\cite{chen2023gaussianeditor} introduced GaussianEditor for 3D editing that employs 3D-GS to augment control and efficiency throughout the editing procedure. 
GaussianEditor employs Gaussian semantic tracing to accurately identify and target specific areas for editing. 
It then utilizes Hierarchical Gaussian Splatting to strike a balance between fluidity and stability. 
Moreover, GaussianEditor includes a specialized 3D inpainting algorithm, which streamlines the removal and integration of objects and significantly reduces editing time.

%\vspace{-0.3cm}
\subsection{4D Manipulation}
The field of 4D scene reconstruction has experienced notable progress with the introduction of dynamic neural 3D representation. 
These advancements have greatly improved the capacity to capture and depict dynamic scenes. 
However, despite these breakthroughs, the interactive editing of these 4D scenes continues to present significant obstacles. 
The main challenge lies in guaranteeing spatial-temporal consistency and maintaining high quality during 4D editing, while also providing interactive and advanced editing capabilities.

Shao et al.~\cite{shao2023control4d} introduce Control4D for editing dynamic 4D portraits using text instructions. Control4D aims to overcome the challenges commonly encountered in 4D editing, particularly the limitations of existing 4D representations and the inconsistent editing outcomes resulting from diffusion-based editors. GaussianPlanes was first proposed as a novel 4D representation, which enhances the structure of Gaussian Splatting through plane-based decomposition in both 3D space and time. This approach improves the efficiency and robustness of 4D editing.
Additionally, a 4D generator is leveraged to learn a more continuous generation space from the edited images produced by the diffusion-based editor, effectively enhancing the consistency and quality of 4D editing. 
Huang et al.~\cite{huang2023sc} proposed SC-GS to model scene motion using sparse control points in conjunction with an MLP. 
% They leverage the understanding that motions within a scene can be effectively represented by a compact subspace comprising a sparse set of bases. 
To facilitate the accurate learning of appearances, geometry, and motion from monocular videos, adaptive learning strategies are employed, along with the incorporation of a regularization loss that enforces rigid constraints. By adopting this sparse motion representation, SC-GS enables motion editing by manipulating the learned control points while preserving high-fidelity appearances.
Yu et al.~\cite{yu2023cogs} introduced Controllable Gaussian Splatting (CoGS) for dynamic scene manipulation.
% Unlike NeRFs and other neural methods, CoGS utilizes an explicit representation, allowing for real-time and controllable manipulation of dynamic scenes. 
The explicit nature of CoGS not only enhances rendering efficiency but also simplifies the manipulation of scene elements.
The control signal can be obtained through a three-step process: (i) Selecting a set of Gaussians to accurately represent movement within the control part. (ii) Applying Principal Component Analysis to identify the primary movement direction. (iii) Normalizing the distances of all points from the starting point to a range between 0 and 1, thereby generating the control signal.
After acquiring the control signal, the subsequent critical step involves integrating it into the network to enable manipulation using these signals. 
This integration is achieved by constructing a unique network for each control signal. 
These networks are specifically designed to generate the corresponding offset for each Gaussian attribute, as determined during the dynamic modeling phase. 
This ultimate step ensures accurate and efficient control over dynamic scene renderings.

% \textcolor{blue}{
% The recent work SC-GS proposed by Huang et al.~\cite{huang2023sc} explicitly decomposes the motion and appearance of dynamic scenes into sparse control points and dense Gaussian distributions, respectively. 
% The key idea is to use a smaller number of sparse control points to learn compact 6-DoF transformation bases. 
% They adopted a deformed MLP to predict the time-varying 6-DoF transformation of each control point, which reduced the learning complexity, enhanced the learning ability, and facilitated the acquisition of temporally and spatially coherent motion patterns. 
% They jointly learn 3D Gaussians, canonical spatial locations of control points, and deformation MLPs to reconstruct the appearance, geometry, and dynamics of 3D scenes. 
% During the learning process, the position and number of control points are adaptively adjusted to adapt to the different motion complexities in different regions and an ARAP loss following the principle of being as rigid as possible is developed to enhance the spatial continuity and local rigidity of the learned motion. 
% }

\textbf{Challenges:}
(i) The selection of RoIs relies on the performance of segmentation models, which are influenced by noises.
(ii) When editing 3D Gaussians, several important physical aspects are often overlooked. large editing will lead to artifacts.
(iii) There is still potential for enhancing frame consistency in 4D editing.

\textbf{Opportunities:}
(i) To enhance the accuracy of editing, it is possible to incorporate more advanced segmentation models into the 3D-GS framework.
(ii) More complex and multiple physics should be considered to ensure smooth editing.
(iii) Existing methods have primarily been tested with minimal motion changes and accurate camera poses. Expanding their applicability to scenarios involving intense movements remains an area of investigation.
(iv) Existing 2D diffusion models encounter difficulties in providing adequate guidance for intricate prompts, resulting in constraints when it comes to 3D editing. Therefore, efficient and accurate 2D diffusion models can be utilized as better guidance for editing 3D Gaussians.

%\vspace{-0.4cm}
\section{Generation}\label{sec:Generation}
\begin{figure}[t]
    \centering
    \includegraphics[width=\linewidth]{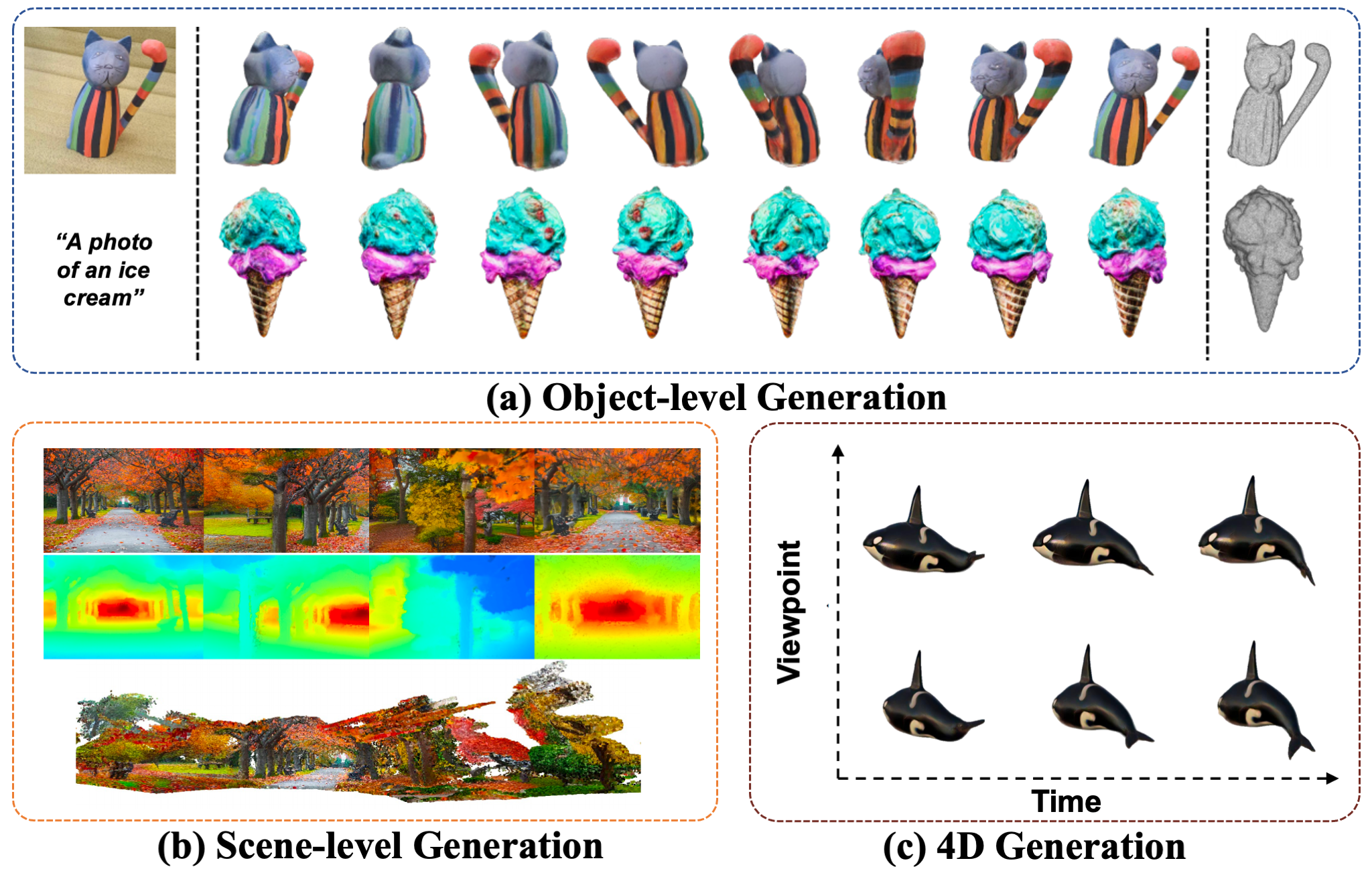}
    %\vspace{-0.8cm}
    \caption{An illustration of 3D-GS on generation tasks: (a) object-level generation, (b) scene-level generation, and (c) 4D generation. Images courtesy of ~\cite{tang2023dreamgaussian, ouyang2023text2immersion, yin20234dgen}.}
    \label{fig:generation}
    %\vspace{-0.6cm}
\end{figure}

Thanks to the significant progress made in diffusion models and 3D representations, generating 3D assets from text/image prompts is now a promising task in AIGC. 
Furthermore, resorting to 3D-GS as the explicit representation of the object (Fig.~\ref{fig:generation}a) and the scene (Fig.~\ref{fig:generation}b) makes fast or even real-time rendering possible. 
% Besides, some works focus on improving the time-consuming optimization process inherent in the Score Distillation Sampling (SDS) pipeline (Fig.~\ref{fig:generation}c). 
While 3D generation has shown some impressive results, 4D generation (Fig.~\ref{fig:generation}c) remains a challenging and under-explored topic.
% Fortunately, numerous attempts have been made to expand the current landscape.

%\vspace{-0.3cm}

%\vspace{-0.2cm}
\subsection{Object-level 3D Generation}

3D diffusion models provide 3D consistency in 3D generation, while 2D diffusion models enjoy strong abilities of generalization. 
Combining the best of both worlds, Yi et al.~\cite{yi2023gaussiandreamer} proposed GaussianDreamer for fast generation and real-time rendering.
GaussianDreamer first initializes 3D Gaussians with the assistance of the 3D diffusion model to acquire geometry priors and introduces two operations of noisy point growing and color perturbation to supplement the initialized Gaussians for further content enrichment. 
Later, 3D Gaussians are optimized with the help of the 2D diffusion model and text prompts via the SDS. 
However, such a method still suffers from multi-face problems.
Chen et al.~\cite{chen2023text} proposed to guide the generation with the ordinary 2D SDS loss and an additional 3D point cloud diffusion prior to mitigating the Janus problem~\cite{armandpour2023re}.
Specifically, this method first initializes 3D Guassians from customized shapes or Point-E~\cite{nichol2022point} and then optimizes the representations with a combination of 2D and 3D SDS loss in the geometry optimization stage. 
% In the appearance refinement stage, a compactness-based densification strategy is iteratively conducted to enhance geometry continuity and improve fidelity.
Also using SDS, Xu et al.~\cite{xu2024agg} proposed an amortized feed-forward pipeline called AGG, which can generalize to unseen objects of similar categories. 
AGG can reduce the need for test-time optimization by trading off the computation cost of the inference stage with the training stage. 
AGG takes a single image as input and decomposes the geometry and texture generation task into two distinct networks to produce 3D Gaussians at a low resolution. 
Subsequently, a UNet with point-voxel layers is utilized to super-resolve the 3D-GS, improving its fidelity. 
% Since the adaptive density control operation is not available in the amortized generation setting, AGG uses a fixed number of 3D Gaussians for each object and warms up the network with 3D pseudo labels. 
However, such a practice also limits the number of Gaussians generated to represent very complex geometry.
Tang et al.~\cite{tang2023dreamgaussian} propose DreamGaussian to improve the 3D generation efficiency by replacing NeRF with 3D-GS. 
Specifically, DreamGaussian simplifies the optimization landscape with the progressive densification of 3D-GS, which initializes Gaussians with random positions and densifies them periodically to align with the generation progress. 
To enhance the generation quality, it further introduces an efficient mesh extraction algorithm with the block-wise local density query and a UV-space texture refinement stage that performs image space supervision.

The over-smoothing effect in SDS-based 3d generation methods has been identified.
It has been revealed that the over-smoothing issue is mainly caused by the inconsistent and low-quality pseudo-GTs generated by the 2D diffusion models. 
Motivated by these observations, Liang et al.~\cite{liang2023luciddreamer} proposed an Interval Score Matching (ISM) to counteract such deficiency. 
Specifically, DDIM inversion is introduced to increase the consistency of pseudo-GTs, making them more aligned with the inputs. 
Furthermore, to obtain pseudo-GTs with better visual quality, ISM chooses to conduct matching between two interval steps in the diffusion trajectory instead of matching the pseudo-GTs with images rendered by the 3D model. 
% Combined with 3D Gaussian Splatting, this method achieves superior results without additional computation cost compared to the state-of-the-art approaches\cite{chen2023fantasia3d}\cite{wang2023prolificdreamer}.

%\vspace{-0.4cm}
\subsection{Scene-level 3D Generation}
Vilesov et al.\cite{vilesov2023cg3d} proposed CG3D to compositionally generate scalable 3D assets to form physically realistic scenes from text input only. 
CG3D represents each object in a scene with a set of Gaussian and converts objects into compositional coordinates with interaction parameters like rotation, translation, and scale.

Diffusion models trained on 3D scan datasets are inherently limited to generating domain-specific scenes due to the unrealistic nature of the training sets. LucidDreamer~\cite{chung2023luciddreamer} is proposed to create domain-free and high-quality 3D scenes from various types of inputs such as text, RGB and RGBD. With the help of Stable Diffusion~\cite{rombach2022high} and the estimated depth map, LucidDreamer first initializes the point cloud and then refines the point cloud through Dreaming and Alignment processes to form the SfM points for the optimization of Gaussian splats. 
The dreaming process creates new 3D points by completing partially projected images and lifting them into 3D space while the Alignment process seamlessly connects the new 3D points to the existing point cloud. 
Similarly, Text2Immersion~\cite{ouyang2023text2immersion} produces consistently immersive and photo-realistic scenes from text prompts with a two-stage optimization of 3D Gaussians. 
In the first stage, an initial coarse Gaussian cloud is constructed incrementally, leveraging pre-trained 2D diffusion and depth estimation models under a set of anchor cameras rotated from the center view. 
Since the anchor cameras are selected only by rotation but not translation, a large portion of missing regions and noisy appearance may exist in the generated scene. 
Thus, the second stage focuses on 3D Gaussian refinement by incorporating additional views to fill in the missing areas and mitigate the issue of Gaussian noises.

\subsection{4D Generation}
Ling et al.~\cite{ling2023align} introduced Align Your Gaussians (AYG) to extend 3D synthesis to 4D generation with an additional temporal dimension. 
The 4D representation combines 3D Gaussian with deformation fields, which model scene dynamics of 3D Gaussians and transform the collection of them to represent object motion. 
AYG begins with generating an initial static 3D shape with a 3D-aware multi-view diffusion model and a regular text-to-image model. 
Then, a text-to-video model and a text-to-image model are used to optimize the deformation field to capture temporal dynamics and maintain high visual quality for all frames, respectively. 
% To stabilize the optimization, a novel regularization method is used to preserve the distribution of the moving 3D Gaussians. 
Furthermore, a motion amplification mechanism as well as a new autoregressive synthesis scheme are adopted to generate and combine multiple 4D sequences for longer generations. 
Notably, due to the explicit nature of 3D Gaussians, different dynamic scenes, each with its own set of Gaussians and deformation field, can be combined, thereby enabling the composition of multiple 4D objects into large dynamic scenes. 
4DGen~\cite{yin20234dgen} also leverages static 3D assets and monocular video sequences to conduct 4D generation. 
Both components can either be specified by users or generated by the multi-view diffusion model SyncDreamer~\cite{liu2023syncdreamer} and the video generation model. 
To guarantee consistency across frames, SyncDreamer is also used to construct point clouds from the anchor frames as 3D pseudo labels for scene deformation.
% 4DGen further adopts seamless consistency priors implemented with SDS and unsupervised smoothness regularization.

NeRF-based 4D generation methods, like their counterparts in 3D domain, also experience long optimization time problems. DreamGaussian4D~\cite{ren2023dreamgaussian4d}, a follow-up work of DreamGaussian~\cite{tang2023dreamgaussian}, utilizes the explicit modeling of spatial transformations in 3D-GS to simplify the optimization in image-to-4D generation. 
It first creates static 3D Gaussians from the input image using an enhanced variant of DreamGaussian called DreamGaussianHD. 
% This variant pipeline aims to alleviate the blurriness problem in the original model. 
For the dynamic optimization, a driving video obtained from Stable Video Diffusion is employed to optimize a time-dependent deformation field upon the static 3D Gaussians so as to learn the controllable and diverse motions. 
Finally, a video-to-video texture refinement strategy is adopted to enhance the quality of exported animated meshes.
% , making the framework easier to deploy in real-world applications. 
% Compared to previous work \cite{singer2023textto4d}\cite{jiang2023consistent4d}\cite{bahmani20234dfy}\cite{zheng2023unified}, DreamGaussian4D reduces the optimization time from hours to just a few minutes. 

To improve the efficiency in 4D generation domain, Pan et al.~\cite{pan2024fast} propose Efficient4D, a fast video-to-4D object generation framework free of heavy supervision back-propagation through the large pre-trained diffusion model. 
To obtain labeled data for the training of 4D Gaussian Splating, it utilizes an enhanced version of SyncDreamer to generate multi-view images of sampled frames in the input single-view video.
Specifically, Efficient4D revises SyncDreamer with a time-synchronous spatial volume and a smoothing filter layer to impose temporal consistency across frames. 
During the reconstruction stage, Efficient4D optimizes 4D Gaussian representation using an inconsistency-aware loss function for coherent modeling of dynamics in both space and time.
% Experimental results showcase a 10-fold speed up in generation compared to Consistent4D \cite{jiang2023consistent4d}. 

% \subsection{Challenges and Opportunities}

\textbf{Challenges:}
(i) Compositional generation remains an open problem since most methods do not support such creation\cite{yin20234dgen, yi2023gaussiandreamer, tang2023dreamgaussian, xu2024agg}. Even though CG3D \cite{vilesov2023cg3d} is a compositional framework, it supports only rigid-body interactions between objects. Besides, the compositional 4D sequences in AYG~\cite{ling2023align} fail to depict the topological changes of the dynamic objects. 
(ii) It is non-trivial to adapt the adaptive density control operation in the original 3D-GS to the generation framework and fix the number of Gaussians used to represent an object. However, such a design severely limits the model’s capability of creating complex geometry.

\textbf{Opportunities:}
(i) The multi-face problem, also known as Janus problem, exists in most 2D lifting methods \cite{tang2023dreamgaussian, vilesov2023cg3d}. GaussianDreamer~\cite{yi2023gaussiandreamer} manages to alleviate such deficiency by introducing 3D priors. In light of this, utilizing 3D-aware diffusion models or multi-view 2D diffusion models may be the possible direction for further improvement. 
(ii) Text-to-3D methods tend to produce unsatisfying results when the textual prompt is comprised of ambiguous information and complicated logic. In this regard, enhancing the language understanding ability of the text encoder may also be able to boost the generation quality. 
(iii) The generation of 3D scenes is currently constrained by the level of detail. Thus, exploring compositional generation for 3D scenes emerges as a promising direction.

%\vspace{-0.3cm}
\section{Perception}\label{sec:Perception}

% \vspace{-0.3cm}
\begin{figure}[t]
    \centering
    \includegraphics[width=0.9\linewidth]{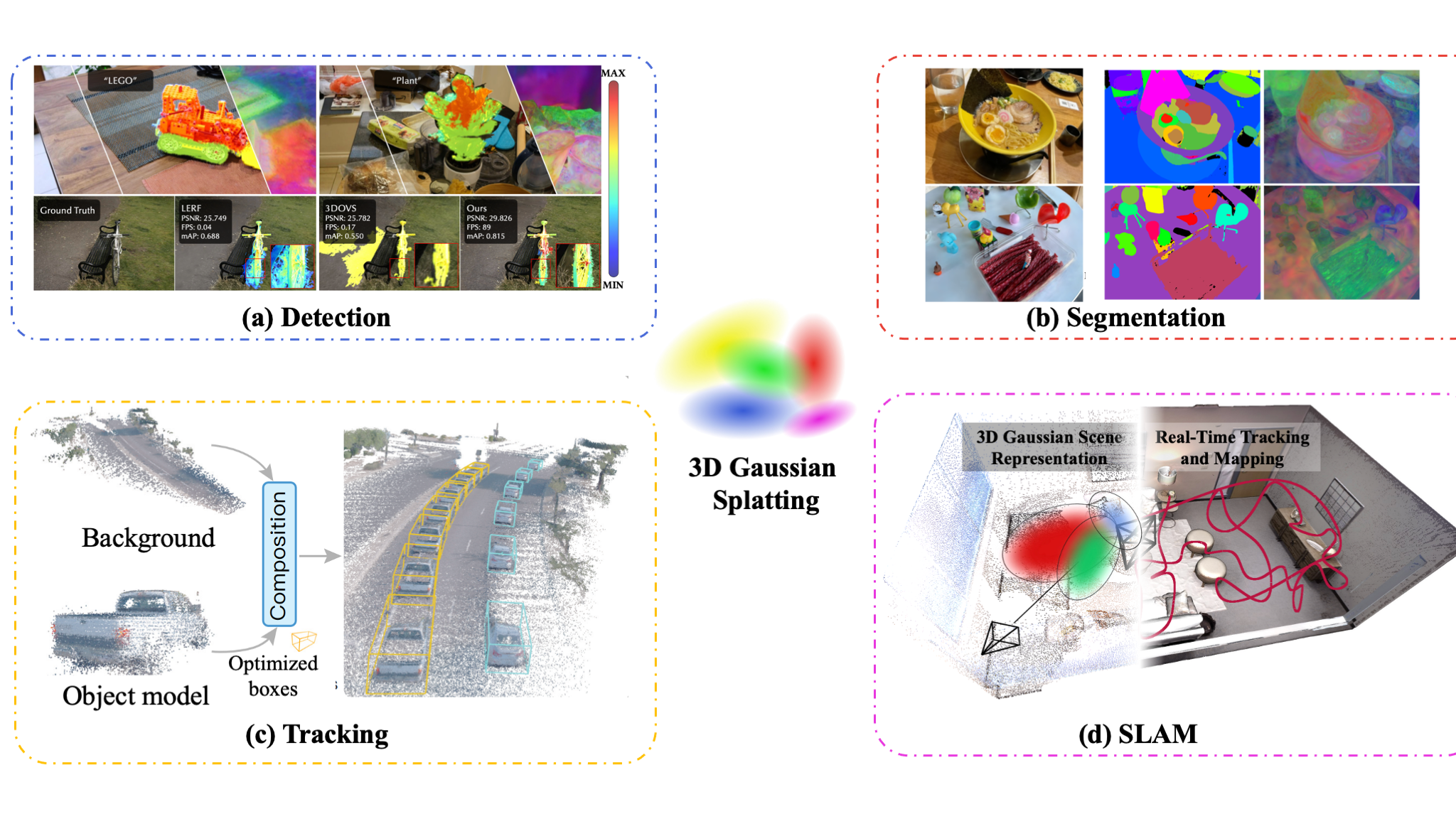}
    %\vspace{-1.1cm}
    \caption{An illustration of 3D-GS on perception: (a) detection, (b) segmentation, (c) tracking, and (d) SLAM. Images courtesy of ~\cite{zhou2023feature,shi2023language,ye2023gaussian,yan2024street}.}
    \label{fig:perception}
    %\vspace{-0.5cm}
\end{figure}

% 3D perception can be subdivided into various tasks, including detection, segmentation, tracking, and Simultaneous Localization and Mapping (SLAM) systems.
Leveraging 3D-GS, 3D perception has the potential to enhance open-vocabulary semantic object detection and localization (Fig.~\ref{fig:perception}a), 3D segmentation (Fig.~\ref{fig:perception}b), tracking of moving objects (Fig.~\ref{fig:perception}c), and Simultaneous Localization and Mapping (SLAM) systems (Fig.~\ref{fig:perception}d).
Note that in domains like perception and scene comprehension in autonomous driving, numerous studies encompass at least two tasks.

%\vspace{-0.3cm}
\subsection{Detection}

The process of semantic object detection or localization in a 3D scene can significantly enhance environmental understanding and perception, as well as benefit applications such as autopilot systems and intelligent manufacturing.

Encouraged by ChatGPT, Shi et al.~\cite{shi2023language} introduced Language Embedded 3D Gaussians, a scene representation specifically crafted for open-vocabulary query tasks, successfully incorporating quantized compact semantic features onto extensive 3D-GS, minimizing memory and storage requirements. 
To mitigate semantic inconsistencies arising from varying perspectives, a feature smoothing procedure is proposed to dynamically reduce the spatial frequency of embedded semantic features, leveraging spatial position and semantic uncertainty of 3D-GS.
At the same time, Zuo et al.~\cite{zuo2024fmgs} proposed Foundation Model Embedded Gaussian Splatting (FMGS), which integrates 3D-GS for the representation of geometry and appearance, along with Multi-Resolution Hash Encodings for efficient language embedding. 
FMGS exhibits multi-view semantic consistency and impressive performance in localizing semantic objects within an open-vocabulary context.
% FMGS aims to tackle memory limitations in room-scale scenes. 
% Moreover, to tackle the issue of pixel misalignment, FMGS incorporates a pixel alignment loss to align the rendered feature distance of identical semantic entities with pixel-level semantic boundaries. 
% The results of FMGS exhibit significant multi-view semantic consistency and impressive performance in localizing semantic objects within an open-vocabulary context.
% Kruse et al.~\cite{kruse2024splatpose} presented a novel pose-agnostic method for anomaly detection, representing the object as a 3D point cloud of Gaussians, which is used for pose estimation, and finding anomalies in images without prior pose information.

%\vspace{-0.4cm}
\subsection{Segmentation}

The significance of 3D scene segmentation lies not only in improving the accuracy of scene segmentation but also in providing robust support for real-world 3D perception tasks. 
The applications of 3D scene segmentation methods, ranging from real-time scene editing and object removal to object inpainting and scene recomposition, have undoubtedly broadened the horizons of computer vision in domains including virtual reality and autonomous driving.

% The incorporation of a 2D segmentation model can be a valuable asset in guiding the segmentation process of 3D-GS. 
% This intuitive concept has the potential to improve both the accuracy and efficiency of the segmentation procedure.
% Lan et al.~\cite{lan20232d} introduced a 3D Gaussian segmentation method that utilizes 2D segmentation as supervision, where each 3D Gaussian is assigned an object code to represent its categorical probability distribution.
% Guidance is provided to ensure the accurate classification of each 3D Gaussian by minimizing the discrepancy between the 2D segmentation map and the rendered segmentation map at a specific pose.
% Moreover, KNN clustering is employed to tackle the issue of semantic ambiguity within 3D Gaussians, while statistical filtering is implemented to eliminate incorrectly segmented 3D Gaussians. This approach successfully acquires the semantic knowledge of a 3D scene and efficiently segments multiple objects from a specific viewpoint within a brief timeframe, yielding compelling outcomes.

Several approaches draw inspiration from 2D visual foundation models~\cite{kirillov2023segment, liu2023segment} to leverage their robust zero-shot segmentation capabilities for guiding the segmentation of 3D-GS. 
In this way, the advantages of the 2D model can be utilized to improve the accuracy and effectiveness of the segmentation process in the 3D domain.
% By employing these methods, researchers can harness the advantageous features of the 2D model to improve the accuracy and effectiveness of the segmentation process in the 3D domain.
For instance, Ye et al.~\cite{ye2023gaussian} introduced Gaussian Grouping for simultaneous reconstruction and segmentation of diverse elements within open-world 3D scenes. 
The proposal includes an Identity Encoding for 3D Gaussians, supervised by both 2D mask predictions from SAM and 3D spatial consistency, which enables each Gaussian to be associated with its represented instances or elements in the 3D scene. 
% Leveraging this discrete and grouped 3D scene representation, Gaussian Grouping showcases its ability to support various scene editing applications, encompassing 3D object removal, object inpainting, and scene recomposition. 
% Gaussian Grouping achieves a high-quality visual effect while maintaining efficient processing time. 
Meanwhile, Cen et al.~\cite{cen2023segment} presented the Segment Any 3D Gaussians (SAGA) to integrate a 2D segmentation foundation model with 3D-GS for 3D interactive segmentation. 
SAGA incorporates multi-granularity 2D segmentation results from the foundation model into 3D Gaussian point features, eliminating the requirement for the time-consuming multiple forwarding of the 2D segmentation model during inference. 
SAGA trains 3D features for Gaussians based on automatically extracted masks, and a set of queries is generated with input prompts, efficiently retrieving the expected Gaussians through feature matching. 
Furthermore, SAGA facilitates multi-granularity segmentation and accommodates various prompts, such as points, scribbles, and 2D masks. 
Also inspired by 2D foundation models, Zhou et al.~\cite{zhou2023feature} proposed Feature-3DGS, a feature field distillation technique based on 3D-GS. 
Feature-3DGS initially acquires a semantic feature from each 3D Gaussian, in addition to the color information. By utilizing differentiable splatting and rasterization techniques on the feature vectors, Feature-3DGS facilitates the extraction of the feature field while being guided by 2D foundation models.
Furthermore, Feature-3DGS acquires knowledge of a well-organized, lower-dimensional feature field. This field is subsequently enhanced through the utilization of a streamlined convolutional decoder during the final stages of the rasterization process.
Feature-3DGS facilitates complex semantic tasks such as editing, segmentation, and language-prompted interactions.
Hu et al.~\cite{hu2024semantic} proposed a SA-GS for segmenting objects in 3D-GS. SA-GS does not necessitate any training process or learnable parameters. The segmentation accomplished through SA-GS facilitates efficient collision detection and scene editing operations, encompassing object removal, translation, and rotation within 3D scenes.

%\vspace{-0.4cm}
\subsection{Tracking}

As discussed in Sec.~\ref{sec:Reconstruction}, utilization of 3D-GS facilitates the reconstruction of dynamic scenes.
Consequently, the tracking of dynamic objects within these scenes has emerged as a novel area of exploration, contributing significantly to applications such as autonomous driving.

%Zhou et al.~\cite{zhou2023drivinggaussian} introduced DrivingGaussian for reconstructing dynamic large-scale driving scenes.
%DrivingGaussian progressively models the static background by utilizing incremental static 3D Gaussians in scenes that contain moving objects.
%DrivingGaussian utilizes a composite dynamic Gaussian graph to accurately reconstruct individual objects, restore their positions, and effectively handle occlusion relationships in the presence of multiple moving objects.
%Moreover, the utilization of LiDAR prior to 3D-GS facilitates the improvement of scene reconstruction by capturing finer details and ensuring the preservation of panoramic consistency.
%DrivingGaussian successfully achieves high-fidelity and multi-camera consistent photorealistic surround-view synthesis, making it applicable for a wide range of tasks, including the simulation of corner cases.

Yan et al.~\cite{yan2024street} introduced StreetGaussian for modeling dynamic urban street scenes from monocular videos. 
StreetGaussian represents the dynamic urban street as a collection of point clouds that are augmented with semantic logits and 3D Gaussians. 
% These components are assigned to either foreground vehicles or the background. 
To accurately represent the movement of foreground objects, they optimize each object's point cloud by incorporating trackable poses and a dynamic spherical harmonics model to account for their dynamic appearance.
%Despite its ability to facilitate the seamless creation of object vehicles and backgrounds for scene editing operations, as well as its capability to achieve high-speed rendering, this approach does have certain limitations. 
%Specifically, StreetGaussian is limited to reconstructing rigid dynamic scenes, such as stationary streets with only moving vehicles, and is incapable of handling non-rigid dynamic objects, such as pedestrians in motion.

Since tracking the movement of small objects can be challenging, Cotton et al.~\cite{cotton2024dynamic} applied dynamic Gaussian splatting~\cite{luiten2023dynamic} with depth supervision to analyze sparse, markerless motion capture data, specifically focusing on the accurate assessment of movement in infants and neonates.
% This approach employs semantic segmentation masks to enhance the emphasis on the infant, thereby resulting in improved scene initialization.
Additionally, an alternative dynamic tracking method using deformation fields was implemented based on Deformable 3D-GS~\cite{yang2023deformable}.
This method models the changes in location, rotation, and scale by considering the initial 3D locations and the desired time point. 
Deformable 3D-GS has illustrated its ability to produce innovative viewpoints of scenes and track the movements of infants.

%\vspace{-0.4cm}
\subsection{Simultaneous Localization and Mapping}

In the realm of 3D perception, the integration of 3D-GS into Simultaneous Localization and Mapping (SLAM) systems has garnered significant attention.
In this section, we will explore the diverse applications and advancements in SLAM by integrating 3D-GS. 
Furthermore, this section underscores the efficacy of current methodologies in addressing real-world scenarios and highlights the continuous growth of possibilities within the SLAM domain.

%speed
The real-time rendering capabilities powered by 3D-GS set the stage for online SLAM systems, paving the way for applications such as autonomous driving.
Due to the significance of efficiency, Yan et al.~\cite{yan2023gs} presented GS-SLAM, which employs a real-time differentiable splatting rendering pipeline, significantly enhancing map optimization and RGB-D re-rendering speed. 
It introduces an adaptive strategy for efficiently expanding 3D-GS to reconstruct newly observed scene geometry. 
Additionally, GS-SLAM uses a coarse-to-fine technique to select reliable 3D-GS, improving camera pose estimation accuracy.
% By balancing efficiency and accuracy, GS-SLAM outperforms recent SLAM methods utilizing neural implicit representations.
%speed
Meanwhile, Matsuki et al.~\cite{matsuki2023gaussian} introduced a real-time SLAM system for incremental 3D reconstruction.
This approach is applicable to both moving monocular and RGB-D cameras, enabling near real-time performance.
The proposed method formulates camera tracking for 3D-GS through direct optimization against the 3D Gaussians, facilitating fast and robust tracking with a broad basin of convergence. 
Leveraging the explicit nature of the Gaussians, the method introduces geometric verification and regularization to handle ambiguities in incremental 3D dense reconstruction.
% Notably, this method achieves great results in novel view synthesis, trajectory estimation, and the reconstruction of tiny and even transparent objects.
Moreover, Huang et al.~\cite{huang2023photo} proposed Photo-SLAM to introduce a hyper primitives map by integrating both explicit geometric features for localization and implicit photometric features for representing texture information. 
%Notably, Photo-SLAM incorporates Gaussian-Pyramid-based learning to efficiently capture multi-level features, thereby enhancing the photorealistic quality of the mapping process.
The efficacy of Photo-SLAM in online photorealistic mapping is demonstrated, showcasing its potential for robotic applications in real-world scenarios.
%Photo-SLAM demonstrates efficacy in online photorealistic mapping and exhibits potential for applications in advanced robotics in real-world scenarios.

% monocular
To effectively utilize monocular input data from SLAM, it is crucial to design meticulous methods that incorporate 3D-GS.
Keetha et al.~\cite{keetha2023splatam} presented SplaTAM for dense SLAM with a single monocular RGB-D camera. 
SplaTAM begins by estimating camera pose using silhouette-guided differentiable rendering. 
It then dynamically expands map capacity by integrating new Gaussians from rendered silhouettes and depth data. 
SplaTAM updates the Gaussian map through differentiable rendering, providing explicit knowledge of map extent and streamlining map densification. 
Yugay et al.~\cite{yugay2023gaussian} introduced strategies to optimize 3D-GS for monocular RGB-D setups. 
This method extends 3D-GS to enhance geometry encoding, enabling reconstruction beyond radiance fields in monocular settings. 
This representation facilitates interactive-time reconstruction and photo-realistic rendering.

%\textcolor{blue}{\subsection{Camera Pose Estimation}}

%Camera pose estimation stands as a foundational aspect in the realms of 3D reconstruction and perception. The incorporation of 3D-GS has the potential to provide insightful approaches to this essential topic.

%In SLAM, the task of estimating the 6D pose poses a considerable challenge.
%To solve this challenge, Sun et al.~\cite{sun2023icomma} introduced iComMa by integrating traditional geometric matching methods with rendering comparison techniques.
%iComMa inverts the 3D-GS to capture pose gradient information for precise pose computation and adopts a render-and-compare strategy to ensure increased accuracy during the final phase of optimization. 
%Additionally, a matching module is incorporated to enhance the model's robustness against unfavorable initializations via minimizing distances between 2D keypoints. 
%iComMa is designed to effectively handle a wide range of complex and challenging scenarios, including cases with significant angular deviations while maintaining a high level of accuracy in predicting outcomes.

% \subsection{Challenges and Opportunities}
\textbf{Challenges:}
(i) The detection of highly reflective or translucent objects remains a challenging task as the modeling capabilities of 3D-GS for these objects are limited.
(ii) A Gaussian distribution can be linked to multiple objects, thereby increasing the complexity of accurately segmenting individual objects utilizing feature matching.
(iii) The existing tracking methods for dynamic scene objects based on 3D-GS may encounter challenges in tracking deformable objects, such as pedestrians. 
(iv) A SLAM system might exhibit sensitivity to various factors, including motion blur, substantial depth noise, and aggressive rotation. Motion blur occurs when there is rapid movement. Substantial depth noise refers to inaccuracies or fluctuations in the distance measurements. Aggressive rotation implies swift and sudden rotations.

\textbf{Opportunities:}
(i) Considering the rapid real-time reconstruction capabilities of 3D-GS, detection can be utilized in dynamic scenes.
(ii) Semantic segmentation based on 3D-GS may leverage other Foundation Models to enhance the segmentation accuracy.
(iii) Real-time tracking based on 3D-GS has the potential to be applied in various medical scenarios, including radiation therapy.
(iv) Inputting known camera intrinsics and dense depth is essential for executing SLAM, and eliminating these dependencies presents an intriguing direction for future exploration.

%\vspace{-0.4cm}
\section{Virtual Humans}
\label{sec:Human}

Learning virtual human avatars with implicit neural representations like NeRF and SDF suffers from long optimization and rendering time and struggles to generate satisfactory-quality novel body poses. 
In contrast, utilizing the 3D-GS is experimentally demonstrated to improve training and rendering speeds and offer explicit control of human body deformation. 
Also, forward skinning in 3D Gaussian-based methods avoids the correspondence ambiguities that are present in inverse skinning used in neural implicit representations~\cite{jena2023splatarmor}.

\begin{figure}[t]
    \centering
    \includegraphics[width=\linewidth]{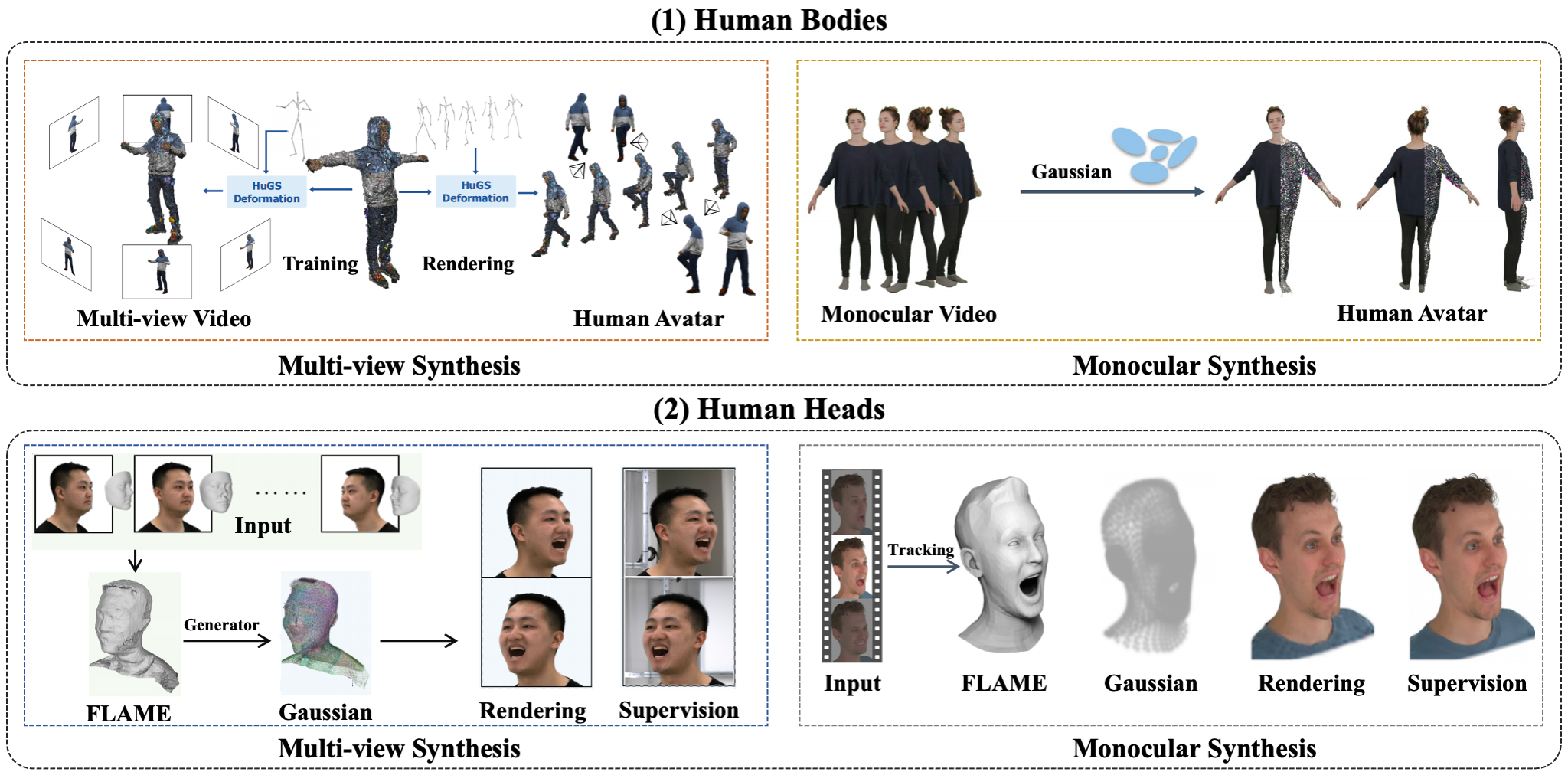}
    %\vspace{-0.8cm}
    \caption{An illustration of 3D-GS on virtual humans for modeling (1) bodies and (2) heads. Images courtesy of ~\cite{moreau2023human, li2024gaussianbody, qian2023gaussianavatars, xu2023gaussian}.}
    \label{fig:human}
    %\vspace{-0.5cm}
\end{figure}

Typically, 3D Gaussian-based methods start with initializing Gaussians with an SMPL body template (Fig.~\ref{fig:human}(1)) and then deform the representations into the observation space using Linear Blend Skinning (LBS). Gaussians are then rendered and supervised by either multi-view (Fig.~\ref{fig:human}(Left)) or monocular videos (Fig.~\ref{fig:human}(Right)). 
Besides, some methods specialize in reconstructing human head avatars (Fig.~\ref{fig:human}(2)).

%\vspace{-0.3cm}
\subsection{Human Bodies}
\subsubsection{Multi-view Video Synthesis}
Moreau et al.~\cite{moreau2023human} proposed HuGS to render photo-realistic human body avatars from multi-view videos with an animatable human body represented with a set of 3D Gaussians. 
3D Gaussians in HuGS build upon the original setting with an additional skinning weight vector that regulates the influence of each body joint on how the Gaussian moves and a latent code that encodes the non-rigid motion. 
HuGS applies LBS to deform the canonical primitives initialized by the SMPL model and learns only the skinning weights.
Since LBS only encodes the rigid deformations of the body joints, HuGS subsequently introduces a local non-rigid refinement stage to model the non-rigid deformation of the garments, considering both the body pose encoding and ambient occlusions. 
% Though achieving competitive performance on novel pose synthesis, HuGS optimizes and deforms each Gaussian independently, ignoring the intrinsic relation between Gaussians in local neighborhoods.
Based on the observation that explicit point-based representations have the potential to be parameterized on 2D maps, Li et al.~\cite{li2023animatable} facilitate Gaussian-based modeling with 2D CNNs to faithfully reconstruct higher-fidelity avatars.
After obtaining the character-specific template from multi-view images and diffusing into it the skinning weights of SMPL, this method predicts two pose-dependent canonical Gaussian maps in the UV space through StyleUNet~\cite{wang2023styleavatar}.
Specifically, given a training pose, the template is first deformed to the posed space via LBS, and two position maps are produced by orthogonal projection of the posed template to the front and back views. 
Subsequently, Gaussian maps are generated with position maps serving as the pose condition, and thus canonical Gaussians are extracted from each pixel.
% Moreover, for the generalization to novel poses, PCA is utilized to project a novel driving pose signal into the distribution of seen training poses.
ASH~\cite{pang2023ash} also attaches Gaussian splats to the deformable character model~\cite{habermann2021real} and parameterizes them in 2D texture space, with each texel covered by a triangle representing a Gaussian. 
Besides, a 2D geometry decoder and a 2D appearance decoder are adopted to separately predict the shape-related parameters and the Spherical Harmonics. 
Subsequently, Gaussian splats from the canonical position are transformed to the posed space through dual quaternion skinning~\cite{kavan2007skinning} and supervised solely on the multi-view videos.

Departing from the commonly used LBS, D3GA~\cite{zielonka2023drivable} employs cage deformation to transform Gaussians to represent deformable human avatars.
Notably, the deformation is conducted in a compositional manner where separate cages deform the body, face, and garment avatars and each part is controlled by three specialized MLPs, i.e., the cage node correction network, the Gaussian correction network, and the shading network. 

HiFi4G~\cite{jiang2023hifi4g} resorts to traditional non-rigid tracking for high-fidelity human performance rendering from multi-view videos. 
A dual-graph mechanism, comprising of a coarse deformation graph and a fine-grained Gaussian graph, is first introduced to bridge 3D Gaussian with non-rigid tracking. 
The former first generates per-frame geometry proxies using NeuS2~\cite{wang2023neus2} and then applies the Embedded Deformation (ED)~\cite{sumner2007embedded} to establish model-to-model motion correspondences. 
Then, the fine-grained Gaussian graph is generated by initializing Gaussians from the mesh of the first frame and applying pruning and densification operations to the subsequent key frames for temporal coherence. 
In the optimization step, HiFi4G employs ED graph to warp Gaussian under the constraint of a temporal regularizer, a smooth term, and an adaptive weighting mechanism. 
% To make the generated results practical for users, HiFi4G demonstrates a companion compression scheme with standard residual compensation, achieving a substantial compression rate of approximately 25 times and only requiring less than 2 MB storage per frame. 

Contrary to most methods relying on per-subject optimization, Zheng et al. \cite{zheng2023gps} presented a generalizable method called GPS-Gaussian to synthesize novel views of unseen human performers in real-time without any fine-tuning or optimization. 
GPS-Gaussian directly regresses Gaussian parameters in a feed-forward manner from massive 3D human scan data with diverse properties to learn abundant human priors, thus enabling instantaneous human appearance rendering. 
Moreover, GPS-Gaussian adopts the efficient 2D CNN to encode the source view image and predict the 2D Gaussian parameter maps. 
The depth estimation module generates a learned depth map, which represents the 3D position map. Meanwhile, the RGB image serves as the color map. These two components are used to formulate the Gaussian representation. Additionally, the remaining parameters of the 3D Gaussians are predicted in a pixel-wise manner. Later, these parameter maps are lifted to 3D space and aggregated for novel view rendering.

%\vspace{-0.3cm}
\subsubsection{Monocular Video Synthesis}
As another research line, Kocabas et al.~\cite{kocabas2023hugs} designed a deformation model to represent the animatable human and the static scene with 3D Gaussians from only a monocular video. 
% Hence, this method supports both the novel pose synthesis of the human and the novel view synthesis of the human and the scene. 
In practice, the human body and the scene are disentangled and separately constructed from the SMPL body model and the SfM point cloud. 
The human Gaussians are parameterized by their center locations in a canonical space, a feature triplane and three MLPs which predict properties of the Gaussians. 
% This method achieves state-of-the-art reconstruction quality on the NeuMan dataset and the ZJU-Mocap dataset. 
Similar to~\cite{kocabas2023hugs}, ParDy-Human~\cite{jung2023deformable} separates the human and the background with two sets of Gaussians. 
Additionally, the human Gaussian representation is extended with two extra features, parent index and surface normal, indicating the mesh face that the Gaussian is generated from. 
Thus, the canonical Gaussian can be deformed to the pose with Per Vertex Deformation (PVD) according to its parent mesh face. 
The garment motions are further refined by a deformation refinement module that predicts residual refinements for PVD deformations.
% Besides, the surface normal is updated through the deformations to obtain a more realistic direction for calculating Spherical Harmonics.
Li et al.~\cite{li2023human101} proposed Human101 for both fast reconstruction and real-time rendering from monocular videos. 
For fast convergence, Human101 first extracts the point cloud from four frames covering the front, back, left and right views of the human body and then converts the generated point cloud into canonical Gaussians. 
In the deformation process, Human101 assigns each Gaussian to its nearest SMPL triangular facet and deforms its rotation and spherical harmonics with Triangular Face Rotation accordingly.
% Such method is able to train 3D Gaussians within 100 seconds and achieve a rendering speed over 100 FPS. 

3DGS-Avatar~\cite{qian20233dgs} also initializes a set of canonical Gaussians via sampling on the SMPL mesh template.
% Following HumanNeRF~\cite{weng2022humannerf}, it decomposes the human deformation into a non-rigid part that encodes pose-dependent cloth deformation, and a rigid transformation controlled by the human skeleton.
It decomposes the human deformation into a non-rigid part that encodes pose-dependent cloth deformation, and a rigid transformation controlled by the human skeleton.

Since the view-dependent color encoding does not suit the monocular setting, 3DGS-Avatar substitutes the convention of learning spherical harmonic coefficients with a color MLP. 
In practice, it canonicalizes the viewing direction used in the color modeling and additionally learns a per-frame latent code to encode the different light effects across frames. 
% It further reports a fast real-time rendering at over 50 FPS with comparatively fast training within 30 minutes.
Instead of optimizing per-Gaussian colors, Jena et al.~\cite{jena2023splatarmor} proposed a novel neural color field to implicitly regularize the color of nearby Gaussians, which later provides an additional 3D supervision to the Gaussian positions. The visualized results prove that this operation effectively reduces the texture artifacts on test frames. 

To circumvent the inaccurate motion estimation in the monocular setting, Hu et al.~\cite{hu2023gaussianavatar} proposed GaussianAvatar to jointly optimize motion and appearance with isotropic 3D Gaussians. 
Specifically, given a fitted SMPL model, a network encodes the pose-dependent feature from a 2D UV position map and further integrates the feature with a learnable feature tensor that encodes the coarse global appearance.
Thus, the integrated feature serves as the input for the Gaussian parameter encoder to predict the Gaussian properties like position offset, color and scale.
Notably, the human avatar is represented by isotropic Gaussians with fixed rotation and opacity since anisotropic 3D Gaussians are prone to overfitting the most frequently seen view via monocular supervision.

The direct application of LBS transformation for 3D Gaussians may introduce inaccurate information derived from SMPL. 
In this regard, GauHuman~\cite{hu2023gauhuman} starts with LBS weights from SMPL and rectifies SMPL poses with an effective pose refinement module. 
GauHuman also integrates the KL divergence of 3D Gaussians to regulate the split and clone process, taking the crucial metric, the distance between 3D Gaussians, into consideration.
To speed up the optimization, a novel merge operation is introduced to merge redundant Gaussians that meet certain criteria. 
The inaccurate mapping from the canonical space to the observation space may induce overfitting and artifacts. 
To tackle this, Li et al. \cite{li2024gaussianbody} proposed to preserve the local geometry property of 3D Gaussians during the deformation with three regularization losses, i.e., the local-rigidity loss, the local-rotation loss, and the local-isometry loss~\cite{luiten2023dynamic}. 
Since the monocular video input inherently lacks multi-view supervision, a split-with-scale operation to address the potential blurs and oversized Gaussians and a pose refinement to optimize the SMPL pose parameters during training are introduced for further refinement.

SplattingAvatar~\cite{shao2024splattingavatar} proposed a hybrid avatar representation of 3D Gaussian and meshes to disentangle the human motion and appearance. Specifically, the low-frequency motion information is controlled by the mesh while the high-frequency geometry and appearance details are modeled by the Gaussians. This method associates Gaussians with mesh triangles using Phong surface~\cite{shen2020phong} embeddings and jointly optimizes Gaussians and the embedding parameters. Besides, the pose-dependent rotation and translation are explicitly defined by the mesh instead of the MLP-based linear blend skinning.
Similarly, GoMAvatar~\cite{wen2024gomavatar} introduced the Gaussian-on-Mesh representation that leverages 3D Gaussians for real-time rendering and utilizes skeleton-based deformable mesh for geometry modeling and compatibility with graphics pipelines. Contrary to the direct computation of the final color in the previous rendering methods, GoMAvatar decomposes the RGB image into the pseudo albedo map and the
pseudo shading map where the former is rendered by Gaussian splatting and the latter is derived from the normal map obtained from mesh rasterization. Following HumanNeRF~\cite{weng2022humannerf}, this method also integrates a pose refinement module to optimize the estimated poses learned from monocular videos.

%\vspace{-0.4cm}
\subsection{Human Heads}
\subsubsection{Multi-view Video Synthesis}
GaussianAvatars~\cite{qian2023gaussianavatars} focuses on the reconstruction of head avatars by rigging 3D Gaussians to a parametric morphable face model. 
In particular, 3D Gaussians are initialized at the center of each triangle of the FLAME \cite{li2017learning} mesh, with their parameters defined by the triangle properties. 
Besides, to accommodate the adaptive density control operations to the method without breaking the connection between triangles and splats, a binding inheritance strategy is designed to additionally parameterize the Gaussian with the index of its parent triangle so that new Gaussian points remain rigged to the FLAME mesh.
However, such a method lacks control over areas that are not modeled by FLAME such as hair and other accessories.
Xu et al.~\cite{xu2023gaussian} proposed a fully learnable expression-conditioned deformation field to avoid the limited capability of the LBS-based formulation. 
The proposed method involves feeding the positions of the 3D Gaussians, along with their expression coefficients, into an MLP to accurately predict the displacements between the neutral expression and the target expression.
Besides, instead of initializing Gaussians with a morphable template, it designs a geometry-guided initialization strategy by optimizing an SDF field, a color field and a deformation field to extract a mesh and construct Gaussians on the basic head surface. 

%\vspace{-0.3cm}
\subsubsection{Monocular Video Synthesis}
Dhamo et al.~\cite{dhamo2023headgas} proposed a head animation method called HeadGas that can work with different 3D morphable models from monocular video inputs. 
To create the correspondence between the specific expression and 3D Gaussians, this method extends Gaussian representation with a base of latent features, which is multiplied with an expression vector and then fed to the MLP to yield the final color and opacity. 
Since 3D Morphable Model (3DMM) based methods often fail to model non-facial structures such as eyeglasses and hairstyles, Zhao et al.~\cite{zhao2024psavatar} proposed a Point-based Morphable Shape Model (PMSM) to replace meshes with points for enhanced representation flexibility.
PMSM builds on FLAME to inherit its morphable capability.
Then, it converts the FLAME mesh to points by uniformly sampling points on the surface of the mesh. 
Additionally, points off the meshes are sampled to capture complex structures ignored by the FLAME model. 
For faster convergence of head shape alignment, the process is conducted in an analysis-by-synthesis manner with point splatting. 
Finally, 3D Gaussian is employed in combination with PMSM for effective fine-detail rendering.

Chen et al.~\cite{chen2023monogaussianavatar} adopted a different approach to deform the Gaussian representation from monocular videos by transforming the mean position of Gaussians from the canonical space to the deformed space with LBS and adjusting other Gaussian parameters correspondingly with a deformation field. 
GaussianHead~\cite{wang2023gaussianhead} disentangles the modeling of the head geometry and texture attributes by utilizing a motion deformation field to fit the head shape as well as dynamic facial movements and a multi-resolution tri-plane to store the appearance information of Gaussians.
To resolve the feature dilution, this monocular-video-based method devises a Gaussian derivation strategy to generate multiple doppelgangers of each core Gaussian in the canonical space through learnable rotation transformations and then integrates sub-features of these doppelgangers to form the final canonical feature for the core Gaussian.

% \subsection{Challenges and Opportunities}
\textbf{Challenges:}
(i) Existing human avatar methods rely on deformable models like SMPL, SMPL-X and FLAME for Gaussian initialization. 
(ii) Unlike carefully captured videos in the universal datasets, in-the-wild videos are relatively coarse and more challenging to reconstruct. 
(iii) The environmental lighting is not parameterized in most methods, which makes relighting the avatars infeasible. Also, the uneven illumination in the video may degrade the synthesis quality.
(iv) The intrinsic structure and connectivity relation between Gaussians in local regions are then ignored.

\textbf{Opportunities:}
(i) The reconstruction performance of both the 3DMM-based methods and SMPL-based methods is constrained by the initialization of model parameters. 
The inaccuracy in the fixed parameters may severely affect the model’s alignment with the supervisions, thus leading to blurred texture. 
In this regard, enhancing the expression ability of the template model during the optimization is a promising breakthrough.
(ii) For human head modeling, methods that utilize 3DMM to control the motions also fail to express subtle facial expressions. 
Exploring a more effective method to separately control the non-rigid deformation is a focus of future work.
% (ii) Explore a more effective method to separately control the non-rigid deformation.
% The reconstruction performance of both the 3DMM-based methods and SMPL-based methods is constrained by the initialization of model parameters and the representation ability of the mesh template. 
% On the one hand, The inaccuracy in the fixed parameters may severely affect the model’s alignment with the supervisions, thus leading to blurred texture. 
% On the other hand, models lack of expressiveness in loose areas and intricate details. For example, methods that utilize 3DMMs to control the motions fail to express subtle facial expressions.
% In this regard, the first solution is to enhance the expression ability of the template model during the optimization.
% Second, exploring a more effective method to separately control the non-rigid deformation can be a focus of future work. 
% Many works have resorted to disentangling the motion of human bodies and the garments for separate representations. 
% Besides, Moreau et al.~\cite{moreau2023human} point out that the modeling of loose garments may also be improved by incorporating more physics priors. 
% }
% \textcolor{blue}{
% Besides, how to extract meshes from learned 3D Gaussians remains a future work to be investigated.}

%\vspace{-0.4cm}
\section{Discussions and Future Work}\label{sec:Discussion}

3D Gaussian Splatting has demonstrated significant potential in the fields of computer graphics and computer vision. 
Nevertheless, various challenges persist due to the intricate structures and diverse tasks associated with 3D Gaussian Splatting. This section aims to address these challenges and present potential avenues for future research.

\textbf{Handling Floating Elements in 3D-GS.}
% A notable issue in 3D Gaussian Splatting is the prevalence of floating elements within the rendered space, primarily originating from image backgrounds. Employing opacity thresholds has been suggested to reduce the occurrence of these floats, enhancing image rendering quality as measured by PSNR and SSIM metrics~\cite{fan2023lightgaussian}. These floating elements significantly impact the visual quality of the rendered images. A potential area of research could focus on strategies to anchor these floats closer to surfaces, thereby enhancing their positional relevance and contribution to image quality.
One significant challenge in 3D Gaussian Splatting pertains to the presence of floating objects in the rendered space, which mainly arise from image backgrounds. To mitigate the occurrence of these floats and improve the quality of image rendering, the utilization of opacity thresholds has been proposed, as demonstrated by the enhancement in image quality measured through metrics such as PSNR and SSIM~\cite{fan2023lightgaussian}. The presence of these floating elements considerably affects the visual fidelity of the rendered images. To address this issue, a potential avenue for further investigation could center around exploring strategies to anchor these floats in closer proximity to surfaces, thereby enhancing their positional relevance and overall contribution to image quality. There are two primary research directions that guide the movement of these floats towards the surfaces: (i) Geometry-guided rendering and (ii) Depth-guided rendering.

\textbf{Trade-off Between Rendering and Reconstruction.}
% The presence of floating elements significantly influences the visual quality of images. 
% However, their impact extends beyond rendering to affect mesh extraction processes. 
% On one hand, directly utilizing 3D-GS can reconstruct meshes efficiently but the meshes tend to be not continuous. 
% For instance, SuGaR~\cite{guedon2023sugar} utilizes an opacity-based approach to generate 3D Gaussians around the mesh surface for reconstruction and compromise rendering quality.
% Similarly, 2D-GS collapses the 3D volume into a set of 2D-oriented planar Gaussian disks to enhance the quality of the reconstructions.
On one hand, the direct utilization of 3D-GS allows for efficient mesh reconstruction. However, it often results in non-continuous meshes. For example, SuGaR~\cite{guedon2023sugar}) employs an opacity-based approach to generate 3D Gaussians around the mesh surface, compromising the rendering quality. Similarly, 2D-GS~\cite{huang20242d} compresses the 3D volume into a collection of planar Gaussian disks oriented in 2D, aiming to improve the quality of the reconstructions.
On the other hand, when 3D-GS is combined with other representations, like signed distance fields, it yields smoother surfaces; however, this comes at the cost of increased reconstruction time. An illustration of this is the optimization of an implicit signed distance field in conjunction with 3D-GS as demonstrated by 3DGSR~\cite{lyu20243dgsr}, aiming to enhance the reconstruction process.
This highlights the importance of adopting a nuanced approach that balances the pursuit of rendering quality with the accurate reconstruction and speed of the process.
Hence, there are two potential directions for further research in this regard. The first involves building upon the existing frameworks of 2D-GS or 3D-GS for reconstruction, utilizing solely one representation. The second entails exploring how 3D-GS could enhance or complement other time-efficient representations, thus presenting a promising area~\cite{chen2023neusg}.

% \subsection{Optimization of 3D Gaussian Splatting}

% \subsubsection{Rendering Quality}

\textbf{Rendering Realism.}
Current lighting decomposition methods show limited effectiveness in scenes with indistinct boundaries, often requiring the inclusion of object masks during optimization. 
This limitation primarily arises from the detrimental influence of the background on the optimization process, a consequence of the distinctive qualities of the point clouds generated through 3D Gaussian Splatting. Unlike conventional surface points, these point clouds display particle-like properties, including color and partial transparency, deviating from conventional surface points. 
Considering these challenges, integrating Multi-View Stereo into the optimization process could significantly improve geometry accuracy.
Furthermore, specular and anisotropic effects are primarily influenced by the properties of the material, while reflections are closely tied to the environment and geometry. It is essential to consider these factors as well.

\textbf{Real-time Rendering.}
To facilitate real-time rendering, Scaffold-GS~\cite{lu2023scaffold} introduces anchor points from a sparse grid~\cite{xu2023grid}, which aids in distributing local 3D Gaussians and thus improves rendering speed. However, the method's reliance on a uniform grid size limits its adaptability. 
The utilization of octree representations has emerged as a promising alternative, providing the flexibility to divide complex regions into smaller grids for detailed processing~\cite{ren2024octree}. However, the usage of octree in Octree-GS still needs to be manually assigned. Consequently, there is an urgent need to devise adaptive octree representations.
While these methods show potential for achieving real-time rendering in small scenes, scaling up to large environments, such as cityscapes~\cite{xu2023grid}, will require further efforts.

\textbf{Few-short 3D-GS.}
Recent few-shot studies~\cite{zhu2023fsgs, chung2023depth} explore optimizing Gaussian splatting with depth guidance in a few-shot setting. While promising, these approaches face notable challenges. 
The success of few-shot methods heavily relies on the accuracy of monocular depth estimation models. Moreover, their performance may vary across different data domains. Additionally, the dependency on fitting the estimated depth to COLMAP~\cite{schonberger2016structure} points introduces reliance on the performance of COLMAP itself. Consequently, these limitations pose challenges in handling textureless areas or complex surfaces where COLMAP~\cite{schonberger2016structure} may encounter difficulties. 
For future research, it would be beneficial to investigate the optimization of 3D scenes using interdependent depth estimates, reducing reliance on COLMAP points. Another avenue for future work involves investigating methods to regularize geometry across diverse datasets, particularly in areas where depth estimation, such as the sky, presents challenges.

\textbf{Intergration of Physics.}
In contrast to the natural world, where the physical behavior and visual appearance of materials are inherently interconnected, the traditional physics-based visual content generation pipeline has been a laborious and multi-stage process. This process involves constructing the geometry, preparing it for simulation (often using techniques like tetrahedralization), simulating the physics, and eventually rendering the scene. Although this sequence is effective, it introduces intermediate stages that can result in discrepancies between the simulation and the final visualization. This disparity is also evident within the NeRF paradigm~\cite{li2023pac,chu2022physics}, where the rendering geometry is embedded within a simulation geometry. 
To address this issue, it is recommended to unite these two aspects by advocating for a unified representation of material substance that can be utilized for both simulation and rendering purposes. In addition, it is essential to conduct further research on large and complex motion as well as the utilization of multiple materials for various objects. Additionally, an area worth exploring is the automatic assignment of materials to 3D-GS, which shows potential for future development.

%物理约束在神经渲染中体现。

\textbf{Accurate Reconstruction.}
The original 3D-GS cannot distinguish between specular and non-specular areas, thereby unable to produce reasonable 3D Gaussians in regions with specular reflections~\cite{jiang2023gaussianshader, gao2023relightable}. 
The presence of irrational 3D Gaussians can significantly influence the reconstruction, leading to the generation of flawed meshes. Furthermore, it has been observed that the inclusion of specular reflection components can also result in the production of unreliable meshes~\cite{guedon2023sugar}. 
Hence, to achieve precise reconstruction, it is essential to decompose the 3D Gaussian through illumination prior to accurately reconstructing the mesh.
Otherwise, another future direction involves large-scale reconstruction, such as VastGaussian~\cite{lin2024vastgaussian}. However, the presence of numerous 3D Gaussians in a massive scene may pose challenges in terms of storage space and rendering speed. Consequently, a promising direction is to develop techniques that can effectively handle 3D Gaussians for large-scale reconstruction, while simultaneously addressing the issues of storage space and memory requirements.

\textbf{Realistic Generation.}
Pioneered by DreamGaussian~\cite{tang2023dreamgaussian}, GaussianDreamer~\cite{yi2023gaussiandreamer}, 3D-GS begins its journey on 3D generation.
However, the geometries and textures of the generated 3D assets still require improvement.
In terms of geometries, the integration of more precise SDF and UDF into the 3D-GS allows for the generation of meshes that are more realistic and accurate. Additionally, various conventional graphic techniques, such as Medial Fields~\cite{rebain2021deep}, can be effectively utilized.
Regarding textures, two recently proposed methods, MVD~\cite{liu2023text} and TexFusion~\cite{cao2023texfusion}, demonstrate impressive capabilities in texture generation. These advancements have the potential to be applied in the context of textured mesh generation with 3D-GS. Additionally, Relightable3DGaussian~\cite{gao2023relightable} and GaussianShader~\cite{jiang2023gaussianshader} have explored the shading aspects of 3D-GS. However, the question of BRDF decomposition on the generated meshes remains unanswered.
Additionally, the exploration of 4D generation holds great promise, requiring further enhancement of temporal consistency.

% \subsection{Versatile Manipulation}

% \textbf{Expanding 3D-GS with
\textbf{Combining 3D-GS with Large Foundation Models.}
Recent studies, such as Shi et al. ~\cite{shi2023language}, have demonstrated that embedding language into 3D-GS can significantly enhance 3D scene understanding.
With the advent of large foundation models in 2023, their remarkable capabilities have been showcased across a broad spectrum of vision tasks. Notably, the SAM model has emerged as a powerful tool for segmentation, finding successful applications within the 3D-GS~\cite{ye2023gaussian,zhou2023feature,cen2023segment}. Beyond segmentation, LLM models have shown promise for language-guided generation, manipulation, and perception tasks. This highlights the versatility and utility of these models in a wide range of applications, further emphasizing their significance in 3D-GS.
Recent studies~\cite{shi2023language} have demonstrated that embedding language into 3D-GS can significantly enhance 3D scene understanding.
With the advent of large foundation models in 2023, their remarkable capabilities have been showcased across a broad spectrum of vision tasks. Notably, the SAM model has emerged as a powerful tool for segmentation, finding successful applications within the 3D-GS~\cite{ye2023gaussian,zhou2023feature,cen2023segment}. Beyond segmentation, LLM models have shown promise for language-guided generation, manipulation, and perception tasks. This highlights the versatility and utility of these models in a wide range of applications, further emphasizing their significance in 3D-GS.

Several works employ 3D-GS as a supplementary tool to improve performance. For instance, NeuSG~\cite{chen2023neusg} utilizes 3D-GS to enhance the reconstruction of NeuS, while SpecNeRF~\cite{ma2023specnerf} incorporates Gaussian directional encoding to model specular reflections. Consequently, the distinctive characteristics of 3D-GS can be seamlessly integrated into existing methods to further boost their performance. It is conceivable that 3D-GS can be combined with the Large Reconstruction Model (LRM)~\cite{xu2023dmv3d}, or with existing perception techniques in Autonomous Vehicles, to enhance their perception capabilities.

\ifCLASSOPTIONcaptionsoff
  \newpage
\fi

\bibliographystyle{IEEEtran}
\bibliography{ref}

% that's all folks
\end{document}

\title{Supplementary Materials for \\ 3D Gaussian Splatting as a New Era: A Survey}

\author{Ben Fei$^*$, Jingyi Xu$^*$, Rui Zhang$^*$, Qingyuan Zhou$^*$, Weidong Yang$^{\dagger}$, and Ying He$^{\dagger}$% <-this % stops a space
% \IEEEcompsocitemizethanks{\IEEEcompsocthanksitem 
% \textit{Ben Fei, Jingyi Xu, Rui Zhang, and Qingyuan Zhou contribute equally to this work and are co-first authors.
% }
% \IEEEcompsocthanksitem \textit{Corresponding authors: Weidong Yang and Ying He}
% \IEEEcompsocthanksitem Ben Fei, Jingyi Xu, Rui Zhang, Qingyuan Zhou, and Weidong Yang are with the School of Computer Science, Fudan University, Shanghai, China, 200433 (e-mails: $\{\text{bfei21}|\text{jyxu22}|\text{22210240379}|\text{zhouqy23}\}$@m.fudan.edu.cn;  wdyang@fudan.edu.cn).
% \IEEEcompsocthanksitem Ying He is with the School of Computer Science and Engineering, Nanyang Technological University, Singapore, 639798 (email: yhe@ntu.edu.sg).
% }
}

\maketitle

\section{Applications}

\subsection{Robotics}

Among the applications of 3D-GS in the field of robotics~\cite{yan2023gs}, one of the most prevalent uses is in SLAM robot navigation~\cite{lei2024gaussnav} and environmental modeling~\cite{deng2024compact}, while 3D-GS is extensively employed in the point cloud processing and reconstruction processes of SLAM systems. 
In SLAM, robots utilize various sensors, such as LiDAR and cameras, to perceive their surroundings and estimate their position while constructing an environment map through the integration of this perceptual data with previous measurements. 
3D-GS plays a significant role in this process. 
By applying 3D-GS, the SLAM system can achieve a denser and more continuous map representation~\cite{yan2023gs}, ultimately improving the accuracy of robot localization and map construction. This map representation can also be utilized for tasks such as path planning~\cite{jiang20243dgs}, obstacle avoidance~\cite{chen2024splat}, and object detection~\cite{tosi2024nerfs}, thereby enhancing the robot's navigation capabilities~\cite{lu2024manigaussian}.

In addition to SLAM, 3D-GS can also play a role in other robotic applications. For instance, it can be used for object recognition and reconstruction to aid the robot in understanding its surrounding environment and performing object manipulation tasks. By processing sensor data with 3D-GS, the robot can generate high-resolution object models, enabling pose estimation~\cite{chen2024splat}, object tracking~\cite{sun2024high}, and pick-and-place operations~\cite{zheng2024gaussiangrasper}.

\subsection{Autonomous Navigations}

The study conducted in DrivingGaussian~\cite{zhou2023drivinggaussian} provides evidence that 3D-GS is applicable for surrounding view synthesis, dynamic scene reconstruction, and corner case simulation. Additionally, StreetGaussians~\cite{yan2024street} demonstrates the ability of 3D-GS to effectively model dynamic urban scenes.

The future research direction of 3D-GS in the field of autonomous driving holds great promise. 
Firstly, there is potential for further optimization of the 3D-GS algorithm to enhance its accuracy and stability in environmental perception and scene modeling~\cite{zhao2024tclc}.
This optimization will contribute to the development of more reliable and precise autonomous driving systems. 
Secondly, the integration of 3D-GS with other visual perception technologies, such as deep learning and sensor fusion, can be explored to enhance the performance of autonomous driving systems~\cite{li2024ho}. 
By combining different perception modules, a more comprehensive and robust environmental understanding and decision-making capability can be achieved~\cite{cui2024letsgo}. 
Moreover, it is important to investigate the application of 3D-GS in real-time decision-making and planning within autonomous driving scenarios. 
By harnessing its high-precision environmental perception capabilities, the system can make more intelligent and secure driving decisions, thereby improving its adaptability and ability to handle complex traffic environments. Lastly, practical applications of 3D-GS in autonomous driving systems, including automatic vehicle parking and interaction between autonomous vehicles and pedestrians, warrant exploration. 
By examining the feasibility and effectiveness of applying 3D-GS technology in real traffic environments, its potential can be further validated.

\subsection{Humans}
The potential of 3D-GS in fields like human medicine~\cite{liu2024endogaussian} and sports analysis~\cite{li2023animatable} is significant.
In the medical field, this technique can be utilized for visualizing and analyzing medical images. 
By integrating medical image data with 3D-GS, internal organs, bones, and muscles can be realistically presented~\cite{zhu2024deformable}. 
This will greatly assist doctors in accurately diagnosing diseases and planning surgeries, ultimately enhancing medical care and improving patient treatment outcomes~\cite{zhu2024deformable}. 
Furthermore, 3D-GS also holds great promise in motion analysis. Through the use of multiple sensors to capture human motion data and incorporating 3D-GS for data interpolation and reconstruction, precise analysis and modeling of human motion can be achieved~\cite{cotton2024dynamic}. 
This has wide-ranging applications in areas such as sports training, posture correction, and human-computer interaction. For instance, in sports training, coaches can leverage 3D-GS technology to analyze athletes' movements and postures, offering personalized guidance and improvement suggestions to enhance their skill levels.

The applications of 3D-GS in the human body are extensive. From medical imaging to sports analysis, this technology offers more accurate, realistic, and practical solutions across various fields. With ongoing technological advancements and application developments, we can anticipate the emergence of innovative applications based on 3D-GS, which will positively impact our lives and work.

% \input{Tables/optimization_sum}
% Please add the following required packages to your document preamble:
% \usepackage{graphicx}
\begin{table*}[htbp]
\centering
\setlength\tabcolsep{1.5pt}
\renewcommand{\arraystretch}{1.4}
\caption{Summary of methods derived from 3D-GS for improving the representation (Part I).}
\resizebox{\textwidth}{!}{%
\begin{tabular}{>{\columncolor{C7!50}}l>{\columncolor{C6!50}}c>{\columncolor{C5!50}}c>{\columncolor{C4!50}}c>{\columncolor{C3!50}}c>{\columncolor{C2!50}}c>{\columncolor{C1!50}}c}
\toprule[1.5pt]
Methods   & Venue & Year & Category &  Sub-category  & Contributions                                                                                                                         & Limitations                                                                       \\ \midrule[1pt]

Scaffold-GS~\cite{lu2023scaffold}  & CVPR & 2024  & Representation  & Efficiency &  \begin{tabular}[c]{@{}c@{}} (1) Utilize underlying scene structure to help prune excessively expanded Gaussian balls.  \\ (2) Predict Gaussian attributes on-the-fly for robust novel view sythesis. \end{tabular} & \begin{tabular}[c]{@{}c@{}} Initialization of the framework suffers from extremely sparse points. \end{tabular}\\ 

Compact3D~\cite{navaneet2023compact3d} & Arxiv & 2023  & Representation & Efficiency &  \begin{tabular}[c]{@{}c@{}} Adopt K-means and vector quantization to compress the parameters of 3D Gaussians. \end{tabular} & \begin{tabular}[c]{@{}c@{}} / \end{tabular}\\ 

EAGLES~\cite{girish2023eagles} & Arxiv & 2023  & Representation & Efficiency &  \begin{tabular}[c]{@{}c@{}} Utilize quantization for parameter compression and a coarse-to-fine training strategy \\ for gradually increasing the rendering size to improve memory efficiency and speed. \end{tabular} & \begin{tabular}[c]{@{}c@{}} / \end{tabular}\\ 

LightGaussian~\cite{fan2023lightgaussian} & Arxiv & 2023  & Representation & Efficiency &  \begin{tabular}[c]{@{}c@{}} (1) Evaluate the global significance of each Gaussian for pruning. \\ (2) Coefficients of trivial Gaussians are quantized to save storage space. \end{tabular} & \begin{tabular}[c]{@{}c@{}} / \end{tabular}\\ 

\begin{tabular}[c]{@{}c@{}}Efficient Representation \\for Dynamic Scenes~\cite{katsumata2023efficient}\end{tabular}   & Arxiv & 2023  & Representation & Efficiency &  \begin{tabular}[c]{@{}c@{}} (1) Parameters are divided according to their time-invariance \\ and only time-variant parameters are stored. \\
(2) The flow information is employed to overcome the ambiguity between \\ consecutive frames across different time steps. \end{tabular} & \begin{tabular}[c]{@{}c@{}} Limited in modeling the change of topology and fluids. \end{tabular}\\ 

Multi-Scale 3D-GS~\cite{yan2023multi} & CVPR & 2024  & Representation & Photorealism &  \begin{tabular}[c]{@{}c@{}} (1) Represent the scene at different levels of detail. \\ (2) 3D Gaussians with appropriate scales are selected accordingly to improve the quality. \end{tabular} & \begin{tabular}[c]{@{}c@{}} Rely on the pixel coverage and the calculation is a considerable overhead. \end{tabular}\\ 

Mip-Splatting~\cite{yu2023mip} & CVPR & 2024  & Representation & Photorealism  &  \begin{tabular}[c]{@{}c@{}} (1) The 3D-smoothing filter removes high-frequency artifacts. \\ (2) The 2D filter mitigates the aliasing issues when rendering at a lower sampling rate. \end{tabular} & \begin{tabular}[c]{@{}c@{}} The approximation of a box filter introduces errors. \end{tabular}\\ 

RelightableGaussian~\cite{gao2023relightable} & Arxiv & 2023  & Representation & Photorealism &  \begin{tabular}[c]{@{}c@{}} (1) Rendering pipeline solely based on 3D-GS. \\ (2) Split the incident light into local and global parts. \end{tabular} & \begin{tabular}[c]{@{}c@{}} Reduced performance in unbordered scenes. \end{tabular}\\ 

GaussianShader~\cite{jiang2023gaussianshader} & CVPR & 2024  & Representation & Photorealism  &  \begin{tabular}[c]{@{}c@{}} (1) Effective approximation of the rendering equation. \\ (2) Novel framework for normal estimation. \end{tabular} & \begin{tabular}[c]{@{}c@{}} Training and rendering time could be further reduced. \end{tabular}\\ 

GS-IR~\cite{liang2023gs} & CVPR & 2024  & Representation & Photorealism  &  \begin{tabular}[c]{@{}c@{}} (1) Leverage forward mapping splatting to deduce the physical attributes of a complex scene. \\ (2) Utilize baking-based method for efficient indirect illumination modeling. \end{tabular} & \begin{tabular}[c]{@{}c@{}} The quality of normal reconstruction. \end{tabular}\\ 

SpecNeRF~\cite{ma2023specnerf} & CVPR & 2024  & Representation & Photorealism &  \begin{tabular}[c]{@{}c@{}} (1) Specifically designed Gaussian directional encoding \\ for spatially varying lighting conditions. \\ (2) Utilization of monocular normal estimation in early training stages. \end{tabular} & \begin{tabular}[c]{@{}c@{}} Blurry results when rendering mirror-like reflections. \end{tabular}\\ 

Depth Regularized~\cite{chung2023depth} & Arxiv & 2023  & Representation & Costs &  \begin{tabular}[c]{@{}c@{}} Depth-guided optimization and early stop training strategy. \end{tabular} & \begin{tabular}[c]{@{}c@{}} (1) Heavily rely on monocular depth estimation. \\ (2) Fall short in scenarios with textureless plains or changing surfaces. \end{tabular}\\ 

FSGS~\cite{zhu2023fsgs} & Arxiv & 2023  & Representation  & Costs &  \begin{tabular}[c]{@{}c@{}} (1) Densify Gaussians by proximity-guided Gaussian unpooling. \\ (2) Employs monocular priors to mitigate overfitting. \end{tabular} & \begin{tabular}[c]{@{}c@{}} / \end{tabular}\\ 

SparseGS~\cite{xiong2023sparsegs} & Arxiv & 2023  & Representation  & Costs &  \begin{tabular}[c]{@{}c@{}} (1) Utilize a generative diffusion model to compensate the absence of images of uncovered regions. \\ (2) Devise a pruning strategy to remove ``floater'' artifacts. \end{tabular} & \begin{tabular}[c]{@{}c@{}} Heavily reliant on the quality of initial point clouds. \end{tabular}\\ 

4D-GS~\cite{wu20234d} & CVPR & 2024  & Representation & Physics &  \begin{tabular}[c]{@{}c@{}} (1) Incorporate efficient Gaussian deformation field. \\ (2) A spatial-temporal encoder for effective feature extraction. \end{tabular} & \begin{tabular}[c]{@{}c@{}} Suffer from large motions, the absence of background points\\ and the unprecise camera pose. \end{tabular}\\ 

MD-Splatting~\cite{duisterhof2023md} & Arxiv & 2023  & Representation & Physics &  \begin{tabular}[c]{@{}c@{}} (1) Learn continuous metric deformations from a canonical space of Gaussians. \\ (2) Employ rigidity, isometry, and conservation of momentum losses to regularize the trajectory. \end{tabular} & \begin{tabular}[c]{@{}c@{}} Require synchronization of multiple calibrated cameras. \end{tabular}\\ 

DynMF~\cite{kratimenos2023dynmf} & Arxiv & 2023  & Representation & Physics &  \begin{tabular}[c]{@{}c@{}} (1) Decompose complex motions within a given scene into a few basis trajectories. \\ (2) Sharing neural basis leads to plausible and consistent dynamics. \end{tabular} & \begin{tabular}[c]{@{}c@{}} / \end{tabular}\\

Real-Time 4D-GS~\cite{yang2023real} & Arxiv & 2023  & Representation & Physics &  \begin{tabular}[c]{@{}c@{}} (1) Model space and time as an entirety to address \\ general dynamic scene representation and rendering. \\ (2) Extend the scaling and rotation matrix to 4D Euclidean space. \end{tabular} & \begin{tabular}[c]{@{}c@{}} / \end{tabular}\\ 

PhysGaussian~\cite{xie2023physgaussian} & CVPR & 2024  & Representation & Physics &  \begin{tabular}[c]{@{}c@{}} (1) Integration of continuum mechanics-based strategy for 3D Gaussian Kinematics. \\ (2) Efficient simulation and rendering pipeline. \end{tabular} & \begin{tabular}[c]{@{}c@{}} Simulating dynamics of shadows and other materials like liquids. \end{tabular}\\ 

Spacetime~\cite{li2023spacetime} & CVPR & 2024  & Representation & Physics &  \begin{tabular}[c]{@{}c@{}} (1) A Spacetime Gaussian derived representation for high-fidelity and efficient dynamic view synthesis. \\ (2) Propose splatted feature rendering to enhance compactness and improve the modeling of time-varying appearance. \\
(3) Introduce a guided sampling to enhance rendering quality at distant sparsely covered areas.\end{tabular} & \begin{tabular}[c]{@{}c@{}} Adapt to monocular setting. \end{tabular}\\ 

FreGS~\cite{zhang2024fregs} & CVPR & 2024  & Representation & Photorealism  &  \begin{tabular}[c]{@{}c@{}} Progressive frequency regularization for exploitation of low-to-high frequency components.  \end{tabular} & \begin{tabular}[c]{@{}c@{}} / \end{tabular}\\ 

DNGaussian~\cite{li2024dngaussian} & CVPR & 2024  & Representation & Photorealism &  \begin{tabular}[c]{@{}c@{}} (1) Restore accurate scene geometry via Hard and Soft regularization. \\ (2) Enhance local depth changes through Global Local Normalization. \end{tabular} & \begin{tabular}[c]{@{}c@{}} Performance drops when input views exceeds certain amount. \end{tabular}\\ 

OptimalGS~\cite{huang2024optimalgaussian} & Arxiv & 2024  & Representation & Photorealism &  \begin{tabular}[c]{@{}c@{}} (1) Thorough analysis of error that could lead to artifacts and derive mathematical expectation for it. \\ (2) Propose optimal projection method for minimization of projection error. \end{tabular} & \begin{tabular}[c]{@{}c@{}} Increased training time. \end{tabular}\\ 

VR-GS~\cite{jiang2024vr} & Arxiv & 2024  & Representation & Physics &  \begin{tabular}[c]{@{}c@{}} (1) Unified framework for rendering and simulation with interactive Gaussian editing. \\ (2) Utilize extended-position-based dynamics for real-time simulation. \end{tabular} & \begin{tabular}[c]{@{}c@{}} (1) Restricted rendering efficiency for high-fidelity scenes. \\ (2) Lack of variety of materials. \end{tabular}\\ 

GSP~\cite{feng2024gaussian} & Arxiv & 2024  & Representation & Physics &  \begin{tabular}[c]{@{}c@{}} (1) Utilize Poisson disk sampling for scene discretization
. \\ (2) Leverage position-based fluids to synthesize fluids-solids interactions
. \end{tabular} & \begin{tabular}[c]{@{}c@{}} Dynamics of fluids are less physically convincing. \end{tabular}\\ 

LIV-GaussMap~\cite{hong2024liv} & Arxiv & 2024  & Representation & Photorealism &  \begin{tabular}[c]{@{}c@{}} (1) Realize accurate representation via Gaussian distribution measurement. \\ (2) Adopt differentiable ellipsoidal surface Gaussians for more evenly distributed point clouds. \end{tabular} & \begin{tabular}[c]{@{}c@{}} / \end{tabular}\\ 

4DGS~\cite{duan20244d} & Arxiv & 2024  & Representation & Physics &  \begin{tabular}[c]{@{}c@{}} (1) Achieve unified and generalized representation of 3D \\ scene dynamics via rotor-based 4D Gaussian. \\ (2) Introduce entropy and 4D consistency losses for more realistic and consistent reconstruction. \end{tabular} & \begin{tabular}[c]{@{}c@{}} / \end{tabular}\\ 

GES~\cite{hamdi2024ges} & Arxiv & 2024  & Representation & Efficiency &  \begin{tabular}[c]{@{}c@{}} Derive Generalized Exponential Function for more memory efficient splatting. \end{tabular} & \begin{tabular}[c]{@{}c@{}} Degradation of quality when facing complex scenes. \end{tabular}\\ 

GaussianPro~\cite{cheng2024gaussianpro} & Arxiv & 2024  & Representation & Photorealism &  \begin{tabular}[c]{@{}c@{}} (1) Joint geometric representation that comprises 3D Gaussians, 2D depth and normal maps. \\ (2) Gradual densification Gaussians based on propagated depth and normal values. \end{tabular} & \begin{tabular}[c]{@{}c@{}} Diversity of experiment datasets. \end{tabular}\\ 

Spec-Gaussian~\cite{yang2024spec} & Arxiv & 2024  & Representation & Photorealism &  \begin{tabular}[c]{@{}c@{}} Leverage anisotropic spherical Gaussian and anchor-based \\ geometry-aware hybrid model to improve rendering quality. \end{tabular} & \begin{tabular}[c]{@{}c@{}} Rendering scenes with reflections. \end{tabular}\\ 

X-Gaussian~\cite{cai2024radiative} & Arxiv & 2024  & Representation & Photorealism &  \begin{tabular}[c]{@{}c@{}} Initialization strategy that leverages parameters of \\ X-ray scanner for adapting 3D Gaussians to X-ray rendering. \end{tabular} & \begin{tabular}[c]{@{}c@{}} / \end{tabular}\\ 

BAGS~\cite{peng2024bags} & Arxiv & 2024  & Representation & Photorealism &  \begin{tabular}[c]{@{}c@{}} Neural network for image de-blurring and predicting of blur regions. \end{tabular} & \begin{tabular}[c]{@{}c@{}} Extra computational burden. \end{tabular}\\ 

ManiGaussian~\cite{lu2024manigaussian} & Arxiv & 2024  & Representation & Physics &  \begin{tabular}[c]{@{}c@{}} Utilize 4D Gaussians to represent future dynamics of scenes and facilitates robotic manipulation. \end{tabular} & \begin{tabular}[c]{@{}c@{}} Heavily rely on calibrated cameras and consistent multi-views. \end{tabular}\\ 

BAD-Gaussians~\cite{zhao2024bad} & Arxiv & 2024  & Representation & Photorealism. &  \begin{tabular}[c]{@{}c@{}} Leverage motion-blurred image to reconstruct de-blurred \\ scenes via joint optimization of 3D Gaussians and camera trajectories. \end{tabular} & \begin{tabular}[c]{@{}c@{}} / \end{tabular}\\ 

RAIN-GS~\cite{jung2024relaxing} & Arxiv & 2024  & Representation & Photorealism &  \begin{tabular}[c]{@{}c@{}}  Comprehensive analysis of the importance of point clouds initialization and \\ loosen the requirement for dense and accurate initialization. \end{tabular} & \begin{tabular}[c]{@{}c@{}} / \end{tabular}\\ 

Touch-GS~\cite{swann2024touch} & Arxiv & 2024  & Representation & Costs &  \begin{tabular}[c]{@{}c@{}} (1) Jointly utilize monocular depth priors and tactile data-informed \\ implicit surface to reduce the required views for training. \\ (2) Realize visual-tactile fusion and uncertainty estimation via Bayesian Inference. \end{tabular} & \begin{tabular}[c]{@{}c@{}} Limited to static scenes. \end{tabular}\\ 

GGRt~\cite{li2024ggrt} & Arxiv & 2024  & Representation & Photorealism &  \begin{tabular}[c]{@{}c@{}} (1) Robust and iterative pose estimation that mitigates the dependency on accurate camera poses. \\ (2) (2) Propose DBP mechanism for efficient training on high-resolution datasets. \end{tabular} & \begin{tabular}[c]{@{}c@{}} / \end{tabular}\\ 

FDGaussian~\cite{feng2024fdgaussian} & Arxiv & 2024  & Representation & Costs &  \begin{tabular}[c]{@{}c@{}} (1) Two-stage pipeline that utilized diffusion model to \\ synthesize novel views based on single input image. \\ (2) Epipolar attention for enhancing multi-view feature consistency. \end{tabular} & \begin{tabular}[c]{@{}c@{}} (1) Fixed number of generated views. \\ (2) Restricted to single-object 3D reconstruction.  \end{tabular}\\ 

GS on The Move~\cite{seiskari2024gaussian} & Arxiv & 2024  & Representation & Photorealism &  \begin{tabular}[c]{@{}c@{}} (1) Capture motion effects within pixel plane for screen space approximation. \\ (2) Rasterization with approximated pixel-wise velocity. \end{tabular} & \begin{tabular}[c]{@{}c@{}} Comparison with more baseline methods. \end{tabular}\\ 

RadSplat~\cite{seiskari2024gaussian} & Arxiv & 2024  & Representation & Efficiency &  \begin{tabular}[c]{@{}c@{}} (1) Leverage radiance field prior to optimize 3D-GS for complex scenes. \\ (2) Prune strategy based on ray contribution. \end{tabular} & \begin{tabular}[c]{@{}c@{}} Prolonged training process.  \end{tabular}\\ 

HAC~\cite{chen2024hac} & Arxiv & 2024  & Representation & Efficiency &  \begin{tabular}[c]{@{}c@{}} (1) Context modeling via hash-grid which exploits consistencies among 3D Gaussians. \\ (2) Adaptive quantization of attributes. \end{tabular} & \begin{tabular}[c]{@{}c@{}} Increased training time. \end{tabular}\\

\bottomrule[1.5pt]
\end{tabular}%
}
\label{summary_representation}
\end{table*}

% Please add the following required packages to your document preamble:
% \usepackage{graphicx}
\begin{table*}[htbp]
\centering
\setlength\tabcolsep{1.5pt}
\renewcommand{\arraystretch}{1.5}
\caption{Summary of methods derived from 3D-GS for improving the representation (Part II).}
\resizebox{\textwidth}{!}{%
\begin{tabular}{>{\columncolor{C7!50}}l>{\columncolor{C6!50}}c>{\columncolor{C5!50}}c>{\columncolor{C4!50}}c>{\columncolor{C3!50}}c>{\columncolor{C2!50}}c>{\columncolor{C1!50}}c}
\toprule[1.5pt]
Methods   & Venue & Year & Category &  Sub-category  & Contributions                                                                                                                         & Limitations                                                                       \\ \midrule[1pt]

MVSplat~\cite{chen2024mvsplat} & Arxiv & 2024  & Representation & Efficiency &  \begin{tabular}[c]{@{}c@{}} Per-view cost volume representation. \end{tabular} & \begin{tabular}[c]{@{}c@{}} Inaccurate results for objects with reflections. \end{tabular}\\ 

Mini-Splatting~\cite{fang2024mini} & Arxiv & 2024  & Representation & Efficiency &  \begin{tabular}[c]{@{}c@{}} (1) Leverage scene and world space information for 3D Gaussian densification and simplification. \\ (2) Support of quality-prioritized rendering or storage-oriented compression. \end{tabular} & \begin{tabular}[c]{@{}c@{}} / \end{tabular}\\ 

Gaussian Frosting~\cite{guedon2024gaussian} & Arxiv & 2024  & Representation & Photorealism &  \begin{tabular}[c]{@{}c@{}} Integration of frosting representation and 3D Gaussian splatting. \end{tabular} & \begin{tabular}[c]{@{}c@{}} Increased storage space. \end{tabular}\\ 

Pixel-GS~\cite{zhang2024pixel} & Arxiv & 2024  & Representation & Photorealism &  \begin{tabular}[c]{@{}c@{}} Mitigate floaters and artifacts via adaptive Gaussian density control based on pixel-wise gradients. \end{tabular} & \begin{tabular}[c]{@{}c@{}} Slight increase of memory usage. \end{tabular}\\ 

Analytic-Splatting~\cite{liang2024analytic} & Arxiv & 2024  & Representation & Photorealism &  \begin{tabular}[c]{@{}c@{}} Derive analytic approximation of the integral of projected 2D Gaussians and utilize it for anti-aliasing. \end{tabular} & \begin{tabular}[c]{@{}c@{}} Drop in rendering speed and increased computation cost. \end{tabular}\\ 

Fed3DGS~\cite{suzuki2024fed3dgs} & Arxiv & 2024  & Representation & / &  \begin{tabular}[c]{@{}c@{}} (1) Integrate 3D-GS to federated framework for large-scale 3D scene reconstruction. \\ (2) Propose appearance modeling to improve quality and distillation-based \\ update strategy for efficient merging of 3D-GS. \end{tabular} & \begin{tabular}[c]{@{}c@{}} Communication costs and require high computational capability for clients. \end{tabular}\\ 

GS-W~\cite{zhang2024gaussian} & Arxiv & 2024  & Representation & Photorealism &  \begin{tabular}[c]{@{}c@{}} (1) Enable each point to model intrinsic and dynamic appearance separately via independent appearance features. \\ (2) Propose an adaptive sampling strategy for capturing detailed local information. \end{tabular} & \begin{tabular}[c]{@{}c@{}} Performance drops in scenes with complex lighting conditions and reflections. \end{tabular}\\ 

TOGS~\cite{zhang2024togs} & Arxiv & 2024  & Representation & Physics &  \begin{tabular}[c]{@{}c@{}} (1) Propose to use temporal opacity offset for connecting 3D Gaussians along the time axis. \\ (2) Integrate smooth loss for regularization of opacity offset table optimization and random pruning for efficiency. \end{tabular} & \begin{tabular}[c]{@{}c@{}} Tend to overfit with limited views. \end{tabular}\\ 

SA-GS~\cite{song2024sa} & Arxiv & 2024  & Representation & Photorealism &  \begin{tabular}[c]{@{}c@{}} (1) Propose 2D scale-adaptive filter to maintain scale consistency. \\ (2) Scale consistency ensures the effective application of super-sampling and integration for anti-aliasing. \end{tabular} & \begin{tabular}[c]{@{}c@{}} Increased rendering time when zooming out. \end{tabular}\\ 

Robust GS~\cite{darmon2024robust} & Arxiv & 2024  & Representation & Photorealism &  \begin{tabular}[c]{@{}c@{}} (1) Model motion blur as Gaussian distribution over camera pose for simultaneous pose and blur correction. \\ (2) Propose off-focus correction for compensation and RGB decoder function for maintaining color consistency. \end{tabular} & \begin{tabular}[c]{@{}c@{}} Limited to defocus blur and camera motion blur. \end{tabular}\\ 

Mirror-3DGS~\cite{meng2024mirror} & Arxiv & 2024  & Representation & Photorealism &  \begin{tabular}[c]{@{}c@{}} (1) Propose virtual viewpoint in the mirror for high-fidelity reconstruction. \\ (2) Introduce a two-stage training procedure for realistic rendering of scenes with mirrors. \end{tabular} & \begin{tabular}[c]{@{}c@{}} Rely on accurate mirror estimation and dataset-specific mirror masks. \end{tabular}\\ 

Z-Splat~\cite{qu2024z} & Arxiv & 2024  & Representation & Photorealism &  \begin{tabular}[c]{@{}c@{}} Integration of various type of acoustic depth measurement for effective geometric regularization. \end{tabular} & \begin{tabular}[c]{@{}c@{}} Do not consider scattering and non-linearity in combination with various modalities. \end{tabular}\\ 

GauStudio~\cite{ye2024gaustudio} & Arxiv & 2024  & Representation & / &  \begin{tabular}[c]{@{}c@{}} Integrate existing 3D-GS techniques as a unified framework that can tailored for desired downstream tasks. \end{tabular} & \begin{tabular}[c]{@{}c@{}} / \end{tabular}\\ 

CoherentGS~\cite{paliwal2024coherentgs} & Arxiv & 2024  & Representation & Costs &  \begin{tabular}[c]{@{}c@{}} (1) Coherent optimization via single and multiview constraints. \\ (2) Optical flow-assisted initialization of 3D-GS. \end{tabular} & \begin{tabular}[c]{@{}c@{}} Rendering scenes with transparent objects. \end{tabular}\\ 

Octree-GS~\cite{ren2024octree} & Arxiv & 2024  & Representation & Efficiency &  \begin{tabular}[c]{@{}c@{}} Integrate octree structure into Scafffold-GS for effective and efficient LOD representation. \end{tabular} & \begin{tabular}[c]{@{}c@{}} (1) Dependent on hyperparameter tuning. \\(2) Inherit existing issues of original 3D-GS. \end{tabular}\\ 

3DGStream~\cite{sun20243dgstream} & CVPR & 2024  & Representation & Physics &  \begin{tabular}[c]{@{}c@{}} (1) Models 3D Gaussian dynamics via neural transformation cache which \\ consists of multi-resolution hash encoding and shallow MLP. \\ (2) Adaptive 3D-GS growing strategy based on view-space positional gradients. \end{tabular} & \begin{tabular}[c]{@{}c@{}} Require accurate initialization of point cloud. \end{tabular}\\ 

SWAG~\cite{dahmani2024swag}  & Arxiv & 2024  & Representation  & Photorealism &  \begin{tabular}[c]{@{}c@{}} (1) Reconstruct scenes from an unconstrained photo collection with different appearances. \\ (2) A specific training scheme that captures occlusions and enables \\ the removal of transient objects from the trained scene. \end{tabular} & \begin{tabular}[c]{@{}c@{}} Suffer from transient objects. \end{tabular}\\ 

SpikeNVS~\cite{dai2024spikenvs}  & Arxiv & 2024  & Representation & Photorealism &  \begin{tabular}[c]{@{}c@{}} (1) Evaluate the advantages of spikes over event streams in terms of quality and cost for NVS. \\ (2) A novel Texture from Spike (TfS) loss that integrates the advantages \\ of both Texture from Interval (TFI) and Texture from Playback (TFP). \\ (3) Synchronized spiking RGB camera system with aligned fields-of-view and triggering times. \end{tabular} & \begin{tabular}[c]{@{}c@{}} Extend it to high-speed capture. \end{tabular}\\ 

GaGS~\cite{lu20243d}  & CVPR & 2024  & Representation & Physics &  \begin{tabular}[c]{@{}c@{}} (1) Geometry-aware feature extraction network to better utilize local geometric information. \\ (2) Adapt 3D-GS to dynamic scenes using a continuous 6D \\ rotational representation and a modified density control strategy.  \end{tabular} & \begin{tabular}[c]{@{}c@{}} /  \end{tabular}\\ 

Revising Densification~\cite{bulo2024revising}  &  Arxiv & 2024  & Representation & Photorealism &  \begin{tabular}[c]{@{}c@{}} (1) Ability to guide the densification process based on an auxiliary per-pixel error function. \\ (2) Correct systematic biases in the original growth process when cloning Gaussian.  \end{tabular} & \begin{tabular}[c]{@{}c@{}} Affected by 3D-GS on (1) strong view-dependent effects, \\ (2) appearance variations across images, \\ (3) errors induced by the linear approximation in the Splatting operation. \end{tabular}\\ 

LoopGaussian~\cite{li2024loopgaussian}  &  Arxiv & 2024  & Representation & Physics &  \begin{tabular}[c]{@{}c@{}} (1) Generate realistic 3D Cinemagraph from multi-view images of static scenes. \\ (2) Describe the dynamics of a scene using Euler motion fields in 3D space.  \end{tabular} & \begin{tabular}[c]{@{}c@{}} / \end{tabular}\\ 

Compact 3D Gaussian~\cite{lee2023compact}  & CVPR & 2024  & Representation & Efficiency &  \begin{tabular}[c]{@{}c@{}} Vector quantization based on k-means, compressed \\ indexing and employing a compact codebook. \end{tabular} & \begin{tabular}[c]{@{}c@{}} /  \end{tabular}\\ 

CompGS~\cite{liu2024compgs}  & Arxiv & 2024  & Representation & Efficiency &  \begin{tabular}[c]{@{}c@{}} (1) Hybrid primitives for efficient 3D scene representation. \\ (2) Regularize optimization under compression rate constraint for more compact primitives. \end{tabular} & \begin{tabular}[c]{@{}c@{}} /  \end{tabular}\\

MCMC~\cite{kheradmand20243d}  &  Arxiv & 2024  & Representation & Photorealism &  \begin{tabular}[c]{@{}c@{}} Reformulate 3D-GS as Markov Chain Monte Carlo.  \end{tabular} & \begin{tabular}[c]{@{}c@{}} / \end{tabular}\\ 

LetsGo~\cite{cui2024letsgo}  &  Arxiv & 2024  & Representation & Efficiency &  \begin{tabular}[c]{@{}c@{}} (1) Model a wide range of challenging garage scenes. \\ (2) Deep priors for training Gaussian primitive representations, \\ and a lightweight renderer for real-time rendering on consumer devices. \end{tabular} & \begin{tabular}[c]{@{}c@{}} (1) Explore lighter devices for scanning and rendering large scenes. \\ (2) Modifying lighting conditions is not supported due to \\ existing pipeline constraints. \end{tabular}\\

SRGS~\cite{feng2024srgs} &  Arxiv & 2024  & Representation & Photorealism &  \begin{tabular}[c]{@{}c@{}} (1) High-quality HRNVS using only existing LR views. \\ (2) Utilize super-resolution Gaussian densification strategies. \\ (3) A texture-guided Gaussian learning strategy is introduced to guide Gaussian elements to \\ learn textures from external priors of the 2D SR model.  \end{tabular} & \begin{tabular}[c]{@{}c@{}} / \end{tabular}\\

AbsGS~\cite{ye2024absgs} &  Arxiv & 2024  & Representation  & Photorealism &  \begin{tabular}[c]{@{}c@{}} (1) Utilize homodirectional view-space positional gradient as guidance for densification. \\ (2) Eliminate large Gaussians in over-reconstructed areas. \end{tabular} & \begin{tabular}[c]{@{}c@{}} / \end{tabular}\\

DeblurGS~\cite{oh2024deblurgs} &  Arxiv & 2024  & Representation  & Photorealism &  \begin{tabular}[c]{@{}c@{}} (1) Propose sub-frame pose estimation for blur view synthesis via optimizable sub-frame alignment parameters. \\ (2) Adopt Gaussian densification annealing to refine the initial estimation of camera poses. \end{tabular} & \begin{tabular}[c]{@{}c@{}} / \end{tabular}\\

GeoGaussian~\cite{li2024geogaussian} &  Arxiv & 2024  & Representation  & Photorealism &  \begin{tabular}[c]{@{}c@{}} (1) Initialize and densify 3D-GS with explicitly geometric meaning that leads to reasonable scene reconstruction. \\ (2) Introduce smooth constraints during optimization for geometric consistency. \end{tabular} & \begin{tabular}[c]{@{}c@{}} / \end{tabular}\\

HO-Gaussian~\cite{li2024ho} &  Arxiv & 2024  & Representation  & Photorealism &  \begin{tabular}[c]{@{}c@{}} (1) Avoid the reliance of accurate point initialization. \\ (2) Propose improved representation and optimize both volume and Gaussians to facilitate end-to-end training. \end{tabular} & \begin{tabular}[c]{@{}c@{}} Longer training time when compared to original 3D-GS. \end{tabular}\\

EfficientGS~\cite{liu2024efficientgs} &  Arxiv & 2024  & Representation  & Efficiency &  \begin{tabular}[c]{@{}c@{}} Combination of selective Gaussian densification, pruning and gradual increase \\ of Spherical order strategies for accelerated training and memory saving.  \end{tabular} & \begin{tabular}[c]{@{}c@{}} / \end{tabular}\\

\begin{tabular}[c]{@{}c@{}}Analysis of \\ SfM Initialization~\cite{foroutan2024does} \end{tabular} &  Arxiv & 2024  & Representation  & / &  \begin{tabular}[c]{@{}c@{}} Comprehensive and sound analysis of SfM initialization based on thorough experiments.  \end{tabular} & \begin{tabular}[c]{@{}c@{}} / \end{tabular}\\

\bottomrule[1.5pt]
\end{tabular}%
}
\label{summary_representation2}
\end{table*}

% Please add the following required packages to your document preamble:
% \usepackage{graphicx}
\begin{table*}[htbp]
\centering
\setlength\tabcolsep{1.5pt}
\renewcommand{\arraystretch}{1.3}
\caption{Summary of methods derived from 3D-GS for reconstruction.}
\resizebox{\textwidth}{!}{%
\begin{tabular}{>{\columncolor{C7!50}}l>{\columncolor{C6!50}}c>{\columncolor{C5!50}}c>{\columncolor{C4!50}}c>{\columncolor{C3!50}}c>{\columncolor{C2!50}}c>{\columncolor{C1!50}}c}
\toprule[1.5pt]
Methods   & Venue & Year & Category &  Sub-category  & Contributions                                                                                                                         & Limitations                                                                       \\ \midrule[1pt]

Splatter Image~\cite{szymanowicz2023splatter}& CVPR & 2024 & Reconstruction& Static& \begin{tabular}[c]{@{}c@{}}(1) A 3D-GS based reconstruction method tailored for single-view\\ scenarios, with potential extension to multi-view reconstruction.\\(2) A 3D reconstruction approach with high quality and speed.   \end{tabular} & \begin{tabular}[c]{@{}c@{}} / \end{tabular}\\

pixelSplat~\cite{charatan2023pixelsplat}& CVPR & 2024 & Reconstruction& Static & \begin{tabular}[c]{@{}c@{}}(1) An approach reconstructing a primitive-based parameterization\\ of the 3D radiance field from two images.\\
(2) Utilizing a dense probability distribution for primitive location parameterization. \\(3) Enabling gradient backpropagation into the parameters. \end{tabular} & \begin{tabular}[c]{@{}c@{}} / \end{tabular}\\ 

Triplane-Gaussian~\cite{zou2023triplane}& CVPR & 2024 & Reconstruction & Static &\begin{tabular}[c]{@{}c@{}} (1) A hybrid representation for single-view reconstruction.\\(2) Applying projection-aware conditioning and geometry-aware\\ encoding to improve the consistency with the input observation. \end{tabular} & \begin{tabular}[c]{@{}c@{}} (1) Limited by the quality of the predicted point cloud.\\(2)  The backside tends to be blurry.\\(3) Relying on the camera parameters. \end{tabular}\\

SuGaR~\cite{guedon2023sugar}& CVPR & 2024 & Reconstruction& Static &  \begin{tabular}[c]{@{}c@{}} (1) Incorporating a regularization term for precise scene geometry capture with Gaussians.\\
(2) Developing an efficient algorithm for accurate mesh extraction from Gaussians.\\
(3) Enhancing mesh accuracy and rendering quality by binding Gaussians to the mesh. \end{tabular} & \begin{tabular}[c]{@{}c@{}} (1) Rendering quality is reduced while extracting meshes. \\ (2) Relatively unsuitable for extracting meshes from large scenes. \end{tabular}\\ 

NeuSG~\cite{chen2023neusg}&ArXiv&2023&Reconstruction&Static&  \begin{tabular}[c]{@{}c@{}}(1) A method that jointly optimizes NeuS and 3D-GS. \\(2) Ensuring that point clouds from extremely\\ thin 3D Gaussians closely adhere to the surfaces.  \end{tabular} & \begin{tabular}[c]{@{}c@{}} (1) Constrained by the number of images input.\\(2) Difficulties arise in environments with sparse or uneven image distribution. \end{tabular}\\

GaussainObject~\cite{yang2024gaussianobject}& ArXiv &2024  &Reconstruction & Static &  \begin{tabular}[c]{@{}c@{}} 
(1) Optimizing 3D Gaussians from sparse views with explicit structural priors.\\
(2) Introducing a Gaussian repair model based on diffusion models.   \end{tabular} & \begin{tabular}[c]{@{}c@{}} (1) Dependency on precise poses for visual hull construction poses daily life challenges.\\
(2) Color shifts on white backgrounds could impact Gaussian repair. \end{tabular}\\ 

latentSplat~\cite{wewer2024latentsplat}&  ArXiv &2024  &Reconstruction& Static &\begin{tabular}[c]{@{}c@{}}  A fast and efficient two-view reconstruction method for both training and rendering. \end{tabular} & /\\ 

DN-Splatter~\cite{turkulainen2024dn}&  ArXiv &2024  &Reconstruction& Static  &  \begin{tabular}[c]{@{}c@{}} (1) Integrating depth and smoothness priors into GS optimization.\\
(2) Introducing monocular normal priors to align Gaussians with scene geometry.  \end{tabular} & \begin{tabular}[c]{@{}c@{}} (1) Limited to densely captured scenes with static cameras.\\
(2) Improved meshing techniques needed for simultaneous\\ optimization of Gaussian parameters and mesh quality. \end{tabular}\\ 

GauMesh~\cite{xiao2024bridging}&ArXiv&2024&Reconstruction& Static& \begin{tabular}[c]{@{}c@{}} (1) A hybrid representation integrating triangle meshes and 3D Gaussians. \\(2) A hybrid differentiable rendering pipeline. \\(3) A joint optimization process.  \end{tabular} & \begin{tabular}[c]{@{}c@{}} / \end{tabular}\\

GSDF~\cite{yu2024gsdf}&ArXiv&2024&Reconstruction& Static&\begin{tabular}[c]{@{}c@{}} (1) A Dual-branch framework based on both 3D-GS and SDF.\\
(2) A bidirectional mutual guidance approach.\end{tabular} & \begin{tabular}[c]{@{}c@{}}(1) Challenging in scenes with reflections\\ and intense lighting variations. \\(2) Increased training durations resulting from asynchronous branching. \end{tabular}\\ 
%%%%%%%%%%%%%%%%%%%

2D-GS~\cite{huang20242d}&ArXiv&2024&Reconstruction& Static&  \begin{tabular}[c]{@{}c@{}} (1) A differentiable 2D Gaussian renderer for accurate surface presentation. \\(2) Regularization losses to enhance\\ surface reconstruction and minimize noise.  \end{tabular} & \begin{tabular}[c]{@{}c@{}} (1) Challenging in accurately handling semi-transparent surfaces. \\ (2) Occasionally inaccurate representations of fine geometric structures. \\(3) Possibly excessively smooth in certain regions.\end{tabular}\\ 

\begin{tabular}[c]{@{}c@{}}Tactile-Informed\\ 3DGS~\cite{comi2024snap}\end{tabular}&ArXiv&2024&Reconstruction& Static& \begin{tabular}[c]{@{}c@{}}(1) An approach for incorporating touch sensing with vision\\ to achieve surface reconstruction and novel view synthesis. \\ (2) SOTA geometry reconstruction on reflective and glossy surfaces.\\(3) Faster scene reconstruction and improved performance.\end{tabular} & \begin{tabular}[c]{@{}c@{}} Random touch sampling for scene reconstruction\\ may not always be efficient. \end{tabular}\\

3DGSR~\cite{lyu20243dgsr}&ArXiv&2024&Reconstruction& Static& \begin{tabular}[c]{@{}c@{}} (1) A differentiable SDF-to-opacity transformation function. \\(2) A consistency regularization based on volumetric\\ rendering and geometry consistency.  \end{tabular} & \begin{tabular}[c]{@{}c@{}} Limited by the trade-off between rendering\\ quality and surface smoothness.  \end{tabular}\\

OmniGS~\cite{li2024omnigs}&ArXiv&2024&Reconstruction&Static&  \begin{tabular}[c]{@{}c@{}} (1) A theoretical analysis of the omnidirectional GS.\\(2) A photorealistic reconstruction system based on\\ a GPU-accelerated omnidirectional rasterizer. \end{tabular} & \begin{tabular}[c]{@{}c@{}} Sacrificing computational resources maybe\\ necessary to enhance quality. \end{tabular}\\

Gs2mesh~\cite{wolf2024surface}&ArXiv&2024&Reconstruction&Static& \begin{tabular}[c]{@{}c@{}} (1) An approach for surface reconstruction from GS models. \\(2) Faster and more accurate reconstruction, \\ and suitable for in-the-wild scenes.  \end{tabular} & \begin{tabular}[c]{@{}c@{}}(1) Anomalies introduced in Gaussians may lead to inaccuracies. \\(2) Accuracy of reconstruction is susceptible to transparent surfaces.\end{tabular}\\

%\begin{tabular}[c]{@{}c@{}}GauU-Scene\\ V2~\cite{xiong2024gauu}\end{tabular}&ArXiv&2024&Reconstruction&Dataset& \begin{tabular}[c]{@{}c@{}} (1) A comprehensive multimodal dataset.\\(2) A method for aligning SfM camera positions with LiDAR data points.\\(3) A detailed bench-marking of current popular 3D reconstruction methods.  \end{tabular} & \begin{tabular}[c]{@{}c@{}} / \end{tabular}\\

360-GS~\cite{bai2024360}&ArXiv&2024&Reconstruction&Static&  \begin{tabular}[c]{@{}c@{}}(1) A layout-guided 3D-GS pipeline designed for sparse panoramic images. \\(2) A high-quality point cloud generation method.\\(3) A layout-guided regularization on 3D Gaussians.  \end{tabular} & \begin{tabular}[c]{@{}c@{}} (1) Possibly inaccurate priors for complex scenes.\\(2) Storage costs and rendering quality are conflicting. \end{tabular}\\

Gamba~\cite{shen2024gamba}& ArXiv & 2024 &Reconstruction &Static&  \begin{tabular}[c]{@{}c@{}} (1) Combining 3D-GS with State Space Models(SSMs). \\ (2)  A 3D-GS Generation pipeline for fast and\\ high-quality single-view reconstruction. \end{tabular} & \begin{tabular}[c]{@{}c@{}} (1) 
Struggling to generate sharp textures for occluded areas.\\ (2) Generalization capability needs to be strengthened.  \end{tabular}\\

\begin{tabular}[c]{@{}c@{}}GS Point Cloud\\ Completion~\cite{huang2024zero}\end{tabular} & ArXiv & 2024 &Reconstruction &Static&  \begin{tabular}[c]{@{}c@{}} (1) A zero-shot framework for completing point clouds, \\leveraging 2D priors from the diffusion model through 3D-GS.\\(2) A point cloud colorization operation. \\(3) A zero-shot fractal completion operation. \end{tabular} & \begin{tabular}[c]{@{}c@{}} A separate optimization process is required to\\ integrate 2D priors from diffusion models.
  \end{tabular}\\

GOF~\cite{yu2024gaussian}& ArXiv & 2024 &Reconstruction &Static&\begin{tabular}[c]{@{}c@{}}(1) Establishing a Gaussian opacity field from a set of 3D Gaussians. \\(2) Approximating the surface normals of 3D Gaussians as the\\ normals of intersection planes between the ray and Gaussians.\\(3) An efficient surface extraction technique based on tetrahedra-grids.  \end{tabular} & \begin{tabular}[c]{@{}c@{}} / \end{tabular}\\ 

TCLC-GS~\cite{zhao2024tclc}& ArXiv & 2024 &Reconstruction &Static&\begin{tabular}[c]{@{}c@{}}(1) A hybrid 3D
representation derived from LiDAR-camera data and 3D-GS.\\(2) Incorporating rendered dense depth data within the 3D mesh.  \end{tabular} & \begin{tabular}[c]{@{}c@{}}  /\end{tabular}\\ 

CityGaussian~\cite{liu2024citygaussian}& ArXiv & 2024 &Reconstruction &Dynamic&  \begin{tabular}[c]{@{}c@{}}   (1) A divide-and-conquer strategy to \\ reconstruct large-scale 3D-GS representations in parallel manner.\\ (2) A block-wise Level-of-Detail strategy to reduce the computation footprint. \end{tabular} & \begin{tabular}[c]{@{}c@{}}  Limited by the hidden static scene assumption. \end{tabular}\\

DrivingGaussian~\cite{zhou2023drivinggaussian}& CVPR & 2024 & Reconstruction& Dynamic& \begin{tabular}[c]{@{}c@{}} (1) A framework based on 3D-GS to represent\\ and model dynamic and large-scale driving scenes. \\ (2) Reconstructing the static background and\\ modeling multiple dynamic objects.  \end{tabular} & \begin{tabular}[c]{@{}c@{}} Modeling numerous extremely small objects and materials featuring\\ total reflection properties presents a considerable challenge.\end{tabular}\\

\begin{tabular}[c]{@{}c@{}}Neural Parametric \\Gaussians~\cite{das2023neural} \end{tabular}& CVPR & 2024 & Reconstruction& Dynamic&  \begin{tabular}[c]{@{}c@{}}(1) A two-stage reconstruction approach based on 3D-GS.\\(2) Utilizing a coarse parametric point model founded on low-rank deformation.  \end{tabular} & \begin{tabular}[c]{@{}c@{}} Constrained by camera movement, lighting\\ conditions, speed, and magnitude of deformations. \end{tabular}\\

Gaussian-flow~\cite{lin2023gaussian}& CVPR & 2024 & Reconstruction &Dynamic& \begin{tabular}[c]{@{}c@{}} (1) A point-based reconstruction method for dynamic 3D scene leveraging 3D-GS.\\(2) A Dual-Domain Deformation Model for efficient reconstruction in 4D scenes. \\ (3) Facilitating the perception operations of both static and dynamic 3D scenes.  \end{tabular} & \begin{tabular}[c]{@{}c@{}} Improving the fidelity of thin structures is a potential enhancement. \end{tabular}\\

\begin{tabular}[c]{@{}c@{}}Deformable \\3D-GS~\cite{yang2023deformable}\end{tabular}& CVPR & 2024 & Reconstruction&Dynamic &  \begin{tabular}[c]{@{}c@{}} (1) A deformable 3D-GS framework for reconstructing monocular dynamic scenes. \\(2) An annealing smoothing training mechanism.\\  \end{tabular} & \begin{tabular}[c]{@{}c@{}} (1) Sparse-view datasets and limited coverage may result in overfitting.\\
(2) Effectiveness relies on the precision of pose estimations.\\
(3) Limited in managing complex human motions.\end{tabular}\\

VastGaussian~\cite{lin2024vastgaussian} &CVPR  &2024  & Reconstruction& Dynamic  &  \begin{tabular}[c]{@{}c@{}}(1) Introducing a progressive data partitioning strategy.\\
(2) Integrating decoupled appearance modeling into the optimization process.  \end{tabular} & \begin{tabular}[c]{@{}c@{}} (1) Lacking an optimal solution for division. \\(2) Large scenes may reduce rendering speed\\ and increase storage requirements. \end{tabular}\\ 

PVG~\cite{chen2023periodic}&ArXiv&2023&Reconstruction&Dynamic&\begin{tabular}[c]{@{}c@{}} (1) A method for large-scale dynamic urban scene reconstruction.\\
(2) A scene flow-based temporal smoothing mechanism for enhanced continuity.\\
(3) A position-aware point adaptive control for unbounded urban scenes.\end{tabular} & \begin{tabular}[c]{@{}c@{}} Challenging to model smoothness across time and space. \end{tabular}\\ 

\begin{tabular}[c]{@{}c@{}}Motion-aware \\3D-GS~\cite{guo2024motion} \end{tabular} &  ArXiv &2024  &Reconstruction&Dynamic  &  \begin{tabular}[c]{@{}c@{}}(1) 
(1) A dynamic scene reconstruction method with flow priors.\\
(2) Uncertainty-aware flow augmentation and transient-aware deformation strategies. \end{tabular} & \begin{tabular}[c]{@{}c@{}} (1) Sometimes constrained by motion blur.\\(2) Challenging in limitation of motion uncertainty. \end{tabular}\\ 
    
HGS-Mapping~\cite{wu2024hgs}&ArXiv&2024&Reconstruction&Dynamic& \begin{tabular}[c]{@{}c@{}} (1)  A hybrid 3D-GS representation facilitates\\ the reconstruction of entire urban scenes.\\ (2)  A hybrid Gaussian initialization mechanism\\ and an adaptive update method. \end{tabular} & \begin{tabular}[c]{@{}c@{}} Performance is limited on rugged roads or significant curvature. \end{tabular}\\

SGD~\cite{yu2024sgd}& ArXiv & 2024 &Reconstruction &Dynamic&\begin{tabular}[c]{@{}c@{}} (1) A framework for novel view synthesis in street scenes,\\ combining 3D-GS with a customized diffusion model.\\(2) A strategy for fine-tuning a diffusion model on autonomous\\ driving datasets and equipping it with NVS capability. \end{tabular} & \begin{tabular}[c]{@{}c@{}} Filtering operation of diffusion model is time-consuming. \end{tabular}\\ 

% &  &  &  & & \begin{tabular}[c]{@{}c@{}} \\  \end{tabular} & \begin{tabular}[c]{@{}c@{}}  \end{tabular}\\ 

\bottomrule[1.5pt]
\end{tabular}
}
\label{summary_reconstruction}
\end{table*}

% Please add the following required packages to your document preamble:
% \usepackage{graphicx}
\begin{table*}[htbp]
\centering
\setlength\tabcolsep{1.5pt}
\renewcommand{\arraystretch}{1.1}
\caption{Summary of methods derived from 3D-GS for manipulation.}
\resizebox{\textwidth}{!}{%
\begin{tabular}{>{\columncolor{C7!50}}l>{\columncolor{C6!50}}c>{\columncolor{C5!50}}c>{\columncolor{C4!50}}c>{\columncolor{C3!50}}c>{\columncolor{C2!50}}c>{\columncolor{C1!50}}c}
\toprule[1.5pt]
Methods   & Venue & Year & Category  &  Sub-category  & Contributions                                                                                                                         & Limitations                                                                       \\ \midrule[1pt]

TIP-Editor~\cite{zhuang2024tip} & ACM TOG & 2024 & Manipulation & 3D Manipulation & \begin{tabular}[c]{@{}c@{}} (1) A versatile 3D scene editing framework \\ allows for various editing operations. \\ 
(2) A novel stepwise 2D personalization strategy. \end{tabular} & \begin{tabular}[c]{@{}c@{}} (1) It performs poorly in complex scenes where \\ the bounding box may contain unwanted elements. \\ 
(2) Extracting a smooth and accurate mesh can \\ be a challenging task.  \end{tabular}\\ 

GaMeS~\cite{waczynska2024games} & Arxiv & 2024 & Manipulation & 3D Manipulation & \begin{tabular}[c]{@{}c@{}} (1) A hybrid representation for 3D objects, \\ integrating mesh and GS. \\ (2) Work on the simple pseudo-mesh and \\ rely only upon essential vector operation. \end{tabular} & \begin{tabular}[c]{@{}c@{}} Artifacts can manifest when significant changes \\ occur in the case of meshes with large faces.  \end{tabular}\\ 

Mesh-based GS~\cite{gao2024mesh} & Arxiv & 2024 & Manipulation & 3D Manipulation & \begin{tabular}[c]{@{}c@{}} (1) Utilize the mesh to guide the splitting of 3D-GS. \\ 
(2) A large-scale Gaussian deformation. \end{tabular} & \begin{tabular}[c]{@{}c@{}} (1) The visual appearance and shadow of 3D-GS \\ remain fixed and cannot be further modified. \\ 
(2) Hinge on the utilization of the extracted mesh as a proxy.  \end{tabular}\\ 

3DGM~\cite{wang20243d} & Arxiv & 2024 & Manipulation & 3D Manipulation & \begin{tabular}[c]{@{}c@{}} (1) A proxy-based representation. \\ 
(2) An injective mapping to optimize \\ 3D-GS in texture space. \\ 
(3) Constrain the 3D-GS with each triangle,\\ facilitating animation and texture mapping. \end{tabular} & \begin{tabular}[c]{@{}c@{}} An approximate geometry is needed to represent the target. \end{tabular}\\ 

GSEdit~\cite{gao2024mesh} & Arxiv & 2024 & Manipulation & 3D Manipulation & \begin{tabular}[c]{@{}c@{}} Using consumer-grade hardware, a large number of modifications \\ can be made to any given shape in a short time.  \end{tabular} & \begin{tabular}[c]{@{}c@{}} (1) Perspective bias. \\ (2) Restrict substantial spatial transformations.  \end{tabular}\\ 

StyleGaussian~\cite{liu2024stylegaussian} & Arxiv & 2024 & Manipulation & 3D Manipulation & \begin{tabular}[c]{@{}c@{}} (1) Achieve real-time rendering and multi-view consistency, \\ while enabling instant style transfer at 10 FPS. \\ 
(2) An efficient strategy for rendering high-dimensional features. \end{tabular} & \begin{tabular}[c]{@{}c@{}} (1) The modification solely pertains to the color of 3D-GS, \\ with no alteration to the underlying geometry. \\ 
(2) Memory and computation costs.  \end{tabular}\\ 

GSS~\cite{saroha2024gaussian} & Arxiv & 2024 & Manipulation & 3D Manipulation & \begin{tabular}[c]{@{}c@{}} Gaussians conditioned on a style image are utilized \\ to produce stylized renderings of a complex 3D scene. \end{tabular} & \begin{tabular}[c]{@{}c@{}} /  \end{tabular}\\ 

GaussCtrl~\cite{wu2024gaussctrl} & Arxiv & 2024 & Manipulation & 3D Manipulation & \begin{tabular}[c]{@{}c@{}}  The depth guidance and the integration of the attention-based \\ latent code alignment module facilitate the \\ enhancement of multi-view consistent editing.\end{tabular} & \begin{tabular}[c]{@{}c@{}} (1) The original geometry of the scene is altered through \\ the utilization of depth maps during the editing process.\\ (2) The results may not always accurately \\ reflect the user's original intention.  \end{tabular}\\ 

Texture-GS~\cite{xu2024texture} & Arxiv & 2024 & Manipulation & 3D Manipulation & \begin{tabular}[c]{@{}c@{}}  (1) Disentangle the geometry and texture for 3D-GS. \\ 
(2) A texture mapping module to aid in the efficient mapping of \\ ray-Gaussian intersections onto a continuous 2D texture space. \end{tabular} & \begin{tabular}[c]{@{}c@{}} (1) Blurring issue in the edges.\\ 
(2) The definition of the texture space as a unit sphere is \\ insufficient for accurately representing multiple objects or outdoor scenes.  \end{tabular}\\ 

VcEdit~\cite{wang2024view} & Arxiv & 2024 & Manipulation & 3D Manipulation  & \begin{tabular}[c]{@{}c@{}}  (1) Multi-view consistency modules to harmonize \\ the inconsistent multi-view guidance images. \\ 
(2) An effective iterative pattern \\ for high-quality 3D-GS editing. \end{tabular} & \begin{tabular}[c]{@{}c@{}} (1) Existing diffusion-based image editing models occasionally fail \\ to deliver high-quality image editing for intricate prompts. \\ 
(2) In non-rigid editing scenarios, existing diffusion models \\ produce highly variant editing results among views. \end{tabular}\\

StylizedGS~\cite{zhang2024stylizedgs} & Arxiv & 2024 & Manipulation & 3D Manipulation & \begin{tabular}[c]{@{}c@{}} An efficient stylization \\ process and flexible control.  \end{tabular} & \begin{tabular}[c]{@{}c@{}} (1) Geometric artifacts from the original 3D-GS reconstruction. \\ 
(2) Cannot fulfill instant stylization. \end{tabular}\\

MANUS~\cite{pokhariya2023manus} & CVPR & 2024 & Manipulation & 3D Manipulation & \begin{tabular}[c]{@{}c@{}} (1) An efficient representation for articulated hands. \\ (2) Utilize MANUS-Hand and a 3D Gaussian \\ representation to accurately model contacts. \\(3) MANUS-Grasps dataset.  \end{tabular} & \begin{tabular}[c]{@{}c@{}} (1) The complexity of hand dynamics. \\ (2) Model single-hand grasping with static objects. \\ (3) Hand-object manipulation for longer \\ time-frames is unaddressed. \end{tabular}\\

Point’n Move~\cite{huang2023point} & Arxiv & 2023 & Manipulation & 3D Manipulation & \begin{tabular}[c]{@{}c@{}} (1) Fulfill interactive scene object \\ manipulation via exposed region inpainting. \\ (2) A dual-stage self-prompting mask propagation strategy. \\ (3) A fast inpainting method to minimize unnecessary inpainting \\ along with a reprojection-based initialization scheme.  \end{tabular} & \begin{tabular}[c]{@{}c@{}} Lack the ability to handle lighting and texture, \\ as it is solely focused on geometry editing. \end{tabular}\\

GaussianEditor~\cite{fang2023gaussianeditor} & CVPR & 2024 & Manipulation & 3D Manipulation & \begin{tabular}[c]{@{}c@{}} (1) Achieve delicate 3D scene editing. \\ (2) A set of techniques has been developed to accurately \\ localize the specific region of interest for editing. \end{tabular} & \begin{tabular}[c]{@{}c@{}} The descriptions of the same object \\ from different perspectives exhibit variations.  \end{tabular}\\

GaussianEditor~\cite{chen2023gaussianeditor} & Arxiv & 2023 & Manipulation & 3D Manipulation & \begin{tabular}[c]{@{}c@{}} (1) Gaussian semantic tracing enables detailed and effective editing. \\(2) Hierarchical Gaussian Splatting demonstrates greater stability in \\ converging towards refined results when subjected to highly random generative guidance. \\(3) A 3D inpainting algorithm enables the efficient removal and addition of objects. \end{tabular} & \begin{tabular}[c]{@{}c@{}} Rely on 2D diffusion models for facilitating effective supervision. \end{tabular}\\

GART~\cite{lei2023gart} & CVPR & 2024 & Manipulation & 3D Manipulation & \begin{tabular}[c]{@{}c@{}} (1) A general and explicit representation for non-rigid articulated subjects. \\ (2) GART can be efficiently animated through learnable forward skinning \\ and allow for the accurate representation of complex deformations. \end{tabular} & \begin{tabular}[c]{@{}c@{}} Rely on a template pose estimator.   \end{tabular}\\

DeferredGS~\cite{wu2024deferredgs}  & Arxiv & 2024 & Manipulation & 3D Manipulation & \begin{tabular}[c]{@{}c@{}} (1) Decoupled 3D-GS with editable geometry, textures and lighting.\\ (2) Apply deferred shading strategies to mitigate blending artifacts.  \end{tabular} & \begin{tabular}[c]{@{}c@{}} / \end{tabular}\\

DeferredGS~\cite{wu2024deferredgs}  & Arxiv & 2024 & Manipulation & 3D Manipulation & \begin{tabular}[c]{@{}c@{}} (1) Incorporate multiview monocular depth estimation as an additional constraint.\\
(2) A feature regularization strategy is proposed to enhance the interaction between Gaussian \\ clusters in both the in-painted and visible regions of the scene.  \end{tabular} & \begin{tabular}[c]{@{}c@{}} / \end{tabular}\\

BAGS~\cite{zhang2024bags} & Arxiv & 2024 & Manipulation & 4D Manipulation & \begin{tabular}[c]{@{}c@{}}  (1) Diffusion priors are integrated to accommodate the \\ limited information pertaining to unseen perspectives in casual videos. \\ 
(2) A robust regularization technique \\ to enhance the utilization of prior information efficiently. \end{tabular} & \begin{tabular}[c]{@{}c@{}} (1) The motion information is dependent on the input video.\\ 
(2) The generation of unseen views is \\ contingent upon the diffusion model.  \end{tabular}\\ 

\begin{tabular}[c]{@{}c@{}} Per-Gaussian \\ Deformation~\cite{bae2024per} \end{tabular} & Arxiv & 2024 & Manipulation & 4D Manipulation & \begin{tabular}[c]{@{}c@{}}  (1) Formulate a product space of per-Gaussian \\ embeddings and temporal embeddings. \\ 
(2) Temporal variations of the parameters are \\ decomposed into coarse and fine components. \end{tabular} & \begin{tabular}[c]{@{}c@{}} Relatively slow.  \end{tabular}\\

Feature Splatting~\cite{qiu2024feature} & Arxiv & 2024 & Manipulation & 4D Manipulation & \begin{tabular}[c]{@{}c@{}} (1) Augment static scenes with semantics and \\ language-grounded physically realistic movements. \\ 
(2) Combine features from multiple foundation \\ models for accurate decomposition.  \end{tabular} & \begin{tabular}[c]{@{}c@{}} Artifacts in the background. \end{tabular}\\

Control4D~\cite{shao2023control4d} & CVPR & 2024 & Manipulation & 4D Manipulation & \begin{tabular}[c]{@{}c@{}} (1) A 4D representation GaussianPlanes for 4D editing \\ by employing plane-based decomposition. \\ (2) 4D generator to minimize the inconsistent supervision signals \\ and enhance the quality of 4D editing. \end{tabular} & \begin{tabular}[c]{@{}c@{}}  (1) Rely on learning flow within the 4D \\ scenes to exhibit simplicity and smoothness. \\ (2) Constrained by ControlNet, the granularity \\ of edits is limited to a coarse level. \end{tabular}\\

CoGS~\cite{yu2023cogs} & CVPR & 2024 & Manipulation & 4D Manipulation & \begin{tabular}[c]{@{}c@{}} (1) Adapt 3D-GS for dynamic environments \\ captured by a monocular camera. \\ (2) Integrate control mechanisms for intuitive \\ and straightforward manipulation.  \end{tabular} & \begin{tabular}[c]{@{}c@{}} /  \end{tabular}\\

SC-GS~\cite{huang2023sc} & CVPR & 2024 & Manipulation & 4D Manipulation & \begin{tabular}[c]{@{}c@{}} (1) Introduce sparse control points for modeling scene motion.\\ (2) Employ adaptive learning strategies and \\ design a regularization loss based on rigid constraints.\\  \end{tabular} & \begin{tabular}[c]{@{}c@{}} The performance is susceptible to inaccuracies in camera poses.  \end{tabular}\\

\bottomrule[1.5pt]
\end{tabular}%
}
\label{summary_manipulation}
\end{table*}

% Please add the following required packages to your document preamble:
% \usepackage{graphicx}
\begin{table*}[htbp]
\centering
\setlength\tabcolsep{1.5pt}
\renewcommand{\arraystretch}{1.2}
\caption{Summary of methods derived from 3D-GS for generation.}
\resizebox{\textwidth}{!}{%
\begin{tabular}{>{\columncolor{C7!50}}l>{\columncolor{C6!50}}c>{\columncolor{C5!50}}c>{\columncolor{C4!50}}c>{\columncolor{C3!50}}c>{\columncolor{C2!50}}c>{\columncolor{C1!50}}c}
\toprule[1.5pt]
Methods   & Venue & Year & Category & Sub-ategory  & Contributions                                                                                                                         & Limitations                                                                       \\ \midrule[1pt]

GaussianDreamer~\cite{yi2023gaussiandreamer} & CVPR & 2024 & \begin{tabular}[c]{@{}c@{}} Generation \end{tabular} & 3D Object-level & \begin{tabular}[c]{@{}c@{}} (1) A text-to-3D method that combines 3D and 2D diffusion models \\
 for both 3D consistency and rich generation details. \\
 (2) Optimizing the initialized Gaussians with two operations of noisy point growing \\
 and color perturbation for content enrichment.\\  \end{tabular} & \begin{tabular}[c]{@{}c@{}} This method suffers from multi-face problems \\ and fails to generate large scale scenes. \end{tabular}\\ 

CG3D~\cite{vilesov2023cg3d} & ArXiv & 2023 & \begin{tabular}[c]{@{}c@{}} Generation  \end{tabular} & 3D Scene-level & \begin{tabular}[c]{@{}c@{}} (1) A scene-level generation framework that compositionally generates scalable 3D assets. \\ (2) It generates physically realistic scenes with  object composition parameters  \\
while maintaining object separability and fidelity. \\
 \end{tabular} & \begin{tabular}[c]{@{}c@{}} This method also inherits common object generation problems \\ such as the Janus problem and geometry/texture ambiguities. \end{tabular}\\

DreamGaussian~\cite{tang2023dreamgaussian} & ICLR & 2024 & \begin{tabular}[c]{@{}c@{}} Generation  \end{tabular} & 3D Object-level & \begin{tabular}[c]{@{}c@{}} (1) First work to adapt 3D-GS into generative settings for 3D content creation. \\ (2) Efficient mesh extraction algorithm for 3D Gaussians and \\ UV space texture refinement stage.
 \end{tabular} & \begin{tabular}[c]{@{}c@{}} Multi-face Janus problem, \\ oversaturated texture, \\ and baked lighting. \end{tabular}\\

LucidDreamer (Scene)~\cite{chung2023luciddreamer} & ArXiv & 2023 & \begin{tabular}[c]{@{}c@{}} Generation  \end{tabular} & 3D Scene-level & \begin{tabular}[c]{@{}c@{}} (1) A domain-free high-quality 3D scene generation. \\ (2) The dream process establishes the point cloud \\ as a geometric guide for each image generation. \\ (3) The alignment process combines the resulting images to form a unified 3D scene.
 \end{tabular} & \begin{tabular}[c]{@{}c@{}} / \end{tabular}\\

 LucidDreamer (Object)~\cite{liang2023luciddreamer} & CVPR & 2024 & \begin{tabular}[c]{@{}c@{}} Generation  \end{tabular} & 3D Object-level & \begin{tabular}[c]{@{}c@{}} (1) An in-depth analysis of Score Distillation Sampling.\\
(2) Interval Score Matching for invertible diffusion trajectories and interval-based matching.
 \end{tabular} & \begin{tabular}[c]{@{}c@{}} / \end{tabular}\\

DreamPropeller~\cite{zhou2023dreampropeller} & ArXiv & 2023 & \begin{tabular}[c]{@{}c@{}} Generation  \end{tabular} & 3D Object-level & \begin{tabular}[c]{@{}c@{}} Generalize the Picard iteration to more complex computational graphs involving \\ multiple gradient updates and changes in variable dimensions.
 \end{tabular} & \begin{tabular}[c]{@{}c@{}} / \end{tabular}\\

Text2Immersion~\cite{ouyang2023text2immersion} & ArXiv & 2023 & \begin{tabular}[c]{@{}c@{}} Generation  \end{tabular} & 3D Scene-level & \begin{tabular}[c]{@{}c@{}} Generate diverse scenes with various objects, \\ and extend to the creation of imaginary scenes.
 \end{tabular} & \begin{tabular}[c]{@{}c@{}} Heavily dependent on monocular depth estimation \end{tabular}\\

AYG~\cite{ling2023align} & CVPR & 2024 & \begin{tabular}[c]{@{}c@{}} Generation  \end{tabular} & 4D Generation & \begin{tabular}[c]{@{}c@{}} (1) Combining 2D, 3D and video diffusion models to \\ solve text-to-4D tasks via score distillation. \\
(2) Regularization method and motion amplification technique.
 \end{tabular} & \begin{tabular}[c]{@{}c@{}} / \end{tabular}\\

DreamGaussian4D~\cite{ren2023dreamgaussian4d} & ArXiv & 2023 & \begin{tabular}[c]{@{}c@{}} Generation  \end{tabular} & 4D Generation & \begin{tabular}[c]{@{}c@{}} (1) Employing deformable 3D-GS representation in 4D content generation. \\
(2) An image-to-4D framework for learning \\ motion from image-conditioned generated videos. \\
(3) Video-to-video texture refinement strategy.
 \end{tabular} & \begin{tabular}[c]{@{}c@{}} / \end{tabular}\\

4DGen~\cite{yin20234dgen} & ArXiv & 2023 & \begin{tabular}[c]{@{}c@{}} Generation  \end{tabular} & 4D Generation & \begin{tabular}[c]{@{}c@{}} (1) Control the appearance and motion of 4D \\ assets by specifying monocular video sequences. \\
(2) Train at high resolutions and long frame lengths. \\
(3) Ensure spatiotemporal consistency from any angle and any time step.
 \end{tabular} & \begin{tabular}[c]{@{}c@{}} Multi-object generation is beyond assumption \end{tabular}\\

GSGen~\cite{chen2023text} & CVPR & 2024 & \begin{tabular}[c]{@{}c@{}} Generation  \end{tabular} & 3D Object-level & \begin{tabular}[c]{@{}c@{}} (1) Exploit joint guidance of 2D and 3D diffusion \\ prior to shaping a coherent rough structure in geometry optimization. \\ (2) Enrich appearance-refined details through compactness-based densification.
\end{tabular} & \begin{tabular}[c]{@{}c@{}} Subject to text prompts, containing complex \\ scene descriptions or complex logic \end{tabular}\\

AGG~\cite{chen2023text} & ArXiv & 2023 & \begin{tabular}[c]{@{}c@{}} Generation  \end{tabular} & 3D Object-level & \begin{tabular}[c]{@{}c@{}} (1) A cascade generation framework that renders 3D Gaussian distributions in one-shot. \\ (2) Generate coarse Gaussian predictions by decomposing a hybrid representation \\ of geometry and texture. 3D Gaussian properties can be jointly and stably optimized by \\ predicting geometry and texture information via two separate transformers. \\ (3) A UNet-based architecture with point-voxel layers \\ is introduced to super-resolve the 3D Gaussians.
\end{tabular} & \begin{tabular}[c]{@{}c@{}} Limit to represent very complex geometry. \end{tabular}\\

Efficient4D~\cite{pan2024fast} & ArXiv & 2023 & \begin{tabular}[c]{@{}c@{}} Generation  \end{tabular} & 4D Generation & \begin{tabular}[c]{@{}c@{}} Generate high-quality training data from large pre-trained models \\ without the need for heavily supervised backpropagation.
 \end{tabular} & \begin{tabular}[c]{@{}c@{}} Handle long-duration videos. \end{tabular}\\

GALA3D~\cite{zhou2024gala3d} & ArXiv & 2024 & \begin{tabular}[c]{@{}c@{}} Generation \end{tabular} & 3D Scene-level & \begin{tabular}[c]{@{}c@{}} (1) A compositional scene generation framework based on layout priors obtained from LLMs \\ and the layout-guided Gaussian representation. \\ (2) A compositional optimization strategy and a layout \\ refinement module are integrated to iteratively update Gaussian parameters and \\ scale of the layouts towards better alignment. \\ (3) This method generates high-quality complex 3D scenes \\ and supports interactive controllable editing.
 \end{tabular} & \begin{tabular}[c]{@{}c@{}} / \end{tabular}\\

IM3D~\cite{melas20243d} & ArXiv & 2024 & \begin{tabular}[c]{@{}c@{}} Generation \end{tabular} & 3D Object-level & \begin{tabular}[c]{@{}c@{}} (1) Improve multi-view generation through video instead of image generators.\\
(2) Produce high-quality 3D predictions combined with 3D-GS. \\
(3) Reduce the number of evaluations of the 2D generator network.
 \end{tabular} & \begin{tabular}[c]{@{}c@{}} / \end{tabular}\\

MVControl~\cite{li2024controllable} & ArXiv & 2024 & \begin{tabular}[c]{@{}c@{}} Generation \end{tabular} & 3D Object-level & \begin{tabular}[c]{@{}c@{}} (1) Controllable fine-grained text-to-multi-view image generation.\\
(2) A multi-stage but efficient 3D generation pipeline. \\
(3) First exploit the potential of a Gaussian-Mesh hybrid representation in 3D generation.
 \end{tabular} & \begin{tabular}[c]{@{}c@{}} / \end{tabular}\\

GVGEN~\cite{he2024gvgen} & ArXiv & 2024 & \begin{tabular}[c]{@{}c@{}} Generation \end{tabular} & 3D Object-level & \begin{tabular}[c]{@{}c@{}} (1) Design GaussianVolume, a structured, volumetric form consisting of 3D Gaussians.\\
(2) A highly efficient text-to-3D generation pipeline that operates in a coarse-to-fine manner. \\
(3) Involve generating a geometry volume as the initial step, \\ followed by the prediction of detailed 3D Gaussian attributes.
 \end{tabular} & \begin{tabular}[c]{@{}c@{}} 
 (1) Constrain when dealing with input texts significantly \\ divergent from the domain of training data.\\
(2) it is time-consuming to scale up to millions of object data.\\
(3) The volume resolution limits the rendering \\effects of 3D assets with very complex textures. \end{tabular}\\
 
STAG4D~\cite{zeng2024stag4d} & ArXiv & 2024 & \begin{tabular}[c]{@{}c@{}} Generation \end{tabular} & 4D Generation & \begin{tabular}[c]{@{}c@{}} (1) Employ the dynamic representation of 3D-GS for the purpose of \\ 4D generation, augmented by an adaptive densification strategy.\\
(2) A novel attention fusion module to integrate temporal anchor\\ frames into the process of multi-view diffusion. \\
 \end{tabular} & \begin{tabular}[c]{@{}c@{}} 
 (1) Constrained in handling fast and complicated motion.\\
(2) Inherent video limitations such as image blurriness.\\
(3) Domain transfer to videos lacking well-segmented foreground \\ or containing multiple foreground objects.
\end{tabular}\\

DreamPolisher~\cite{lin2024dreampolisher} & ArXiv & 2024 & \begin{tabular}[c]{@{}c@{}} Generation \end{tabular} & 3D Object-level & \begin{tabular}[c]{@{}c@{}} 
(1) Guided by geometric diffusion to get high-quality 3D objects.\\
(2) A ControlNet-based appearance refiner to improve the visual quality of the 3D objects. \\
(3) A geometric consistency loss to ensure the view consistency.
 \end{tabular} & \begin{tabular}[c]{@{}c@{}} 
(1) Take about 30 minutes for each 3D object.\\
(2) Rely on text instructions as input.\\
\end{tabular}\\

GaussianCube~\cite{zhang2024gaussiancube} & ArXiv & 2024 & \begin{tabular}[c]{@{}c@{}} Generation \end{tabular} & 3D Object-level & \begin{tabular}[c]{@{}c@{}} 
(1) Employ a novel densification-constrained fitting algorithm to \\ accurately fit a constant number of Gaussians to each 3D object.\\
(2) The obtained Gaussians are organized into a spatially structured representation by solving \\ the Optimal Transport problem between the positions of the Gaussians and the predefined voxel grid. \\
(3) GaussianCube exhibits a high level of expressiveness, efficiency, and a spatially coherent structure.
 \end{tabular} & \begin{tabular}[c]{@{}c@{}} 
/
\end{tabular}\\

GaussianFlow~\cite{gao2024gaussianflow} & ArXiv & 2024 & \begin{tabular}[c]{@{}c@{}} Generation \end{tabular} & 4D Generation & \begin{tabular}[c]{@{}c@{}} 
(1) Connect the three-dimensional Gaussian dynamics with the resultant pixel velocities.\\
(2) The Gaussian flow is obtained through splatting Gaussian dynamics into the image space. \\
(3) Implement the dynamics splatting in CUDA with minimal overhead.
 \end{tabular} & \begin{tabular}[c]{@{}c@{}} 
Long-term flow supervision across multiple frames.
\end{tabular}\\

BrightDreamer~\cite{jiang2024brightdreamer} & ArXiv & 2024 & \begin{tabular}[c]{@{}c@{}} Generation \end{tabular} & 3D Object-level & \begin{tabular}[c]{@{}c@{}} 
(1) Cpropose the Text-guided Shape Deformation (TSD) network to alleviate \\ the challenges associated with directly generating 3D Gaussians.\\
(2) Introduce the Text-guided Triplane Generator (TTG), which generates the spatial characteristics \\ of the object and subsequently decodes them into 3D Gaussians. \\
 \end{tabular} & \begin{tabular}[c]{@{}c@{}} 
Constrained to generating a singular 3D content for each input.
\end{tabular}\\

SyncTweedies~\cite{kim2024synctweedies} & ArXiv & 2024 & \begin{tabular}[c]{@{}c@{}} Generation \end{tabular} & \begin{tabular}[c]{@{}c@{}} 3D Object-level \& \\ 3D Scene-level \end{tabular}  & \begin{tabular}[c]{@{}c@{}} 
(1) Present a comprehensive investigation of all possible scenarios for \\ synchronizing multiple diffusion processes through a canonical space.\\
(2) Explore the practice of averaging the outputs of Tweedie's formula \\ while performing denoising in multiple instance spaces. \\
 \end{tabular} & \begin{tabular}[c]{@{}c@{}} 
Average the outputs of Tweedie's formula can still result in a decline in performance.
\end{tabular}\\

RealmDreamer~\cite{shriram2024realmdreamer} & ArXiv & 2024 & \begin{tabular}[c]{@{}c@{}} Generation \end{tabular} &  3D Scene-level  & \begin{tabular}[c]{@{}c@{}} 
(1) A robust scene initialization for scene-level 3D-GS with 2D priors.\\
(2) Learn consistent 3D with 2D inpainting priors, within an occluded volume.\\
(3) Distillation from diffusion-based depth estimators.
 \end{tabular} & \begin{tabular}[c]{@{}c@{}} 
 /
\end{tabular}\\

DreamScene360~\cite{zhou2024dreamscene360} & ArXiv & 2024 & \begin{tabular}[c]{@{}c@{}} Generation \end{tabular} &  3D Scene-level  & \begin{tabular}[c]{@{}c@{}} 
(1) Enable the unrestricted generation of high-quality, immersive 3D scenes with complete 360 \\ coverage, encompassing various styles, from textual inputs.\\
(2) Utilize Panorama as an intermediary tool, enabling us to engage in self-refinement through \\ the process of seeking quality assessment and prompt revision from GPT-4V. \\
(3) Construct a geometric field that initializes 3D Gaussians.
 \end{tabular} & \begin{tabular}[c]{@{}c@{}} 
Limited at the resolution of the text-to-image diffusion model.
\end{tabular}\\

Hash3D~\cite{yang2024hash3d} & ArXiv & 2024 & \begin{tabular}[c]{@{}c@{}} Generation \end{tabular} &  3D Object-level  & \begin{tabular}[c]{@{}c@{}} 
(1) A flexible and straightforward acceleration technique that does not require training. \\
(2) Utilize an adaptive grid-based hashing technique to effectively retrieve features.
 \end{tabular} & \begin{tabular}[c]{@{}c@{}} 
/
\end{tabular}\\

DreamScape~\cite{yuan2024dreamscape} & ArXiv & 2024 & \begin{tabular}[c]{@{}c@{}} Generation \end{tabular} &  3D Scene-level  & \begin{tabular}[c]{@{}c@{}} 
(1) 3DG$^2$ efficiently orchestrates the entire scene planning process by initializing scenes and \\ facilitating subsequent local-global 3D Gaussian optimization.\\
(2) A progressive scale constraint enables the model to modify the scale\\ proportions of objects while maintaining their visual representation.\\
(3) Propose sparse initialization and develop corresponding \\ densification and pruning strategies for such objects.
 \end{tabular} & \begin{tabular}[c]{@{}c@{}} 
(1) Necessitate a comparatively extensive guidance scale to \\ guarantee the convergence of the model due to SDS loss. \\
(2) The capabilities are limited by the single-object generation models.
\end{tabular}\\

Gaussian Decoder~\cite{barthel2024gaussian} & ArXiv & 2024 & \begin{tabular}[c]{@{}c@{}} Generation \end{tabular} &  3D Object-level  & \begin{tabular}[c]{@{}c@{}} 
(1) Enable the synthesis of explicit 3D-GS scenes using GANs. \\
(2) Avoid the usage of superresolution modules commonly employed \\ in generating implicit scene representations.\\
(3) A novel architecture for sequential decoding.\\
(4) Propose a strategy for sampling Gaussian splat positions around human heads, \\ and suggest a technique for fine-tuning the generator backbone to enhance the capacity of the decoders.
 \end{tabular} & \begin{tabular}[c]{@{}c@{}} 
(1) The output fidelity is bounded by the fidelity of 3D GANs.\\
(2) Eyes of the faces appear uncanny or blurry.
\end{tabular}\\

\bottomrule[1.5pt]
\end{tabular}%
}
\label{summary_generation}
\end{table*}

% Please add the following required packages to your document preamble:
% \usepackage{graphicx}
\begin{table*}[htbp]
\centering

\setlength\tabcolsep {1.5pt}
\renewcommand{\arraystretch}{1.2}
\caption{Summary of methods derived from 3D-GS for perception.}
\resizebox{\textwidth}{!}{%
\begin{tabular}{>{\columncolor{C7!50}}l>{\columncolor{C6!50}}c>{\columncolor{C5!50}}c>{\columncolor{C4!50}}c>{\columncolor{C3!50}}c>
{\columncolor{C2!50}}c>{\columncolor{C1!50}}c}
\toprule[1.5pt]
Methods   & Venue & Year & Category  &  Sub-category & Contributions                                                                                                                         & Limitations                                                                       \\ \midrule[1pt]

\begin{tabular}[c]{@{}c@{}} Language Embedded\\ 3D Gaussians~\cite{shi2023language} \end{tabular}&CVPR  & 2024 &Perception&Detection &\begin{tabular}[c]{@{}c@{}} 
(1) An efficient quantization scheme compressing\\ semantic features into dense 3D-GS.\\
(2) Utilizing spatial position and semantic uncertainty\\ to address semantic ambiguity from visual inconsistencies. \end{tabular} & \begin{tabular}[c]{@{}c@{}} (1) Struggling to accurately identify detailed object geometries at\\ high resolutions with CLIP-derived semantics.\\
(2) Difficulty in detecting highly reflective or translucent objects. \end{tabular}\\

Feature 3DGS~\cite{zhou2023feature}&CVPR&2024&Perception&Segmentation &\begin{tabular}[c]{@{}c@{}}(1) A method for
feature field distillation\\ using guidance from 2D foundation models based on 3D-GS. \\ (2) A distillation pipeline capable of\\ working with a variety of feature fields. \end{tabular} & \begin{tabular}[c]{@{}c@{}} (1) The student feature’s limited access to the GT\\ feature restricts the overall performance. \\(2) Challenging in handling noise-inducing floaters.\\(3) Introducing a bottleneck in rendering speed.\end{tabular}\\

SplaTAM~\cite{keetha2023splatam}&CVPR&2024&Perception&SLAM& \begin{tabular}[c]{@{}c@{}} An online SLAM system based on 3D-GS and\\ a silhouette mask for capturing scene density. \end{tabular} & \begin{tabular}[c]{@{}c@{}} Sensitive to motion blur, significant depth\\ noise, and aggressive rotation. \end{tabular}\\

GS-SLAM~\cite{yan2023gs}&CVPR&2024&Perception&SLAM&  \begin{tabular}[c]{@{}c@{}}(1) A dense RGB-D SLAM approach based on 3D-GS. \\(2) An adaptive 3D Gaussians expansion strategy.\\(3)  A coarse-to-fine technique to select reliable
3D\\ Gaussians to improve camera pose estimation.  \end{tabular} & \begin{tabular}[c]{@{}c@{}} Constrained by the quality of the obtained depth information.  \end{tabular}\\

Photo-SLAM~\cite{huang2023photo}&CVPR&2024&Perception&SLAM& \begin{tabular}[c]{@{}c@{}}(1) A 3D-GS SLAM system based on hyper primitives map. \\(2) Proposing Gaussian-Pyramid-based learning allowing the model to efficiently\\ and effectively learn multi-level features realizing high-fidelity mapping.  \end{tabular} & \begin{tabular}[c]{@{}c@{}} / \end{tabular}\\

\begin{tabular}[c]{@{}c@{}} Gaussian Splatting\\ SLAM~\cite{matsuki2023gaussian}\end{tabular}&CVPR&2024&Perception&SLAM&\begin{tabular}[c]{@{}c@{}}(1) A near real-time SLAM system based on 3D-GS. \\(2) The analytic Jacobian for camera pose estimation, \\Gaussian shape regularisation and geometric verification.  \end{tabular} & \begin{tabular}[c]{@{}c@{}} (1) Possibly constrained by the inevitable trajectory\\ drift in larger real-world scenes.\\(2) Hard real-time operation is not achieved. \end{tabular}\\

LangSplat~\cite{qin2023langsplat}&CVPR&2024&Perception&Detection\&Segmentation& \begin{tabular}[c]{@{}c@{}}(1) A GS-based method for 3D language fields. \\ (2) Learning the hierarchical semantics defined by SAM to address\\ the point ambiguity issue for 3D language field modeling. \end{tabular} & \begin{tabular}[c]{@{}c@{}} / \end{tabular}\\

SplatPose\&Detect~\cite{kruse2024splatpose} &CVPR Workshop& 2024 &Perception &Detection &  \begin{tabular}[c]{@{}c@{}} A pose-agnostic anomaly detection method based on 3D-GS. \\  \end{tabular} & \begin{tabular}[c]{@{}c@{}} / \end{tabular}\\

High-Fidelity SLAM~\cite{sun2024high}&IROS&2024&Perception&SLAM& \begin{tabular}[c]{@{}c@{}} (1) A densification strategy based on 3D-GS.\\(2) A regularized optimization during the mapping.  \end{tabular} & \begin{tabular}[c]{@{}c@{}}  Tracking accuracy may not be guaranteed\\ on real-world motion datasets.\end{tabular}\\

Gaussian-LIC~\cite{lang2024gaussian}&IROS&2024&Perception&SLAM&  \begin{tabular}[c]{@{}c@{}}(1) A 3D-GS-based SLAM system based\\ on LiDAR-Inertial-Camera fusion. \\(2) Strategies for the incremental mode.  \end{tabular} & \begin{tabular}[c]{@{}c@{}}(1) Redundant LiDAR points need to be discarded.\\(2) Constrained by auto-exposure and\\ motion blur of the cameras.  \end{tabular}\\

SGS-SLAM~\cite{li2024sgs}&ArXiv&2024&Perception&SLAM
& \begin{tabular}[c]{@{}c@{}}(1) A semantic dense visual SLAM system
based on 3D-GS. \\(2) A multi-channel parameter optimization strategy. \end{tabular} & \begin{tabular}[c]{@{}c@{}}  (1) Effectiveness will be compromised with\\ limited or inaccessible information.\\(2) Deploying to large scenes entails\\ significant memory consumption. \end{tabular}\\

SemGauss-SLAM~\cite{zhu2024semgauss}&ArXiv&2024&Perception&SLAM& \begin{tabular}[c]{@{}c@{}} (1) A 3D-GS semantic SLAM system with feature-level loss.\\(2) Semantic-informed bundle adjustment\\ utilizing multi-view semantic constraints.  \end{tabular} & \begin{tabular}[c]{@{}c@{}}/  \end{tabular}\\

Semantic Gaussians~\cite{guo2024semantic}&ArXiv&2024&Perception&Segmentation\&Tracking&\begin{tabular}[c]{@{}c@{}} (1) An approach to open-vocabulary 3D scene understanding by\\ bringing a novel semantic component to 3D-GS.\\
(2) A versatile semantic feature projection framework.  \end{tabular} & \begin{tabular}[c]{@{}c@{}} (1) Limited by the performance of 2D pre-trained\\ models and off-the-shelf 3D Gaussians.\\(2) 
Limited by the quality of novel perspective synthesis of 3D-GS. \end{tabular}\\

CG-SLAM~\cite{hu2024cg}&ArXiv&2024&Perception& SLAM&\begin{tabular}[c]{@{}c@{}}(1) A GPU-accelerated framework for real-time dense RGB-D SLAM based on 3D-GS. \\(2) Multiple loss terms and a depth uncertainty model.  \end{tabular} & \begin{tabular}[c]{@{}c@{}}  (1) Limited by memory usage.\\(2) A weak prediction ability for unobserved areas.\\(3) Not applicable to dynamic scenes.\end{tabular}\\

MM3DGS SLAM~\cite{sun2024mm3dgs}&ArXiv&2024&Perception&SLAM\&Dataset& \begin{tabular}[c]{@{}c@{}} (1) A multi-modal real-time 3D-GS SLAM framework based on 3D-GS.\\(2) A multi-modal indoor dataset collected using a mobile robot.  \end{tabular} & \begin{tabular}[c]{@{}c@{}} Constrained on scenes that involve longer trajectories. \end{tabular}\\

GS-ICP SLAM~\cite{ha2024rgbd}&ArXiv&2024&Perception&SLAM& \begin{tabular}[c]{@{}c@{}} (1) A real-time dense representation SLAM\\ that combines G-ICP and 3D-GS.\\(2) Shared covariances of G-ICP and 3D-GS\\ with scale aligning techniques.\end{tabular} & \begin{tabular}[c]{@{}c@{}} Limited by inherent depth noise in RGB-D sensors. \end{tabular}\\

GS-Pose~\cite{cai2024gs}&ArXiv&2024&Perception&Detection\&Tracking& \begin{tabular}[c]{@{}c@{}} (1) A multi-stage framework for object\\ localization and 6D pose estimation based on 3D-GS.\\(2) A generalizable co-segmentation approach.\\(3) A 3D-GS-based pose refinement and tracking methods. \end{tabular} & \begin{tabular}[c]{@{}c@{}}  Objects with slender or thin structures may pose challenges. \end{tabular}\\

\begin{tabular}[c]{@{}c@{}} Compact 3D-GS \\SLAM~\cite{deng2024compact}\end{tabular}&ArXiv&2024&Perception& SLAM&\begin{tabular}[c]{@{}c@{}} (1)  A GS-based SLAM system with compact scene representation.\\ (2) A sliding window-based online masking method.\\(3) A codebook-based method to restore\\ the geometry of Gaussian points. \end{tabular} & \begin{tabular}[c]{@{}c@{}} / \end{tabular}\\

3DGS-ReLoc~\cite{jiang20243dgs}&ArXiv&2024&Perception&SLAM& \begin{tabular}[c]{@{}c@{}} An approach to creating detailed and \\geometrically accurate environmental representations based on 3D-GS.\\  \end{tabular} & \begin{tabular}[c]{@{}c@{}}  Artifacts may appear in rendered images.\end{tabular}\\

NEDS-SLAM~\cite{ji2024neds}&ArXiv&2024&Perception&SLAM&  \begin{tabular}[c]{@{}c@{}} (1) A fusion module combining semantic\\ features with appearance features.\\(2) A semantic SLAM framework based on 3D-GS.\\(3) A virtual camera view pruning method to filter noisy Gaussians.  \end{tabular} & \begin{tabular}[c]{@{}c@{}} Virtual view pruning method increases computational load\\ and can impact real-time performance. \end{tabular}\\

Gaussian Grouping~\cite{ye2023gaussian}&ArXiv&2023&Perception&Segmentation\&Dataset&\begin{tabular}[c]{@{}c@{}} (1) A joint reconstruction and segmentation method in the open-world 3D scene based on 3D-GS. \\ (2) An Identity Encoding for 3D Gaussians that is supervised both by\\ 2D mask predictions from SAM and 3D spatial consistency.\end{tabular} & \begin{tabular}[c]{@{}c@{}}  Limited to the static 3D scene. \end{tabular}\\

SAGA~\cite{cen2023segment}&ArXiv&2023&Perception&Segmentation&  \begin{tabular}[c]{@{}c@{}}(1) An interactive 3D segmentation method based on 3D-GS and SAM.\\ (2) Losses designed for distilling knowledge\\ from the SAM into 3D Gaussians. \end{tabular} & \begin{tabular}[c]{@{}c@{}}(1) The Gaussians learned by 3D-GS are ambiguous\\ without
any constraint on geometry.\\(2) The masks automatically extracted by SAM\\ tend to exhibit a certain level of noise.  \end{tabular}\\

%iComMa~\cite{sun2023icomma}&ArXiv&2023&Perception&Calibration&\begin{tabular}[c]{@{}c@{}}(1) A novel 6D camera pose estimation method by inverting 3D-GS. \\(2) Integration of a matching strategy. \\(3) A render-and-compare method to ensure accuracy. \end{tabular} & \begin{tabular}[c]{@{}c@{}}  \end{tabular}\\

Gaussian-SLAM~\cite{yugay2023gaussian}&ArXiv&2023&Perception&SLAM&  \begin{tabular}[c]{@{}c@{}}(1) A dense RGBD SLAM approach based on 3D-GS. \\ (2) An extension of GS that better encodes geometry and allows\\ reconstruction beyond radiance fields in a single-camera setup. \\(3) An online optimization method for GS and a frame-to-model\\ tracker with the GS scene representation.\end{tabular} & \begin{tabular}[c]{@{}c@{}}(1) Challenging in tackling motion blur and low-quality depth maps. \\(2) The trajectory drift is inevitable.\end{tabular}\\

\begin{tabular}[c]{@{}c@{}} 2D-Guided 3D Gaussian\\ Segmentation~\cite{lan20232d} \end{tabular}&ArXiv&2023&Perception&Segmentation&  \begin{tabular}[c]{@{}c@{}}  A 3D Gaussian segmentation
method supervised by 2D segmentation. \end{tabular} & \begin{tabular}[c]{@{}c@{}} / \end{tabular}\\

Street Gaussians~\cite{yan2024street}&ArXiv&2024&Perception&Detection\&Tracking& \begin{tabular}[c]{@{}c@{}} (1)  A scene representation for modeling complex dynamic street scenes.\\(2) A tracked pose optimization strategy along
with a 4D spherical\\ harmonics appearance model to handle the dynamics of moving vehicles.  \end{tabular} & \begin{tabular}[c]{@{}c@{}} (1) limited to reconstructing rigid dynamic scenes.\\(2) Dependence on the recall rate of off-the-shelf trackers. \end{tabular}\\

FMGS~\cite{zuo2024fmgs}&ArXiv&2024&Perception&Detection\&Segmentation&  \begin{tabular}[c]{@{}c@{}} (1) An approach combining 3D-GS for geometry and appearance \\representation with MHE for efficient language embedding.\\(2) a multi-view consistent language embedding method.\\(3) Addressing pixel misalignment. \end{tabular} & \begin{tabular}[c]{@{}c@{}}   (1) Relying heavily on the presence of high-quality\\ and calibrated input images.\\(2) The performance is entirely contingent on the\\ quality of the base foundation models.\end{tabular}\\

Splat-Nav~\cite{chen2024splat}&ArXiv&2024&Perception  &SLAM &\begin{tabular}[c]{@{}c@{}} (1) A navigation pipeline for robots based on 3D-GS.\\(2)  A guaranteed-safe planning module and a pose estimation module.  \end{tabular} & \begin{tabular}[c]{@{}c@{}} / \end{tabular}\\

Beyond Uncertainty~\cite{liu2024beyond}&ArXiv&2024&Perception  &SLAM&\begin{tabular}[c]{@{}c@{}} An active view acquisition framework that focuses on enhancing the risk and\\ safety assessment while concurrently advancing 3D scene construction. \\  \end{tabular} & \begin{tabular}[c]{@{}c@{}} / \end{tabular}\\

GaussNav~\cite{lei2024gaussnav}&ArXiv&2024&Perception  &SLAM&\begin{tabular}[c]{@{}c@{}}(1) A modular approach for visual navigation based on 3D-GS.\\(2) A novel map representation that maintains 3D scene geometry, semantic\\ labels for each Gaussian element, and detailed textures.  \end{tabular} & \begin{tabular}[c]{@{}c@{}}  (1) Inability to consistently match the target from instance renderings.\\(2) Inaccuracies in goal localization. \end{tabular}\\

%3DGS-Calib~\cite{herau20243dgs}&ArXiv&2024&Perception&Calibration&\begin{tabular}[c]{@{}c@{}}(1) A multimodal spatiotemporal calibration method based on 3D-GS. \\  \end{tabular} & \begin{tabular}[c]{@{}c@{}}Assuming LiDAR points concentrate in the bottom half of the images.  \end{tabular}\\
\bottomrule[1.5pt]
\end{tabular}%
}
\label{summary_perception}
\end{table*}

% Please add the following required packages to your document preamble:
% \usepackage{graphicx}
\begin{table*}[htbp]
\centering

\setlength\tabcolsep {1.5pt}
\renewcommand{\arraystretch}{1.2}
\caption{Summary of methods derived from 3D-GS for virtual human.}
\resizebox{\textwidth}{!}{%
\begin{tabular}{>{\columncolor{C7!50}}l>{\columncolor{C6!50}}c>{\columncolor{C5!50}}c>{\columncolor{C4!50}}c>{\columncolor{C3!50}}c>{\columncolor{C2!50}}c>{\columncolor{C1!50}}c}
\toprule[1.5pt]
Methods   & Venue & Year & Category  & Sub-category & Contributions                                                                                                                         & Limitations                                                                       \\ \midrule[1pt]

HuGS~\cite{moreau2023human} & CVPR & 2024 & Virtual Human & Human bodies &  \begin{tabular}[c]{@{}c@{}} (1) A human body model for novel view synthesis from multi-view videos. \\ (2) Skinning weight vector for skeleton-based rigid motions and a local \\ non-rigid refinement for the non-rigid garment deformation.  \end{tabular} & \begin{tabular}[c]{@{}c@{}} Only optimize and deform each Gaussian independently, \\ overlooking the intrinsic relation between Gaussians in local neighbourhoods. \end{tabular}\\ 

Animatable Gaussians~\cite{li2023animatable} & CVPR & 2024 & Virtual Human & Human bodies &  \begin{tabular}[c]{@{}c@{}} 
(1) A new avatar representation that leverages 2D CNNs \\ and 3D-GS to create high-fidelity avatar. \\ 
(2) Utilize two pose-dependent canonical Gaussian maps to \\ associate 3D-GS with the SMPL template. \\ 
(3) Introduce a pose projection strategy that employs PCA on the \\ driving signal for better generalization given novel poses. \end{tabular} & \begin{tabular}[c]{@{}c@{}} (1) Entangle the modeling of the human body and clothes. \\ 
(2) Rely on the multi-view input to recover a parametric template. \end{tabular}\\

ASH~\cite{pang2023ash} & CVPR & 2024 & Virtual Human &  Human bodies & \begin{tabular}[c]{@{}c@{}} 
(1) Attach the Gaussians onto a deformable character model \\ and learn their parameters in 2D texture space. \\ 
(2) Adopt a 2D geometry decoder and a 2D appearance decoder to predict the \\ shape-related parameters and the Spherical Harmonics separately. \end{tabular} & \begin{tabular}[c]{@{}c@{}} Unable to update the underlying deformable template
mesh.\end{tabular}\\

HiFi4G~\cite{jiang2023hifi4g} & CVPR & 2024 & Virtual Human & Human bodies &  \begin{tabular}[c]{@{}c@{}} 
(1) Present a compact 4D Gaussian representation for human \\ performance rendering with traditional non-rigid tracking. \\ 
(2) A dual-graph mechanism to organically bridge 3D-GS with non-rigid tracking. \\ 
(3) Achieve a substantial compression rate. \end{tabular} & \begin{tabular}[c]{@{}c@{}}  
(1) Heavily reliance on segmentation.\\ 
(2) Necessitate per-frame reconstruction and mesh tracking. \\ 
(3) Long Gaussian optimization time. \end{tabular}\\

D3GA~\cite{zielonka2023drivable} & ArXiv & 2023 & Virtual Human & Human bodies &  
\begin{tabular}[c]{@{}c@{}} 
(1) Employing cage-based deformation to transform Gaussians to represent drivable human avatars. \\
(2) Compositional deformation where separate cages deform body, face and garment avatar. 
\end{tabular} & 
\begin{tabular}[c]{@{}c@{}}  
(1) High-frequency patterns, like stripes, may
result in blurry regions. \\
(2) Self-collisions for loose garments are still challenging.
\end{tabular}\\

HUGS~\cite{kocabas2023hugs} & CVPR & 2024 & Virtual Human & Human bodies &  
\begin{tabular}[c]{@{}c@{}} 
Supporting both the novel pose synthesis of the 
human and the novel view \\ synthesis of the human and the scene from only a monocular video. 
\end{tabular} & 
\begin{tabular}[c]{@{}c@{}}  
(1) Unable to capture the general deformable structure of loose clothing such as dresses. \\
(2) Trained on in-the-wild videos that do not cover the pose-space of the human body. \\
(3) Does not account for environment lighting. \\
\end{tabular}\\

ParDy-Human~\cite{jung2023deformable} & ArXiv & 2023 & Virtual Human & Human bodies &  
\begin{tabular}[c]{@{}c@{}} 
(1) Provide a mask-free training approach that automatically separates background and human. \\
(2) The human motion is represented by two consecutive deformations, \\ a naked body deformation followed by motion-related cloth movement. 
\end{tabular} & 
\begin{tabular}[c]{@{}c@{}}  
Uni-coloured garments with large deformations can pose problems \\ for animations of out-of-distribution poses. 
\end{tabular}\\

Human101~\cite{li2023human101} & ArXiv & 2023 & Virtual Human & Human bodies &  
\begin{tabular}[c]{@{}c@{}} 
(1) Propose a Canonical Human Initialization to enhance the model’s \\ convergence rate and improve visual detail representation. \\
(2) A deformation methodology that is composed of Human-centric \\ Forward Gaussian Animation and Human-centric Gaussian \\ Refinement for fast reconstruction and real-time rendering. 
(3) Train 3D-GS within 100 seconds and \\ achieve a rendering speed over 100 FPS. 
\end{tabular} & 
\begin{tabular}[c]{@{}c@{}}  
/
\end{tabular}\\

3DGS-Avatar~\cite{qian20233dgs} & CVPR & 2024 & Virtual Human & Human bodies &  
\begin{tabular}[c]{@{}c@{}} 
(1) Integrate rigid human articulation with pose-dependent non-rigid garment deformation. \\
(2) Decode color with a small MLP instead of learning spherical harmonic coefficients. \\
(3) An as-isometric-as-possible regularization to maintain the \\ geometric consistency and realistic deformation of the avatar. 
\end{tabular} & 
\begin{tabular}[c]{@{}c@{}}  
/
\end{tabular}\\

SplatArmor~\cite{jena2023splatarmor} & ArXiv & 2023 & Virtual Human & Human bodies &  
\begin{tabular}[c]{@{}c@{}} 
Propose a neural color field to implicitly \\ regularize the color of nearby Gaussians. 
\end{tabular} & 
\begin{tabular}[c]{@{}c@{}}  
(1) Bad initialization can lead to texture artifacts.\\
(2) Inconsistent texture in unseen regions or
regions with self-shadows. \\
(3) Generalization to unseen poses can not be guaranteed.
\end{tabular}\\

GaussianAvatar~\cite{qian2023gaussianavatars} & CVPR & 2024 & Virtual Human & Human bodies &  
\begin{tabular}[c]{@{}c@{}} 
(1) Enable the creation of animatable head avatars \\ by rigging 3D Gaussians to parametric mesh models.\\
(2) A binding inheritance strategy that effectively supports \\ the addition and removal of 3D Gaussians while maintaining controllability.
\end{tabular} & 
\begin{tabular}[c]{@{}c@{}}  
(1) Relighting the avatar is not feasible.\\
(2) Lack of control over areas that are not modeled by FLAME.
\end{tabular}\\

GauHuman~\cite{hu2023gauhuman} & CVPR & 2024 & Virtual Human & Human bodies &  
\begin{tabular}[c]{@{}c@{}} 
(1) A pose refinement module to rectify SMPL poses for more accurate LBS transformation.\\
(2) Integrate the KL divergence to regulate the split and clone process.\\
(3) A novel merge operator to merge 3D Gaussians for speeding up. 
\end{tabular} & 
\begin{tabular}[c]{@{}c@{}}  
(1) The current pipeline does not support extracting 3D meshes.\\
(2) Recovering 3D human details like cloth wrinkles from \\ monocular videos is still a challenging problem. 
\end{tabular}\\

GaussianBody~\cite{li2024gaussianbody} & ArXiv & 2023 & Virtual Human & Human bodies &  
\begin{tabular}[c]{@{}c@{}} 
(1) Propose to preserve the local geometry property of 3D \\ Gaussians with three regularization losses. \\
(2) A split-with-scale operation to address the potential blurries and oversized Gaussians. \\
(3) A pose refinement to optimize the SMPL pose parameters.
\end{tabular} & 
\begin{tabular}[c]{@{}c@{}}  
/
\end{tabular}\\

SplattingAvatar~\cite{shao2024splattingavatar} & CVPR & 2024 & Virtual Human & Human bodies &  
\begin{tabular}[c]{@{}c@{}} 
(1) Embed the Gaussians on the mesh for explicit control over pose-dependent parameters. \\
(2) Disentangle motion and appearance with explicit
mesh geometry and implicit modeling.\\
(3) Apply lifted optimization to jointly optimize the embedding points and the triangle mesh.
\end{tabular} & 
\begin{tabular}[c]{@{}c@{}}  
Unable to provide separate motion representations for clothes
and hair. 
\end{tabular}\\

HAHA~\cite{svitov2024haha} & ArXiv & 2024 & Virtual Human & Human bodies & 
\begin{tabular}[c]{@{}c@{}} 
(1) Represent the human avatar with the combination of 3D-GS and textured meshes.\\
(2) Utilize SMPL-X parametric human model to \\ handle the animation of highly articulated parts. \\
(3) Generate Gaussians only in out-of-mesh details and  \\ significantly reduce the amount of Gaussians. 
\end{tabular} & 
\begin{tabular}[c]{@{}c@{}}  
Obtaining an accurate SMPL-X mesh from a monocular video is still challenging. 
\end{tabular}\\

OccGaussian~\cite{ye2024occgaussian} & ArXiv & 2024 & Virtual Human & Human bodies & 
\begin{tabular}[c]{@{}c@{}} 
(1) Render human avatars under the obstructed scenarios.\\ 
(2) Replace occluded points with the aggregated pixel-align features from visible points.\\
(3) Design two occlusion-aware loss functions.
\end{tabular} & 
\begin{tabular}[c]{@{}c@{}}  
Can't fully recover regions that are occluded for extended periods. 
\end{tabular}\\

GoMAvatar~\cite{wen2024gomavatar} & CVPR & 2024 & Virtual Human & Human bodies & 
\begin{tabular}[c]{@{}c@{}} 
(1) A Gaussian-on-Mesh representation combining 3D-GS with \\ skeleton-based deformable meshes. \\
(2) A unique differentiable shading module for view dependency.\\
(3) Achieve real-time and free-viewpoint rendering that is compatible with graphics engines.
\end{tabular} & 
\begin{tabular}[c]{@{}c@{}}  
/
\end{tabular}\\

GaussianAvatars~\cite{qian2023gaussianavatars} & CVPR & 2024 & Virtual Human & Human heads &  
\begin{tabular}[c]{@{}c@{}} 
(1) A head avatar model rigging 3D Gaussians to a parametric morphable face model FLAME. \\
(2) A binding strategy to accommodate the adaptive density control operations \\ without breaking the connection between triangles and splats.
\end{tabular} & 
\begin{tabular}[c]{@{}c@{}}  
(1) The avatar relighting is not feasible.\\
(2) Garments that are not modeled by the FLAME system lack controllability.
\end{tabular}\\

HeadGaS~\cite{dhamo2023headgas} & ArXiv & 2023 & Virtual Human & Human heads &  
\begin{tabular}[c]{@{}c@{}} 
(1) A head animation method that can work with different 3D \\ morphable models from monocular video inputs. \\
(2) Extending Gaussian representation with a base of latent \\ features for head expression controllability. 
\end{tabular} & 
\begin{tabular}[c]{@{}c@{}}  
(1) Failures of the head tracker may result in inaccurate  \\ facial expressions or blurriness. \\
(2) Expressions spread among different views are \\ required for reasonable convergence.
\end{tabular}\\

PSAvatar~\cite{zhao2024psavatar} & ArXiv & 2024 & Virtual Human & Human heads &  
\begin{tabular}[c]{@{}c@{}} 
(1) A Point-based Morphable Shape Model capable of modeling non-facial structures.  \\
(2) Conduct in an analysis-by-synthesis manner with point  \\ splatting instead of Gaussian splatting for faster convergence. 
\end{tabular} & 
\begin{tabular}[c]{@{}c@{}}  
/
\end{tabular}\\

MonoGaussianAvatar~\cite{chen2023monogaussianavatar} & ArXiv & 2023 & Virtual Human & Human heads &  
\begin{tabular}[c]{@{}c@{}} 
(1) A Gaussian deformation field to adjust Gaussian parameters.\\ 
(2) A Gaussian points insertion and deletion
strategy.
\end{tabular} & 
\begin{tabular}[c]{@{}c@{}}  
(1) Unable to model the reflection of eyeglass lenses. \\
(2) Expression ability is limited by the prior knowledge derived from the 3DMM. 
\end{tabular}\\

Gaussian Head Avatar~\cite{xu2023gaussian} & CVPR & 2024 & Virtual Human & Human heads &  
\begin{tabular}[c]{@{}c@{}} 
(1) A fully learnable expression-conditioned deformation field \\ to avoid the limited capability of the LBS-based formulation.\\ 
(2) A geometry-guided initialization strategy by optimizing an \\ SDF field, a color field and a deformation field to extract \\ a mesh and construct Gaussians.
\end{tabular} & 
\begin{tabular}[c]{@{}c@{}}  
Blurs sometimes exist due to the lack of tracking methods. 
\end{tabular}\\

GaussianHead~\cite{wang2023gaussianhead} & ArXiv & 2024 & Virtual Human & Human heads &  
\begin{tabular}[c]{@{}c@{}} 
(1) Disentangle the modeling of the head geometry and texture attributes \\ with a motion deformation field and a multi-resolution tri-plane. \\ 
(2) A Gaussian derivation strategy to tackle the feature dilution problem. 
\end{tabular} & 
\begin{tabular}[c]{@{}c@{}}  
Does not separate the motion of the head and torso.
\end{tabular}\\

Rig3DGS~\cite{rivero2024rig3dgs} & ArXiv & 2024 & Virtual Human & Human heads &  
\begin{tabular}[c]{@{}c@{}} 
(1) Generate reanimatable portraits with a learnable prior based \\deformation field derived from the FLAME mesh. \\
(2) Support full control over facial expressions, head-pose \\ and novel view synthesis from causal phone videos. 
\end{tabular} & 
\begin{tabular}[c]{@{}c@{}}  
(1) Unable to model illumination with non-uniformity.\\
(2) During filming, the subject is required to remain relatively still. 
\end{tabular}\\

SplatFace~\cite{luo2024splatface} & ArXiv & 2024 & Virtual Human & Human heads &  
\begin{tabular}[c]{@{}c@{}} 
(1) A face reconstruction framework incorporating a 3D \\ morphable model to produce realistic faces from only a limit of input images.\\
(2) A non-rigid alignment process to jointly optimize Gaussians and the morphable model surface. \\
(3) A splat-to-surface distance metric leveraging Gaussian \\ position and covariance to enhance alignment. \\
(4) A world-space densification process to complement the \\ original adaptive densification operation in 3D-GS.
\end{tabular} & 
\begin{tabular}[c]{@{}c@{}}  
(1) Over-regularization in areas where the \\ morphable model has insufficient expressiveness.\\
(2) Only applicable in scenes that the surface model can reach. \\
(3) Artifacts still exist. 
\end{tabular}\\

GaussianTalker~\cite{yu2024gaussiantalker} & ArXiv & 2024 & Virtual Human & Human heads &  
\begin{tabular}[c]{@{}c@{}} 
(1) A talking head synthesis method combining 3D-GS with the FLAME \\ model for vivid lip movements and realistic visual effects. \\
(2) A speaker-specific motion translator to extract  \\speaker-agnostic features and fuse with the target  \\identity embedding to produce more target-aligned poses and expressions. \\
(3) A dynamic Gaussian renderer with a set of speaker-specific \\ blend shapes for enhanced visual effects.
\end{tabular} & 
\begin{tabular}[c]{@{}c@{}}  
/
\end{tabular}\\

TalkingGaussian~\cite{yu2024gaussiantalker} & ArXiv & 2024 & Virtual Human & Human heads &  
\begin{tabular}[c]{@{}c@{}} 
(1) Address the facial distortion problem in dynamic areas by \\ applying deformations on persistent head Gaussians. \\
(2) A face-mouth decomposition module to realize lip \\ synchronization and high-fidelity mouth structure. 
\end{tabular} & 
\begin{tabular}[c]{@{}c@{}}  
/
\end{tabular}\\

\bottomrule[1.5pt]
\end{tabular}%
}
\label{summary_human}
\end{table*}

\ifCLASSOPTIONcaptionsoff
  \newpage
\fi

\newpage
\clearpage

\bibliographystyle{IEEEtran}\balance
\bibliography{ref}